\documentclass{article}

\usepackage{arxiv}

\usepackage[utf8]{inputenc} 
\usepackage[T1]{fontenc}    

\usepackage{natbib} 
\usepackage{mathtools} 
\usepackage{amsfonts}
\usepackage{booktabs} 

\usepackage{algorithmic}
\usepackage{graphicx}
\usepackage{textcomp}
\usepackage{hyperref}

\usepackage{caption}
\usepackage{subcaption}
\usepackage{float}
\usepackage{bm} 
\usepackage{listings}
\usepackage[table]{xcolor}
\usepackage{geometry}
\usepackage{adjustbox} 
\usepackage{makecell}  
\usepackage{array}
\usepackage{authblk}

\hypersetup{
    colorlinks=true,
    citecolor=blue,
    linkcolor=blue,
    urlcolor=blue
}

\title{ConDiSim: Conditional Diffusion Models for Simulation-Based Inference}

\author[1]{Mayank Nautiyal}

\author[2]{Andreas Hellander}
\author[1]{Prashant Singh}

\affil[1]{Science for Life Laboratory, Uppsala University}

\affil[2]{Uppsala University}
\date{}
  
\begin{document}
\maketitle

\begin{abstract}
We present ConDiSim, a conditional diffusion model for simulation-based inference in complex systems with intractable likelihoods. ConDiSim leverages denoising diffusion probabilistic models to approximate posterior distributions, consisting of a forward process that adds Gaussian noise to parameters, and a reverse process learning to denoise, conditioned on observed data. This approach effectively captures complex dependencies and multi-modalities within posteriors. ConDiSim is evaluated across ten benchmark problems and two real-world test problems, where it demonstrates effective posterior approximation accuracy while maintaining computational efficiency and stability in model training. ConDiSim provides a robust and extensible framework for simulation-based inference, well suited to parameter estimation tasks that demand fast methods for handling noisy, time series observations.
\end{abstract}

\section{Introduction}
Statistical inference of model parameters from empirical observations is a fundamental challenge in scientific research, enabling researchers to derive meaningful insights from complex simulation models. These parameters govern the behavior of simulators that replicate real-world phenomena, providing a bridge between theoretical constructs and empirical observations \citep{Lavin2021SimulationIT}. Calibrating these parameters to ensure that simulator outputs align with observed data constitutes an inverse problem, formally defined within the framework of simulation-based inference (SBI) \citep{Cranmer20}. Such inverse problems are often ill-posed, meaning they may lack unique or stable solutions, especially when the simulator is stochastic or when different parameter configurations produce similar observations \citep{sisson2018handbook}. To address these challenges, the Bayesian formulation of SBI places a prior distribution over the parameters and infers a posterior distribution conditioned on observed data. This posterior captures the full set of plausible solutions and enables principled uncertainty quantification, offering a more robust alternative to relying on single-point estimates.

Recent advances in machine learning have introduced powerful generative models capable of approximating complex data distributions without explicit likelihood computations. Among these, diffusion models have emerged as a particularly promising class \citep{pmlr-v37-sohl-dickstein15, NEURIPS2020_4c5bcfec, pmlr-v115-song20a}. They model data by learning to reverse a fixed noising process, gradually denoising samples from pure noise. Moreover, diffusion models naturally support amortized inference \citep{kingma2022autoencodingvariationalbayes}: a single trained model approximates posteriors for any observation without per-instance optimization, enabling faster inference for new data. 

This work explores diffusion models for addressing common SBI challenges: limited simulation budgets from expensive simulators (particularly in systems biology and epidemiology with stiff ODE/PDE solvers), noisy and stochastic outputs, multimodal posteriors from nonlinearities and identifiability issues, and the need for architectures that integrate parameters, timesteps, and observations to recover diverse posterior structures. ConDiSim is demonstrated to be well-suited for SBI problems involving learning low-dimensional posteriors from noisy, high-dimensional discrete-time observations.

\textbf{Contributions:} (i)We adapt conditional Denoising Diffusion Probabilistic Models (DDPMs) for SBI to learn complex posterior distributions under limited simulation budgets; (ii) ConDiSim introduces a novel reverse diffusion architecture for SBI that employs time-dependent Feature-wise Linear Modulation (FiLM) to jointly condition on noisy parameters, diffusion timesteps, and observations for posterior estimation; (iii) We empirically evaluate ConDiSim systematically across benchmarks and real-world SBI tasks, analyzing performance under limited simulation budgets, noisy outputs, and multimodal posteriors.

\textbf{Outline}: The paper is organized as follows. Section~\ref{sec:related} outlines the related work. Section~\ref{Problem Formulation} formally defines the SBI problem. Section~\ref{condisim_method} introduces the ConDiSim model, and Section~\ref{condition} details its conditioning mechanism and sampling procedure, including the proposed architecture. Section~\ref{Experiments} presents a comprehensive evaluation on benchmark and real-world problems, with additional results provided in the Appendix. Section~\ref{Discussion} presents a discussion of the results, while Section~\ref{Conclusion} concludes the paper.

\section{Related Work}\label{sec:related}

Simulation-based inference has advanced significantly in recent years, evolving from traditional statistical likelihood-free methods to contemporary deep learning frameworks. Classical techniques like Approximate Bayesian Computation (ABC) \citep{Beaumont2002} and Sequential Monte Carlo-ABC (SMC-ABC) \citep{Toni2009} offer theoretical convergence guarantees but may scale poorly to high dimensional problems \citep{sisson2018handbook}. While neural summary statistics \citep{jiang2017learning, wrede2022robust} can reduce dimensionality, they introduce trade-offs: additional computational burden and the possibility of losing information necessary for accurate posterior estimation.

Neural Posterior Estimation (NPE) \citep{papamakarios2016fast} pioneered the use of deep generative models for amortized SBI, introducing conditional normalizing flows to model the posterior distribution, substantially improving accuracy and scalability over traditional methods. Sequential extensions like SNPE-A \citep{papamakarios2018fast}, SNPE-B \citep{NIPS2017_addfa9b7}, and SNPE-C (also known as Automatic Posterior Transformation or APT) \citep{Greenberg19} enhanced accuracy by iteratively refining the model to focus on relevant parameter regions. However, these sequential methods require multiple rounds of training and are constrained by the invertible transformations required in flow-based models. Recently, Flow Matching Posterior Estimation (FMPE) \citep{fmpe} introduced continuous normalizing flows based on conditional vector fields, overcoming architectural constraints and efficiently scaling to high-dimensional inference tasks. However, FMPE requires ODE solvers at inference time, introducing additional computational costs, and its performance depends critically on the choice of flow interpolant, which can be challenging to optimize under tight simulation budgets.

Beyond flow-based methods, other deep generative approaches to SBI include adversarial models. Generative Adversarial Training for SBI (GATSBI) \citep{Ramesh22} uses adversarial networks to model posteriors but suffers from training instability, mode collapse \citep{Arjovsky17a,Arjovsky17b}, and poor performance in low-dimensional settings \citep{Ramesh22}. More recently, Simformer \citep{gloecklerall} combined score-based diffusion models with a transformer architecture, enabling high-fidelity posterior estimation but at the expense of increased computational cost. Score-based diffusion models have also 
been explored in the tall-data setting of SBI, where multiple observations are combined efficiently using 
amortized score networks \citep{Linhart2024DiffusionPS}. In contrast, ConDiSim uses DDPMs with a lightweight, FiLM-based architecture that avoids heavy attention mechanisms while achieving competitive performance with significantly lower inference and training overhead. This makes it especially suitable for settings like iterative model exploration, where many inference queries must be resolved quickly \citep{10.1093/bioinformatics/btz420}.

In this study, we evaluate ConDiSim against five distinct SBI approaches: NPE \citep{papamakarios2016fast}, APT \citep{Greenberg19}, GATSBI \citep{Ramesh22}, FMPE \citep{fmpe}, and Simformer \citep{gloecklerall}.
Together, these methods span a wide spectrum of generative modeling paradigms, including discrete flow-based methods in both one-shot and sequential forms, continuous flow-based approaches, adversarial models, and diffusion models in both discrete and continuous formulations. This selection enables a balanced evaluation that emphasizes efficiency and scalability in addition to posterior accuracy. For a broader overview of deep learning approaches to SBI, see \citep{zammit2024neural, Radev23}.

\section{Simulation-Based Inference}\label{Problem Formulation}
Simulation-based inference (SBI) aims to infer the posterior distribution \( p(\boldsymbol{\theta} \mid \mathbf{y}) \) of model parameters \( \boldsymbol{\theta} \), given observed data \( \mathbf{y} \). From Bayes' theorem:
\begin{equation}
p(\boldsymbol{\theta} \mid \mathbf{y}) \propto p(\mathbf{y} \mid \boldsymbol{\theta}) \, p(\boldsymbol{\theta}),
\label{eq:bayes}
\end{equation}
where \( p(\mathbf{y} \mid \boldsymbol{\theta}) \) is the likelihood and \( p(\boldsymbol{\theta}) \) is the prior distribution. In many scientific  models, particularly those involving stochastic simulators, the likelihood function \( p(\mathbf{y} \mid \boldsymbol{\theta}) \) is either intractable or computationally expensive to evaluate, rendering likelihood-based methods such as Markov Chain Monte Carlo (MCMC) infeasible \citep{mcmc_ml}.
 
When the likelihood is inaccessible, a common strategy is to sample parameters from a prior, 
$\boldsymbol{\theta} \sim p(\boldsymbol{\theta})$, where $p(\boldsymbol{\theta})$ may be informed by domain knowledge 
or chosen to be noninformative, such as a broad uniform prior. For each sampled parameter, a corresponding observation 
is generated via the simulator, $\mathbf{y} \sim p(\mathbf{y}\mid\boldsymbol{\theta})$. This yields joint samples 
$(\boldsymbol{\theta}, \mathbf{y}) \sim p(\boldsymbol{\theta}, \mathbf{y})$, which can then be used to train generative models that approximate the posterior 
$p(\boldsymbol{\theta}\mid \mathbf{y})$ without requiring explicit likelihood evaluation.

\subsection{ConDiSim: Conditional Diffusion Framework for SBI} \label{condisim_method}
We adapt conditional denoising diffusion probabilistic models \citep{NEURIPS2020_4c5bcfec} to SBI, where the goal is to approximate the posterior \(p(\boldsymbol{\theta} \mid \mathbf{y})\) in likelihood-free settings using joint samples \((\boldsymbol{\theta}, \mathbf{y}) \sim p(\boldsymbol{\theta}, \mathbf{y})\). The model employs a forward diffusion process that gradually corrupts the parameters \(\boldsymbol{\theta}\) with Gaussian noise until they follow a tractable prior distribution. The reverse process then learns to denoise these corrupted parameters, conditioned on observations \(\mathbf{y}\), thereby recovering samples from the posterior distribution \(p(\boldsymbol{\theta} \mid \mathbf{y})\).

In the \textbf{forward diffusion process}, we construct a discrete-time Markov chain \(\{\boldsymbol{\theta}_t\}_{t=0}^T\), initialized at \(\boldsymbol{\theta}_0 = \boldsymbol{\theta}\). At each step \(t\), the new state \(\boldsymbol{\theta}_t\) is derived from the old state \(\boldsymbol{\theta}_{t-1}\) by perturbing it with Gaussian noise according to the transition kernel:
\begin{equation}
q(\boldsymbol{\theta}_t \mid \boldsymbol{\theta}_{t-1}) = \mathcal{N}\big( \boldsymbol{\theta}_t;\, \sqrt{1 - \beta_t} \, \boldsymbol{\theta}_{t-1},\, \beta_t \mathbf{I} \big),
\label{eq:forward_diffusion}
\end{equation}
where \(\beta_t \in (0, 1)\) determines the noise level at each step, and \(\mathbf{I}\) is the identity matrix. This step-wise transformation progressively corrupts \(\boldsymbol{\theta}_0\), moving its mean towards zero and increasing its variance to unity, such that as \(t \to T\), the state converges to \(q(\boldsymbol{\theta}_T) = \mathcal{N}(\mathbf{0}, \mathbf{I})\).

Alternatively, \(\boldsymbol{\theta}_t\) can be directly computed from \(\boldsymbol{\theta}_0\) using the marginal distribution:
\begin{equation}
q(\boldsymbol{\theta}_t \mid \boldsymbol{\theta}_0) = \mathcal{N}\big( \boldsymbol{\theta}_t;\, \sqrt{\bar{\alpha}_t} \, \boldsymbol{\theta}_0,\, (1 - \bar{\alpha}_t) \mathbf{I} \big),
\label{eq:marginal_forward_diffusion}
\end{equation}
 where $\bar{\alpha}_t = \prod_{s=1}^t (1 - \beta_s)$ is the cumulative product of the noise scaling factors. This formulation allows efficient computation of \(\boldsymbol{\theta}_t\) at any step \(t\), directly linking \(\boldsymbol{\theta}_t\) to \(\boldsymbol{\theta}_0\). The joint forward  distribution is given by:
\begin{equation}
q(\boldsymbol{\theta}_{1:T}\mid \boldsymbol{\theta}_0)
=
q(\boldsymbol{\theta}_T\mid \boldsymbol{\theta}_0)
\prod_{t=2}^T
q(\boldsymbol{\theta}_{t-1}\mid \boldsymbol{\theta}_t,\boldsymbol{\theta}_0).
\label{eq:joint_forward}
\end{equation}

In the \textbf{reverse diffusion process}, latent states $\{\boldsymbol{\theta}_t\}_{t=T}^0$ are denoised from $\boldsymbol{\theta}_T \sim \mathcal{N}(\boldsymbol{0}, \boldsymbol{I})$ to recover $\boldsymbol{\theta}_0$, with every step conditioned on observed data $\mathbf{y}$ \citep{saharia2021imagesuperresolutioniterativerefinement}. The joint reverse distribution is formulated as:
\begin{equation}
p(\boldsymbol{\theta}_{0:T}\mid \mathbf{y})
=
p(\boldsymbol{\theta}_T)\,\prod_{t=1}^T
p(\boldsymbol{\theta}_{t-1}\mid \boldsymbol{\theta}_t,\mathbf{y}).
\label{eq:joint_reverse}
\end{equation}
Training the reverse model involves maximizing the log‑posterior of the latent variables, given the observations:
\begin{equation}
\mathcal{L}_{\mathrm{ConDiSim}}
= \max \log p(\boldsymbol{\theta}_0 \mid \mathbf{y}),
\label{eq:neg_log_post}
\end{equation}
\begin{equation}
\text{where,} \quad \log p(\boldsymbol{\theta}_0 \mid \mathbf{y}) = \log \int p(\boldsymbol{\theta}_{0:T} \mid \mathbf{y}) \, d\boldsymbol{\theta}_{1:T}.
\label{eq:log_posterior}
\end{equation}
However, evaluating \(\log p(\boldsymbol{\theta}_0 \mid \mathbf{y})\)
 directly is intractable, as it entails integrating over the full trajectory of intermediate latent states. Instead, we employ a variational approach by introducing the joint forward distribution \(q(\boldsymbol{\theta}_{1:T}\mid \boldsymbol{\theta}_0)\). Using Jensen’s inequality, the lower bound on the log-posterior is:
\begin{equation}
\log p(\boldsymbol{\theta}_0 \mid \mathbf{y})
\;\ge\;
\mathbb{E}_{q(\boldsymbol{\theta}_{1:T}\mid \boldsymbol{\theta}_0)}
\!\biggl[
\log \frac{p(\boldsymbol{\theta}_{0:T}\mid \mathbf{y})}
{q(\boldsymbol{\theta}_{1:T}\mid \boldsymbol{\theta}_0)}
\biggr].
\label{eq:elbo}
\end{equation}
Applying Eqs.~\eqref{eq:joint_forward} and \eqref{eq:joint_reverse} to Eq.~\eqref{eq:elbo} leads to the following decomposition:
\begin{equation}
\mathcal{L}_{\mathrm{ConDiSim}} = \mathcal{L}_{\mathrm{recon}} + \mathcal{L}_{\mathrm{prior}} + \mathcal{L}_{\mathrm{denoise}},
\end{equation}
where, the reconstruction term, \(\mathcal{L}_{\mathrm{recon}} = \mathbb{E}_{q}[\log p(\boldsymbol{\theta}_0 \mid \boldsymbol{\theta}_1, \mathbf{y})]\), evaluates how well the original parameters are recovered from the final denoised state. The prior term, \(\mathcal{L}_{\mathrm{prior}} = -D_{\mathrm{KL}}(q(\boldsymbol{\theta}_T \mid \boldsymbol{\theta}_0) \,\|\, p(\boldsymbol{\theta}_T))\), enforces consistency with the prior at the final noised step. The denoising term, \(\mathcal{L}_{\mathrm{denoise}} = -\sum_{t=2}^T \mathbb{E}_{q}[D_{\mathrm{KL}}(q(\boldsymbol{\theta}_{t-1} \mid \boldsymbol{\theta}_t, \boldsymbol{\theta}_0) \,\|\, p(\boldsymbol{\theta}_{t-1} \mid \boldsymbol{\theta}_t, \mathbf{y}))]\), aligns reverse transitions with the forward process. For sufficiently large $T$, the reconstruction and prior terms vanish \citep{NEURIPS2020_4c5bcfec}, leaving the denoising KL as the dominant training objective (i.e., $\mathcal{L} \approx \mathcal{L}_{\text{denoise}}$):
\begin{equation}
\mathcal{L}
=
\sum_{t=2}^T
\mathbb{E}_{q}
\bigl[
D_{\mathrm{KL}}\bigl(
q(\boldsymbol{\theta}_{t-1}\mid\boldsymbol{\theta}_t,\boldsymbol{\theta}_0)
\;\|\;
p_\phi(\boldsymbol{\theta}_{t-1}\mid\boldsymbol{\theta}_t,\mathbf{y})
\bigr)
\bigr].
\label{eq:loss_kl}
\end{equation}
The first term in the KL divergence, \(q(\boldsymbol{\theta}_{t-1} \mid \boldsymbol{\theta}_t, \boldsymbol{\theta}_0)\), can be written using Bayes’ theorem as \(q(\boldsymbol{\theta}_{t-1} \mid \boldsymbol{\theta}_t, \boldsymbol{\theta}_0) = \frac{q(\boldsymbol{\theta}_t \mid \boldsymbol{\theta}_{t-1}, \boldsymbol{\theta}_0) \, q(\boldsymbol{\theta}_{t-1} \mid \boldsymbol{\theta}_0)}{q(\boldsymbol{\theta}_t \mid \boldsymbol{\theta}_0)}\). From Eq.~\eqref{eq:forward_diffusion} and Eq.~\eqref{eq:marginal_forward_diffusion}, each term is Gaussian, so the resulting distribution is also Gaussian: \(q(\boldsymbol{\theta}_{t-1} \mid \boldsymbol{\theta}_t, \boldsymbol{\theta}_0) = \mathcal{N}(\boldsymbol{\theta}_{t-1}; \boldsymbol{\mu}_q, \boldsymbol{\Sigma}_q)\), where:
\begin{equation}
\begin{aligned}
\boldsymbol{\mu}_q
&=
\frac{\sqrt{\bar{\alpha}_{t-1}}(1-\alpha_t)}{1-\bar{\alpha}_t}\,\boldsymbol{\theta}_0
+
\frac{\sqrt{\alpha_t}(1-\bar{\alpha}_{t-1})}{1-\bar{\alpha}_t}\,\boldsymbol{\theta}_t,\\
\boldsymbol{\Sigma}_q
&=
\frac{(1-\bar{\alpha}_{t-1})\alpha_t}{1-\bar{\alpha}_t}\,\mathbf{I}.
\end{aligned}
\label{eq:forward_posterior}
\end{equation}
To eliminate the explicit dependence on $\boldsymbol{\theta}_0$, we substitute $\boldsymbol{\theta}_0 = (\boldsymbol{\theta}_t - \sqrt{1 - \bar{\alpha}_t}\,\boldsymbol{\epsilon}) / \sqrt{\bar{\alpha}_t}$ (with $\boldsymbol{\epsilon}\sim\mathcal{N}(\mathbf{0},\mathbf{I})$) into Eq.~\eqref{eq:marginal_forward_diffusion}, reparameterizing $\boldsymbol{\mu}_q$ to yield a noise-parameterized expression for the mean:
\begin{equation}
\begin{aligned}
\boldsymbol{\mu}_q(\boldsymbol{\theta}_t,t)
&=
\tfrac{1}{\sqrt{\alpha_t}}
\Bigl(\boldsymbol{\theta}_t - \tfrac{\beta_t}{\sqrt{1-\bar{\alpha}_t}}\,\boldsymbol{\epsilon}\Bigr).
\end{aligned}
\label{eq:mean_param}
\end{equation}
Assuming the learned reverse process preserves the covariance structure ($\boldsymbol{\Sigma}_p = \boldsymbol{\Sigma}_q$) \citep{NEURIPS2020_4c5bcfec}, the KL divergence in Eq.~\eqref{eq:loss_kl} reduces to the squared Euclidean distance between the means:
\begin{equation}
\mathcal{L}
=
\sum_{t=2}^T
\mathbb{E}_{q(\boldsymbol{\theta}_{1:T}\mid\boldsymbol{\theta}_0)}
\Bigl[
\big\lVert \boldsymbol{\mu}_q(\boldsymbol{\theta}_t,\boldsymbol{\theta}_0) \;-\; \boldsymbol{\mu}_p(\boldsymbol{\theta}_t,\mathbf{y},t)\big\rVert^2
\Bigr].
\label{eq:loss_mu}
\end{equation}
From Eq.~\eqref{eq:mean_param}, the mean of learned reverse process can be expressed as:
\begin{equation}
\begin{aligned}
\boldsymbol{\mu}_p(\boldsymbol{\theta}_t,\mathbf{y},t)
&=
\tfrac{1}{\sqrt{\alpha_t}}
\Bigl(\boldsymbol{\theta}_t - \tfrac{\beta_t}{\sqrt{1-\bar{\alpha}_t}}\,
\boldsymbol{\epsilon}_\phi(\boldsymbol{\theta}_t,\mathbf{y},t)\Bigr),
\end{aligned}
\label{eq:mean_reparam_reverse}
\end{equation}
where, $\boldsymbol{\epsilon}_\phi(\boldsymbol{\theta}_t,\mathbf{y},t)$ is a neural network that, given the noisy latent $\boldsymbol{\theta}_t$, conditioning $\mathbf{y}$, and time step $t$, predicts the noise. Substituting the mean parameterizations from Eqs.\eqref{eq:mean_param} and \eqref{eq:mean_reparam_reverse} into the loss of Eq.\eqref{eq:loss_mu} and discarding all constant factors, yields the compact noise prediction objective:
\begin{equation}
\mathcal{L}_{\text{\tiny ConDiSim}}
\approx
\mathbb{E}_{\,t,\,\boldsymbol{\epsilon},\,(\boldsymbol{\theta},\mathbf y)\sim p(\boldsymbol{\theta},\mathbf{y})}
\Bigl[\big\lVert \boldsymbol{\epsilon} - \boldsymbol{\epsilon}_\phi(\boldsymbol{\theta}_t,\mathbf{y},t)\big\rVert^2\Bigr].
\label{eq:loss_epsilon}
\end{equation}

\begin{figure*}[t]
    \centering
    \includegraphics[width=\textwidth]{new.pdf}
    \caption{ConDiSim Architecture: Each diffusion block is FiLM-modulated by timestep and observation embeddings to estimate the noise injected at step $t$  during forward diffusion.}
    \label{fig:diffusion_architecture}
\end{figure*}

\subsection{Conditioning and Sampling} \label{condition}
The objective in Eq.~(\ref{eq:loss_epsilon}) requires a denoising network that predicts the noise 
$\boldsymbol{\epsilon}$ from the noisy state $\boldsymbol{\theta}_t$, conditioned on the observation 
$\mathbf{y}$ and the diffusion timestep $t$. Preserving this dual conditioning throughout the denoising trajectory is essential to align with \(p(\boldsymbol{\theta}\mid\mathbf y)\). A naive scheme that feeds \([\boldsymbol{\theta}_t;\,\mathbf y;\,t]\) into a standard network often performs poorly in SBI, where \(\dim(\mathbf y)\!\gg\!\dim(\boldsymbol{\theta})\), input-level concatenation forces the network to disentangle \(\mathbf y\) and \(t\) from \(\boldsymbol{\theta}_t\) in a single step, yielding an ill-conditioned coupling that dilutes the posterior signal and destabilizes training. To address this, the proposed ConDiSim (Fig.~\ref{fig:diffusion_architecture}) model takes \((\boldsymbol{\theta}_t, t, \mathbf y)\) as inputs; while a condition encoder \(f_y(\mathbf y)\) and a timestep encoder \(f_t(t)\) produce embeddings that are concatenated into a context vector \(\mathbf c=[\,f_y(\mathbf y);\;f_t(t)\,]\). A lightweight MLP maps \(\mathbf c\) to per-block FiLM parameters \((\boldsymbol{\gamma}_\ell(\mathbf c),\boldsymbol{\beta}_\ell(\mathbf c))\), which modulate each diffusion block via \(\operatorname{FiLM}_\ell(\mathbf h_\ell)=\boldsymbol{\gamma}_\ell(\mathbf c)\odot\mathbf h_\ell+\boldsymbol{\beta}_\ell(\mathbf c)\) \citep{10.5555/3504035.3504518}. This ensures that hidden features and thus the noise predictor \(\boldsymbol{\epsilon}_\phi(\boldsymbol{\theta}_t,\mathbf y,t)\) remain observation-aware and time-aware at every step, avoiding the signal dilution of naive concatenation. 

Given this architecture, to perform amortized inference we draw \(N\) initial states \(\boldsymbol{\theta}_T^{(i)}\sim\mathcal N(\mathbf 0,\mathbf I)\) and evolve them backward for \(t=T,\ldots,1\) :
\begin{equation}\label{eq:reverse_update}
\boldsymbol{\theta}_{t-1}^{(i)}=\frac{1}{\sqrt{\alpha_t}}\!\left(\boldsymbol{\theta}_t^{(i)}-\frac{1-\alpha_t}{\sqrt{1-\bar{\alpha}_t}}\,\hat{\boldsymbol{\epsilon}}_{\phi}(\boldsymbol{\theta}_t^{(i)},\mathbf y,t)\right)+\sigma_t\,\mathbf z_t,
\end{equation}
where \(\mathbf z_t\sim\mathcal N(\mathbf 0,\mathbf I)\) and \(\sigma_t=\sqrt{\beta_t}\). To balance fidelity to \(\mathbf y\) with sample diversity, we apply classifier-free guidance by interpolating conditional and unconditional predictions,
\(
\hat{\boldsymbol{\epsilon}}_\phi=(1+\lambda)\,\boldsymbol{\epsilon}_\phi(\boldsymbol{\theta}_t,\mathbf y,t)-\lambda\,\boldsymbol{\epsilon}_\phi(\boldsymbol{\theta}_t,t),
\)
with guidance scale \(\lambda\ge 0\). The final set \(\{\boldsymbol{\theta}_0^{(i)}\}_{i=1}^N\) provides approximate samples from \(p(\boldsymbol{\theta}\mid\mathbf y)\) \citep{ho2021classifierfree}.

\section{Experiments}\label{Experiments}
We evaluate ConDiSim on 10 benchmark problems from the \texttt{sbibm} suite~\citep{lueckmann2021benchmarking}, and two real‑world problems. 
For each benchmark, we constrain the training to simulation budgets of  $10{,}000$, $20{,}000$, and $30{,}000$ simulations, and infer the posterior by drawing $10{,}000$ samples from the learned model. We assess posterior quality using Maximum Mean Discrepancy (MMD), the Classifier Two-Sample Test (C2ST) \citep{Friedman03}, and Empirical Cumulative Distribution Function (ECDF) diagnostics from the \texttt{BayesFlow} framework \citep{bayesflow_2023_software}. C2ST results are reported in the main text; MMD scores and full ECDF plots for all benchmarks are provided in Appendix~\ref{sbc_supp}. Complete hyperparameter configurations for all methods are detailed in Appendix~\ref{hyperparams}.

\begin{figure}[!h]
    \centering
    \subcaptionbox{Posterior Plot\label{fig:two_moons}}{%
        \includegraphics[width=0.45\textwidth]{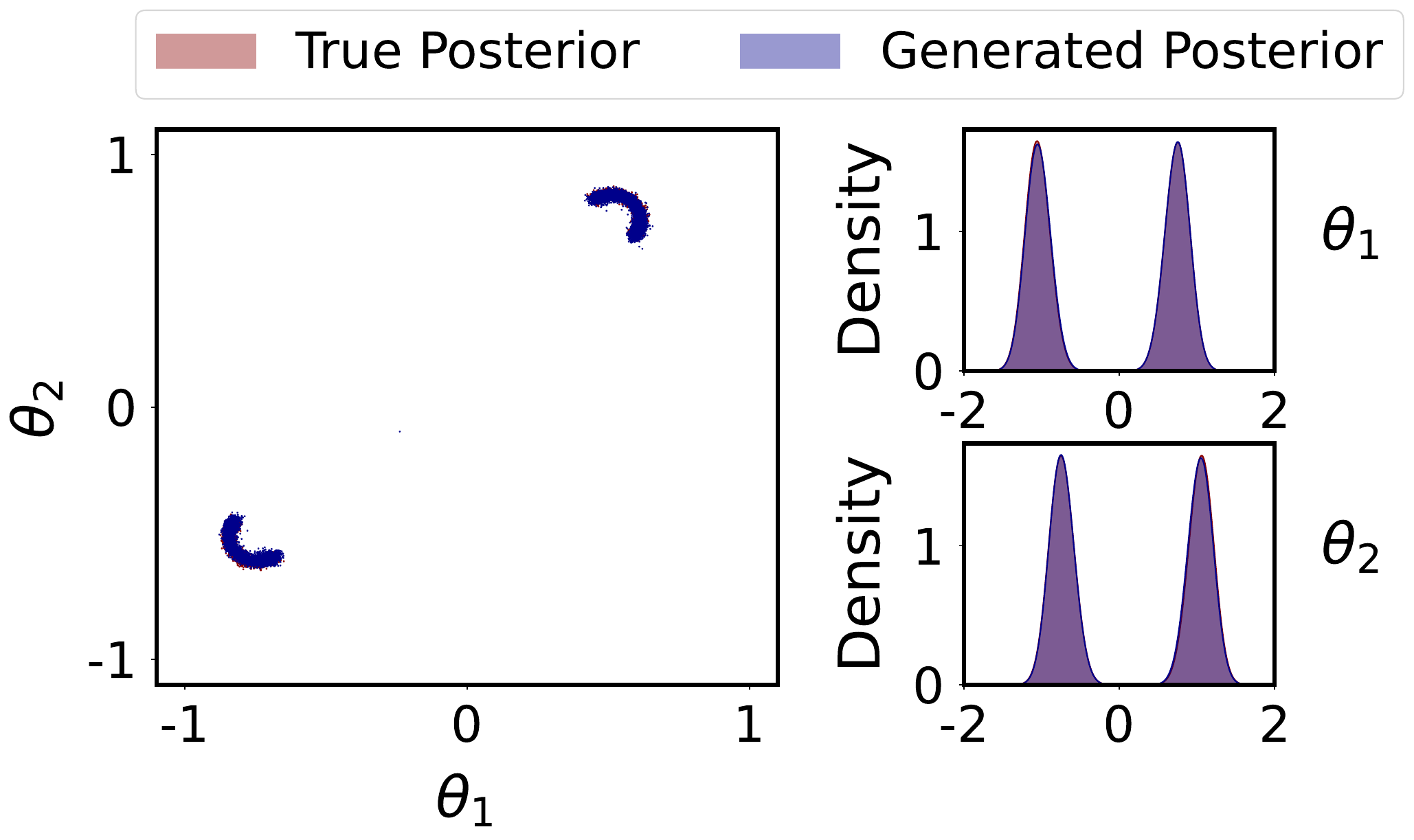}%
    }\hspace{0.5cm}
    \subcaptionbox{ECDF Plot\label{fig:two_moons_ecdf}}{%
        \includegraphics[width=0.45\textwidth]{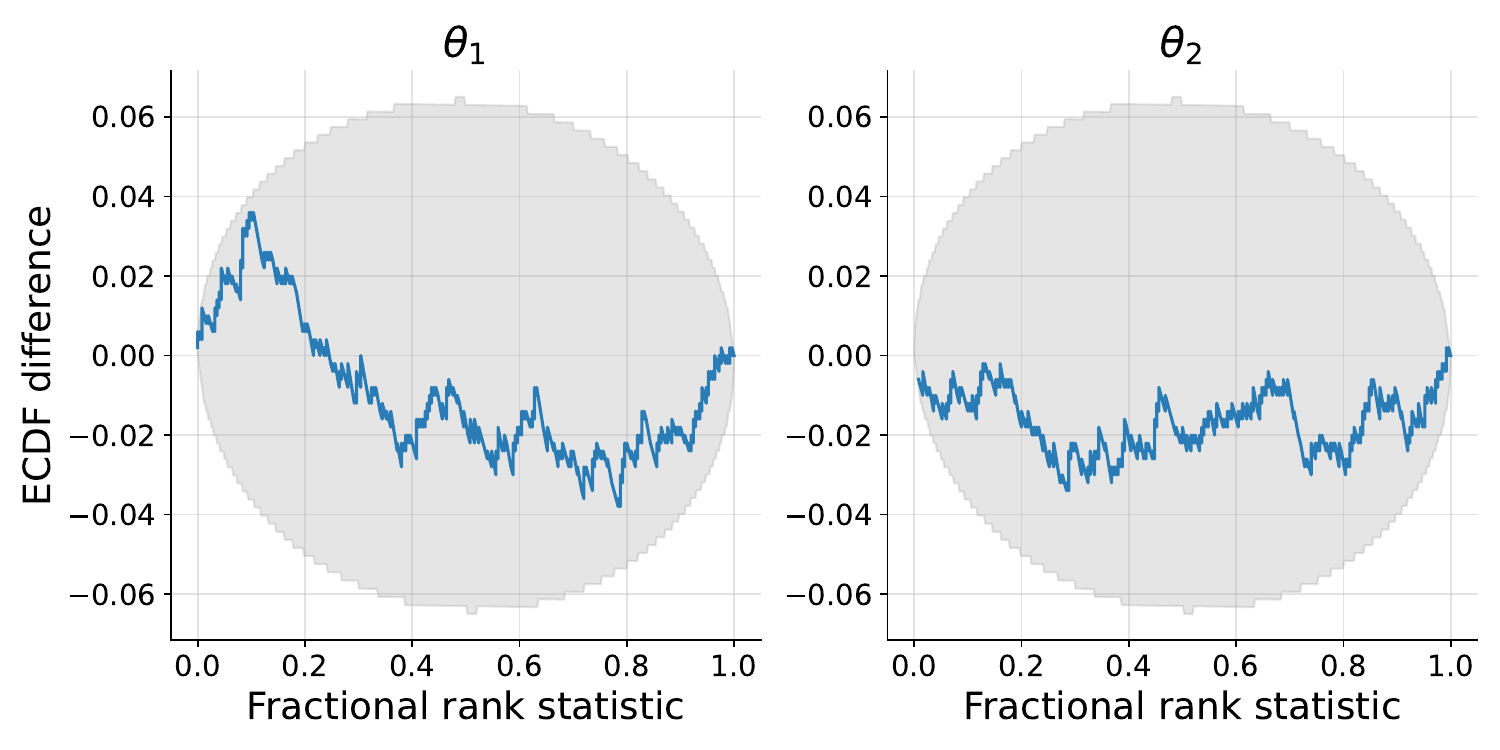}%
    }
    \caption{\textbf{Two Moons:} Comparison of generated and reference posterior distributions, with the ECDF plot representing the cumulative distribution of fractional rank statistics from posterior draws.}
    \label{fig:models}
\end{figure}

\begin{figure}[!h]
    \centering
    \subcaptionbox{SLCP.\label{fig:slcp_posterior}}{%
        \includegraphics[width=0.48\linewidth]{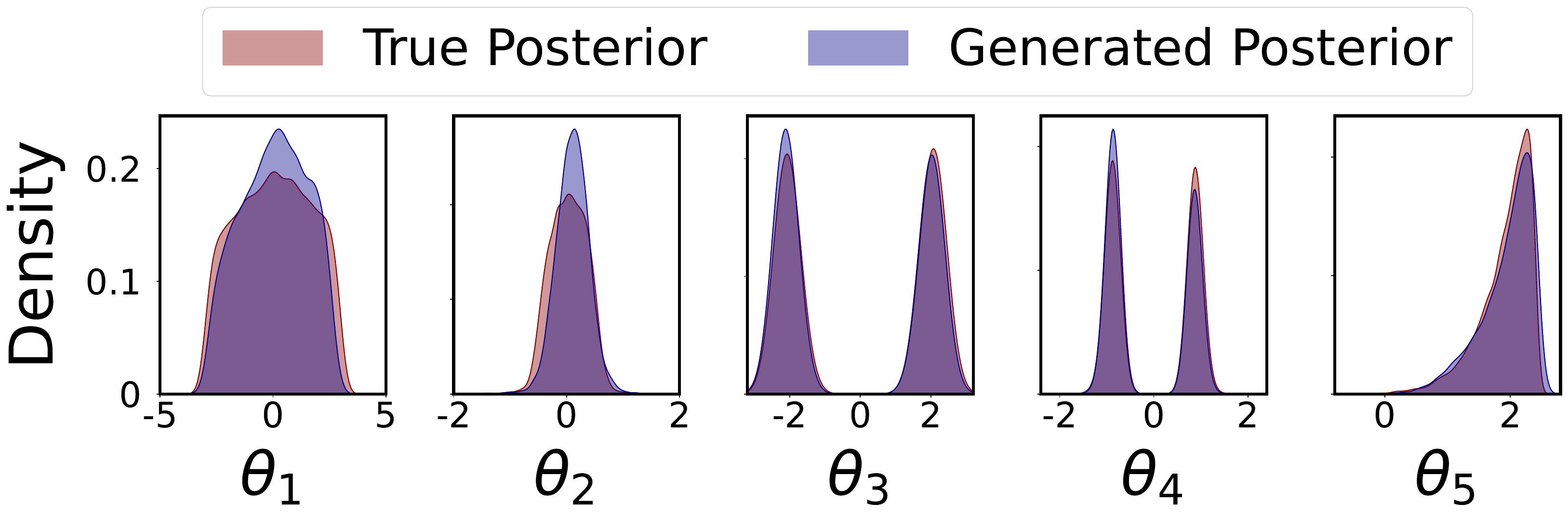}%
    }\hfill
    \subcaptionbox{SLCP Distractors.\label{fig:slcp_dist_posterior}}{%
        \includegraphics[width=0.48\linewidth]{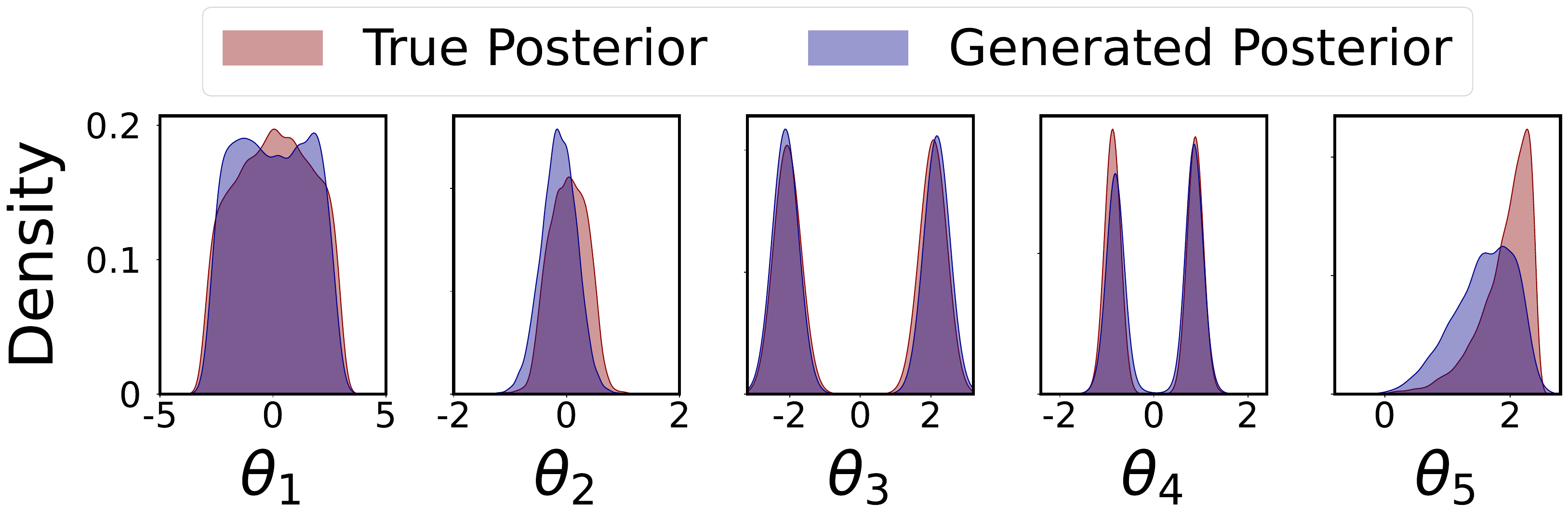}%
    }
    \caption{\textbf{SLCP:} Comparison of generated vs. reference posteriors for SLCP and SLCP Distractors.}
    \label{fig:posteriors}
\end{figure}

\begin{figure*}[!t]
    \centering
    \begin{subfigure}[t]{0.45\linewidth}
        \centering
        \includegraphics[width=\linewidth]{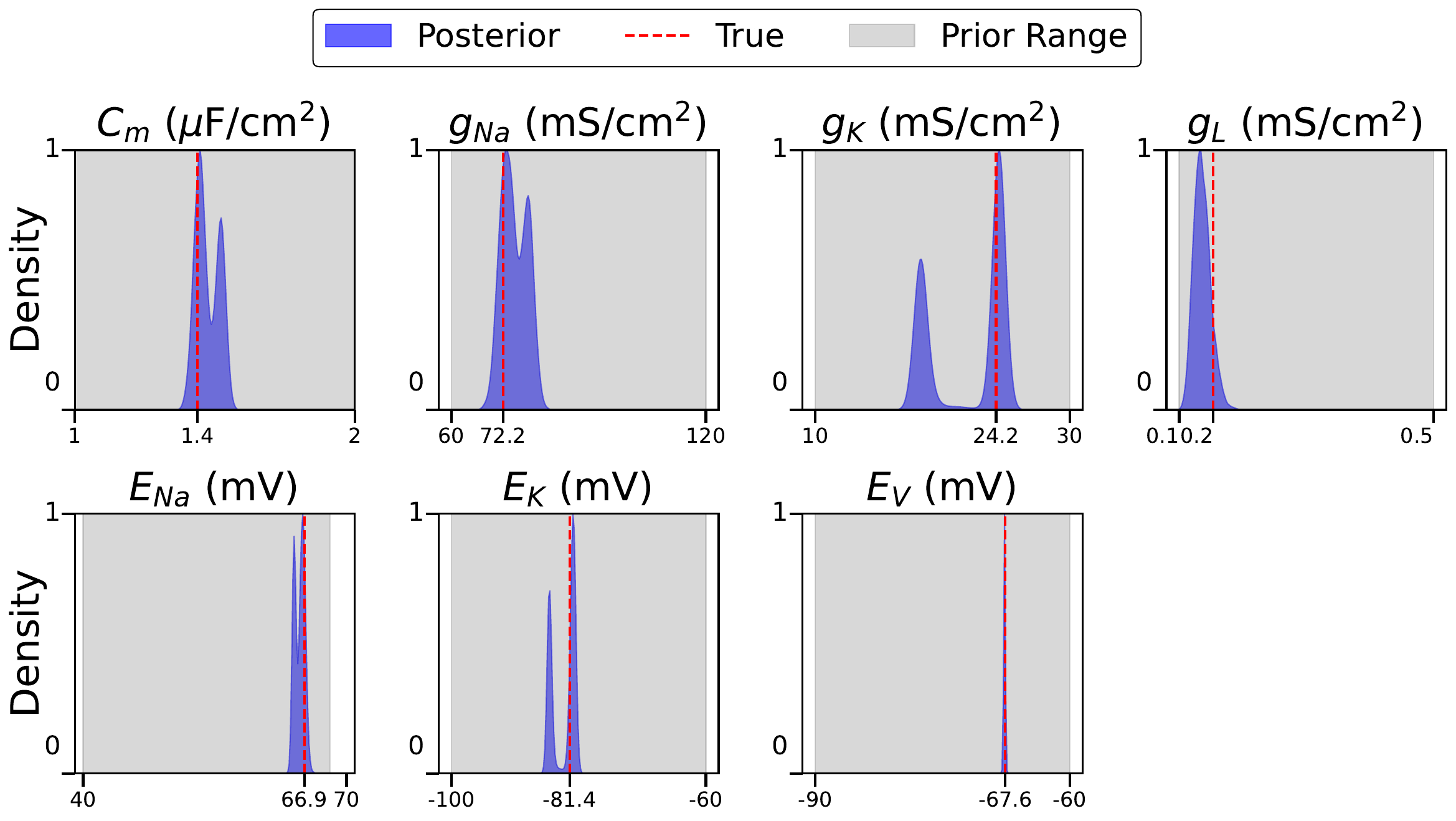}
        \caption{Hodgkin–Huxley: Approximated Posterior}
        \label{fig:hh_posterior}
    \end{subfigure}%
    \hspace{4mm}
    \begin{subfigure}[t]{0.45\linewidth}
        \centering
        \includegraphics[width=\linewidth]{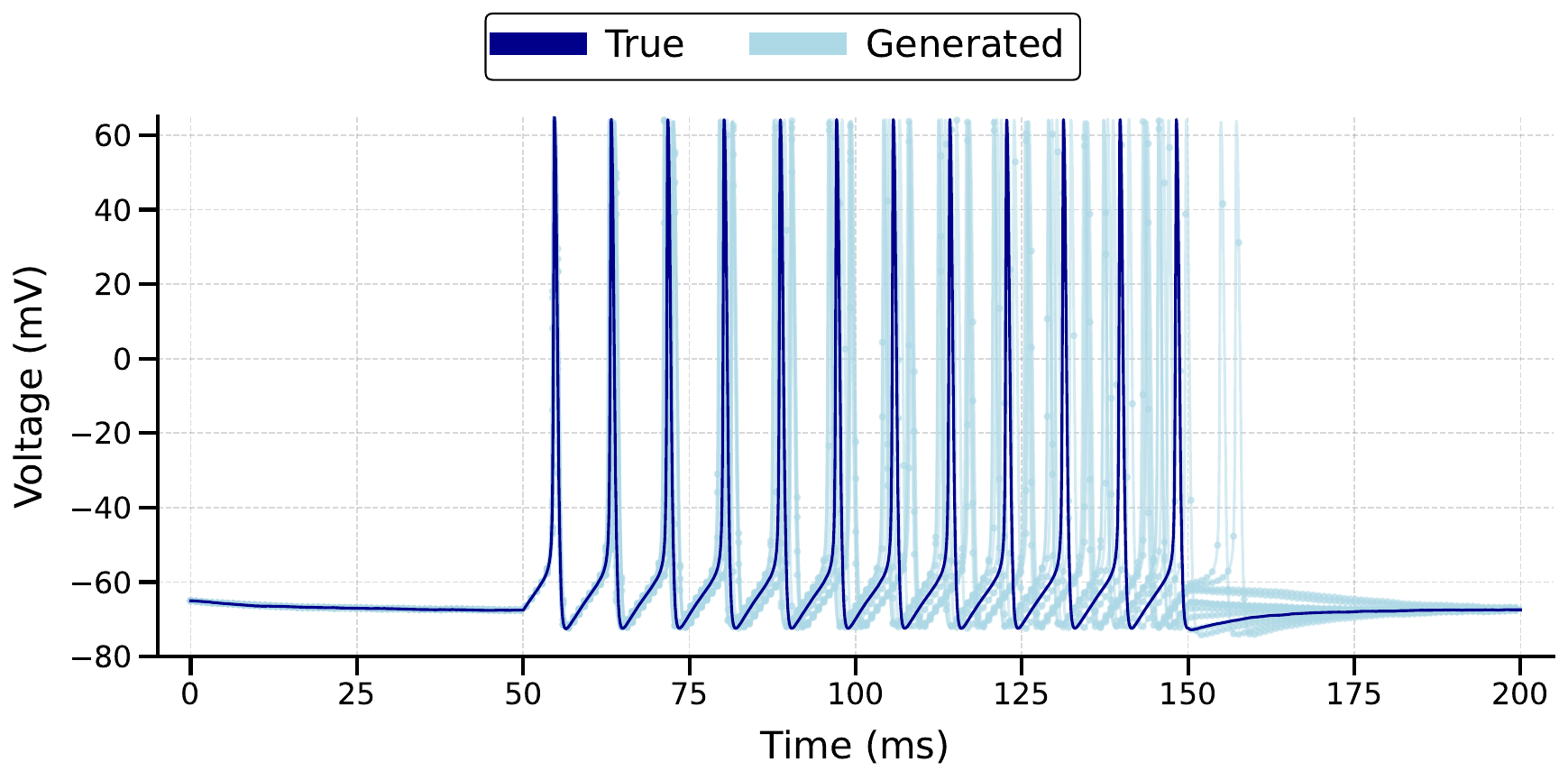}
        \caption{Posterior Predictive Voltage Traces}
        \label{fig:hh_voltage}
    \end{subfigure}
    \caption{Posterior Distributions for Hodgkin–Huxley model.}
    \label{fig:hh_posteriors}
\end{figure*}

\begin{figure*}[!t]
    \centering
    \begin{subfigure}[t]{0.5\linewidth}
        \centering
        \includegraphics[width=\linewidth]{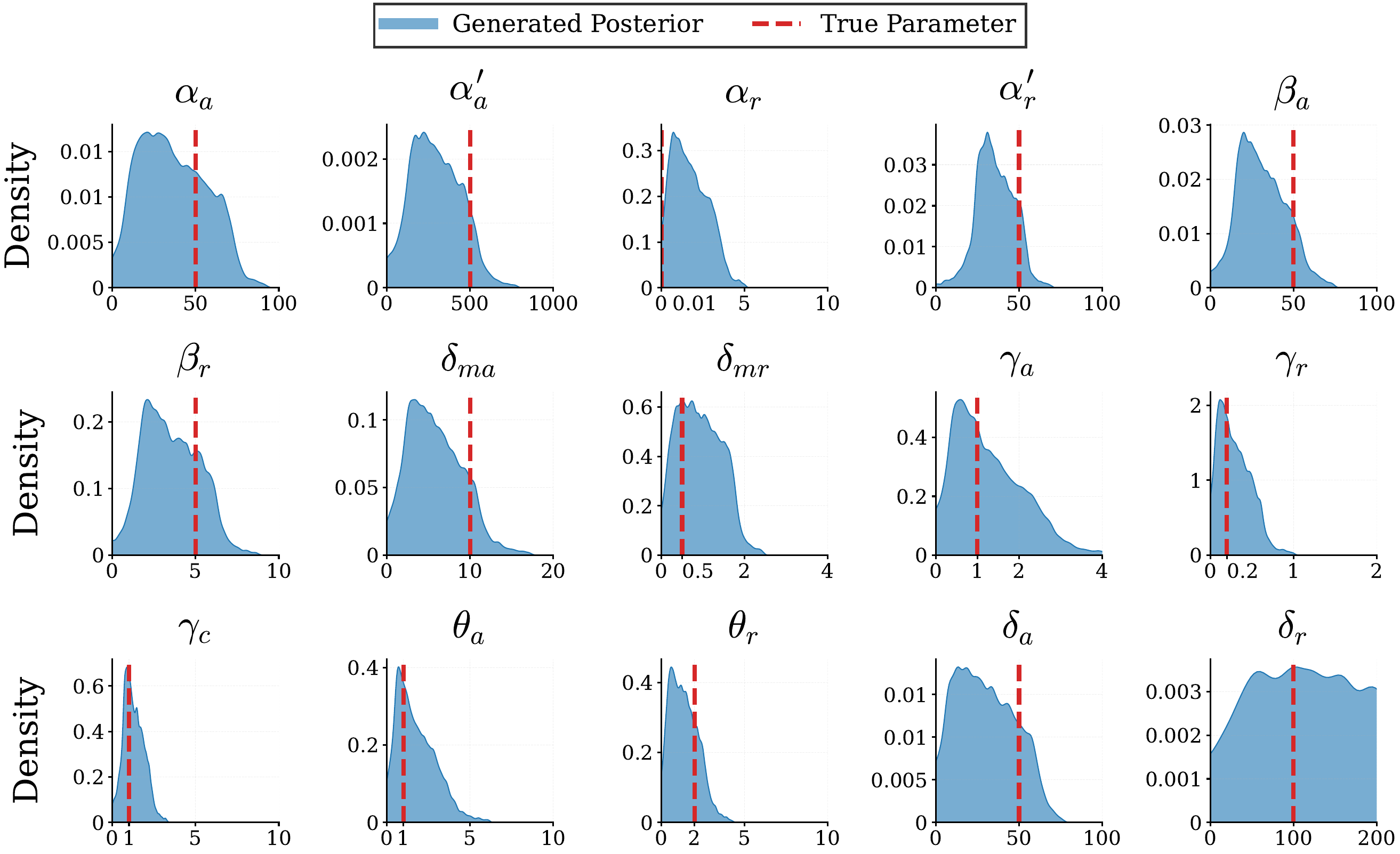}
        \caption{Genetic Oscillator: Approximated Posterior}
        \label{fig:vilar_posterior}
    \end{subfigure}%
    \hspace{6mm}
    \begin{subfigure}[t]{0.3\linewidth}
        \centering
        \includegraphics[width=\linewidth]{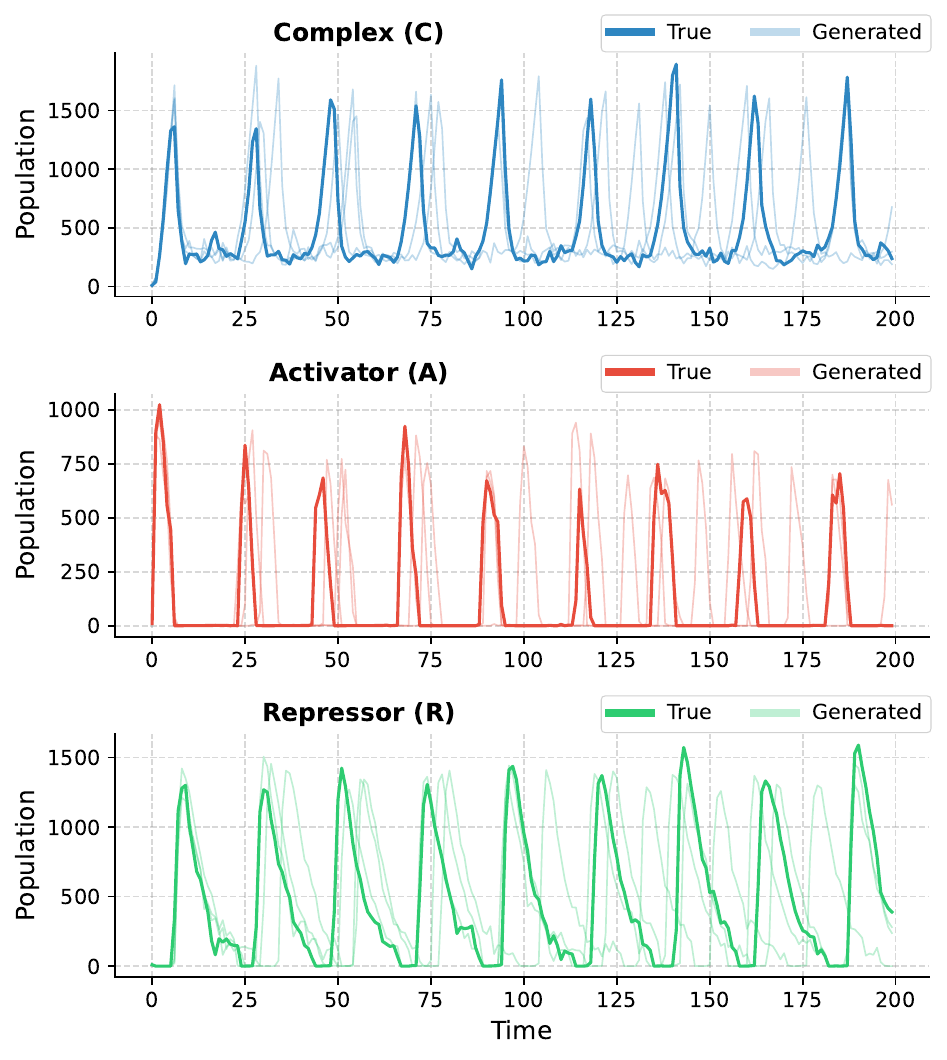}
        \caption{Posterior Predictive Trajectories.}
        \label{fig:vilar_traj}
    \end{subfigure}
    \caption{Posterior Distributions for Genetic Oscillator model.}
    \label{fig:vilar_posteriors}
\end{figure*}

\subsection{Benchmark Posterior Analysis}
\label{posterior_plots_main}
In this section, we analyze three representative benchmarks: \textbf{Two Moons}, \textbf{SLCP}, and \textbf{SLCP Distractors}, selected for their distinct and complementary challenges in posterior inference. The Two Moons benchmark poses the challenge of capturing a bi-modal posterior with a distinctive crescent geometry. As depicted in Figure~\ref{fig:two_moons}, ConDiSim accurately recovers both modes and the complex posterior structure. The ECDF plots for \(\theta_1\) and \(\theta_2\) (Figure~\ref{fig:two_moons_ecdf}) demonstrate that the inferred distributions are well-calibrated, with the majority of rank statistics falling within the 90\% confidence bands; a minor deviation for \(\theta_1\) suggests a slight degree of overconfidence.

The SLCP benchmark encompasses a range of parameter behaviors including uniform, Gaussian, multi-modal, and skewed distributions. Its variant, SLCP Distractors, further tests the model's robustness by incorporating uninformative noise dimensions, thereby simulating scenarios with heavily corrupted data. Figures~\ref{fig:slcp_posterior} and~\ref{fig:slcp_dist_posterior} illustrate that ConDiSim is able to capture these complex posterior landscapes with high fidelity, and also effectively isolate the informative parameters from the distractors. The rank ECDF plots for both SLCP variants (Figures~\ref{fig:slcp_ecdf} and~\ref{fig:slcp_dist_ecdf} in the Appendix) also remain centered within the confidence intervals, confirming that the model reliably captures parameter variability even under noisy conditions. Detailed descriptions of these benchmarks, are provided in Appendix~\ref{sbi_problems}.

Table~\ref{tab:c2st_main} reports C2ST scores across the 10 benchmarks. On tasks such as Two Moons, ConDiSim achieves performance comparable to Simformer, APT, and NPE, while requiring substantially shorter training times (Table~\ref{tab:train_times_all}). Although ConDiSim does not always attain the top C2ST score, the method offers a favorable trade-off between efficiency and posterior quality, which is crucial under limited simulation budgets. Notably, the SLCP task remains challenging for all methods, highlighting the inherent difficulty of accurately recovering high-dimensional, multi-modal posteriors. Moreover, C2ST can be overly sensitive to small errors in individual dimensions, making it less reliable in high-dimensional settings. For example, as shown in Figure~\ref{fig:slcp_dist_posterior}, ConDiSim closely matches the true posterior in most dimensions, with deviations concentrated in the final parameter. To provide additional insight, we report the MMD metric for each benchmark problem in the Supplementary Table~\ref{tab:mmd_global}.

\subsection{Application to Real-World Scenarios}
We evaluate the performance of ConDiSim on two real-world simulation models. First, the \textbf{Hodgkin-Huxley (HH)} model \citep{gloecklerall} which simulates neuronal action potentials by modeling the membrane voltage dynamics governed by sodium, potassium, and leak currents. Second, we consider a \textbf{Genetic Oscillator} model which is a relatively high-dimensional gene regulatory network  that exhibits oscillatory behavior, driven by a positive-negative feedback loop mimicking a circadian clock \citep{doi:10.1073/pnas.092133899}. The model encompasses 9 species undergoing 18 reactions, parameterized by 15 reaction rate constants.  Although the full system involves multiple interacting components, we focus on the inference of only three species (C, A, R).
Detailed description is available in Appendix~\ref{sbi_problems}.

Unlike the SBI benchmarking problems, the true posterior distributions for these real-world models are unknown. To assess ConDiSim, we use well motivated parameter combinations from literature to generate corresponding synthetic observations (e.g., parameter combination where the genetic oscillator results in stable oscillations). Performing inference based on these observations enables us to evaluate the inferred posterior by comparing its concentration around the true parameter value. We also perform a posterior predictive check to validate the model. In this procedure, five posterior samples are randomly selected and used to simulate corresponding observations, which are then compared with the observed data. This check provides an additional measure of the consistency and reliability of the inferred posteriors.

For the HH model, Figure~\ref{fig:hh_posterior} shows that ConDiSim effectively infers a posterior distribution concentrated around the true parameter values, while Figure~\ref{fig:hh_voltage} confirms that the simulated voltage trajectories closely match the reference signals.
The parameter estimation task for the genetic oscillator is more challenging due to broad and varied parameter priors. Figure~\ref{fig:vilar_posterior} indicates that the peak of the posterior distribution is near the true parameter values. Figure~\ref{fig:vilar_traj} illustrates both the true and generated trajectories for three selected species, obtained by randomly sampling three posterior sets and plotting the corresponding simulations. The trajectories align well with the true data, with slight deviations emerging over time due to the inherent stochastic nature of the model, as expected. Figure~\ref{fig:vilar_stochastic} (Appendix) depicts multiple trajectories simulated using the true parameter combination - which closely match the inferred results. In both cases, the trajectories are very close to each other during initial time steps, and develop small deviations owing to model stochasticity as time steps evolve. The plots highlight the system's inherent variability, and also validate the inferred solution. From Figs.\ref{fig:vilar_posterior}, \ref{fig:vilar_posterior_simformer}, we also see that ConDiSim is able to infer a more accurate, tighter posterior as compared to Simformer, highlighting its suitability for challenging inference problems involving stochastic, discrete-time observations.

\section{Discussion}\label{Discussion}
In this section, we compare ConDiSim against established SBI methods (Section~\ref{compare}) and analyze its behavior along three practical axes: scalability with problem dimensionality (Section~\ref{scale}), robustness to simulator noise or misspecification (Section~\ref{robust}), and reliability in capturing posterior uncertainty (Section~\ref{reliable}). This analysis provides practical guidance for deploying diffusion-based inference methods by identifying regimes where ConDiSim excels and scenarios where alternative approaches may be more suitable.

\subsection{ConDiSim vs. Other SBI Methods}\label{compare}
Tables~\ref{tab:c2st_main} and~\ref{tab:mmd_global} show that diffusion-based methods (Simformer, ConDiSim) often achieve higher posterior accuracy than flow-based (NPE, APT, FMPE) and GAN-based (GATSBI) approaches, particularly on problems with multimodal or highly nonlinear posteriors. Simformer attains the best overall performance on several benchmarks (Table~\ref{tab:mmd_global}), owing to its transformer architecture with cross-attention conditioning; however, this comes at the cost of substantially longer training times (Table~\ref{tab:train_times_all}). In contrast, ConDiSim uses a lightweight FiLM-based conditioning mechanism that avoids costly attention computations, achieving competitive accuracy with significantly lower training overhead - while its iterative denoising incurs no practical slowdown in inference. NPE and APT remain strong baselines on simpler or lower-dimensional tasks, where their one-shot or sequential flow-based strategies are sufficient. GATSBI, meanwhile, is hampered by the well-known instabilities of adversarial training. We also note from Table~\ref{tab:c2st_main} that ConDiSim outperforms other methods on problems involving noisy, discrete observations, such as Bernoulli GLM, Bernoulli GLM Raw and SIR, leveraging FiLM. This makes the proposed approach well-suited for SBI problems in computational biology where parameter inference of stochastic models with time-series observations is often encountered (e.g., the genetic oscillator test problem, as noted before).
\newpage
\begin{table*}[!t]
\centering
\renewcommand{\arraystretch}{1.25}  
\setlength{\tabcolsep}{5pt} 
\caption{C2ST scores across SBI benchmarks (mean $\pm$ std.) at varying simulation budgets; best in \textbf{bold}, second-best \underline{underlined}.}
\label{tab:c2st_main}
\begin{subtable}[t]{\textwidth}
\centering

\begin{adjustbox}{max width=\textwidth}  
\begin{tabular}{>{\columncolor{gray!10}}l|c|c|c|c|c|c|c|c|c}
\hline
\rowcolor{gray!10}
\textbf{Method} 
  & \multicolumn{3}{c|}{\textbf{SLCP Distractors}} 
  & \multicolumn{3}{c|}{\textbf{Bernoulli GLM Raw}} 
  & \multicolumn{3}{c}{\textbf{Lotka Volterra}} \\
\hline
\rowcolor{gray!10}
\textbf{Budget}
  & \textbf{10,000} & \textbf{20,000} & \textbf{30,000}
  & \textbf{10,000} & \textbf{20,000} & \textbf{30,000}
  & \textbf{10,000} & \textbf{20,000} & \textbf{30,000} \\
\hline

NPE     
& 0.9714 \tiny{$\pm$0.0144}  
& 0.9468 \tiny{$\pm$0.0157}  
& 0.9359 \tiny{$\pm$0.0127}  
& 0.9354 \tiny{$\pm$0.0093}  
& 0.9114 \tiny{$\pm$0.0082}  
& 0.9065 \tiny{$\pm$0.0105}  
& 0.9999 \tiny{$\pm$0.0001}  
& 0.9999 \tiny{$\pm$0.0000}  
& 0.9999 \tiny{$\pm$0.0000}  
\\

APT     
& 0.9823 \tiny{$\pm$0.0055}  
& 0.9451 \tiny{$\pm$0.0158}  
& 0.9130 \tiny{$\pm$0.0301}  
& \underline{0.8455} \tiny{$\pm$0.0116}  
& 0.8596 \tiny{$\pm$0.0152}  
& 0.8461 \tiny{$\pm$0.0135}  
& 0.9999 \tiny{$\pm$0.0001}  
& 0.9999 \tiny{$\pm$0.0000}  
& 0.9999 \tiny{$\pm$0.0001}  
\\

GATSBI
& 0.9995 \tiny{$\pm$0.0007}  
& 0.9978 \tiny{$\pm$0.0016}  
& 0.9994 \tiny{$\pm$0.0004}  
& 0.9998 \tiny{$\pm$0.0001}  
& 0.9999 \tiny{$\pm$0.0001}  
& 0.9999 \tiny{$\pm$0.0000}  
& 0.9999 \tiny{$\pm$0.0001}  
& 0.9999 \tiny{$\pm$0.0000}  
& 1.0 \tiny{$\pm$0.0000}
\\

Simformer     
& \textbf{0.9198} \tiny{$\pm$0.0451}  
& \textbf{0.8814} \tiny{$\pm$0.0433}  
& \textbf{0.8717} \tiny{$\pm$0.0453}  
& 0.9814 \tiny{$\pm$0.0106}  
& 0.8600 \tiny{$\pm$0.0452}  
& 0.7750 \tiny{$\pm$0.0473}  
& 0.9999 \tiny{$\pm$0.0003}  
& 0.9999 \tiny{$\pm$0.0012}  
& 1.0000 \tiny{$\pm$0.0002}

\\

FMPE
& 0.9713 \tiny{$\pm$ 0.0117}  
& 0.9779 \tiny{$\pm$ 0.0095}  
& 0.9387 \tiny{$\pm$ 0.0377}  
& 0.9742 \tiny{$\pm$ 0.0113}  
& \underline{0.6661} \tiny{$\pm$ 0.0307}  
& \underline{0.6653} \tiny{$\pm$ 0.0377}  
& 0.9999 \tiny{$\pm$ 0.0178}  
& 0.9999 \tiny{$\pm$ 0.0345}  
& 0.9999 \tiny{$\pm$ 0.0018}

\\

\rowcolor{blue!5}

ConDiSim 
& \underline{0.9456} \tiny{$\pm$0.0245}  
& \underline{0.9222} \tiny{$\pm$0.02609}  
& \underline{0.8832} \tiny{$\pm$0.08519}  
& \textbf{0.6673} \tiny{$\pm$0.0668}  
& \textbf{0.6099} \tiny{$\pm$0.0272}  
& \textbf{0.6041} \tiny{$\pm$0.0237}  
& 0.9995 \tiny{$\pm$0.0001}  
& 0.9995 \tiny{$\pm$0.0005}  
& 0.9995 \tiny{$\pm$0.0003}    
\\
\hline
\end{tabular}
\end{adjustbox}
\end{subtable}

\vspace{0.1cm}


\begin{subtable}{\textwidth}
\centering
\renewcommand{\arraystretch}{1.25}  
\setlength{\tabcolsep}{2pt} 
\begin{adjustbox}{max width=\textwidth} 
\begin{tabular}{>{\columncolor{gray!10}}l|ccc|ccc|ccc|ccc}
\hline
\rowcolor{gray!10}
\textbf{Method}
  & \multicolumn{3}{c|}{\textbf{Gaussian Mixture}}
  & \multicolumn{3}{c|}{\textbf{Gaussian Linear}}
  & \multicolumn{3}{c|}{\textbf{Gaussian Linear Uniform}}
  & \multicolumn{3}{c}{\textbf{SLCP}} \\
\hline
\rowcolor{gray!10}
\textbf{Budget}
  & \textbf{10,000} & \textbf{20,000} & \textbf{30,000}
  & \textbf{10,000} & \textbf{20,000} & \textbf{30,000}
  & \textbf{10,000} & \textbf{20,000} & \textbf{30,000}
  & \textbf{10,000} & \textbf{20,000} & \textbf{30,000} \\
\hline

NPE     
& 0.6118 \tiny{$\pm$0.0155}  
& 0.5830 \tiny{$\pm$0.0240}  
& 0.5818 \tiny{$\pm$0.0158}  
& \underline{0.5472} \tiny{$\pm$0.0076}  
& \textbf{0.5238} \tiny{$\pm$0.0112}  
& \underline{0.5266} \tiny{$\pm$0.0037}  
& \underline{0.5562} \tiny{$\pm$0.0147}  
& \textbf{0.5292} \tiny{$\pm$0.0068}  
& \textbf{0.5256} \tiny{$\pm$0.0066}  
& 0.8996 \tiny{$\pm$0.0206}  
& 0.8502 \tiny{$\pm$0.0174}  
& 0.8546 \tiny{$\pm$0.0213}  
\\

APT     
& 0.5457 \tiny{$\pm$0.0058}  
& \underline{0.5425} \tiny{$\pm$0.0124}  
& \underline{0.5467} \tiny{$\pm$0.0050}  
& \textbf{0.5367} \tiny{$\pm$0.0038}  
& \underline{0.5293} \tiny{$\pm$0.0052}  
& \textbf{0.5258} \tiny{$\pm$0.0047}  
& \textbf{0.5401} \tiny{$\pm$0.0115}  
& \underline{0.5302} \tiny{$\pm$0.0069}  
& \underline{0.5264} \tiny{$\pm$0.0075}  
& \underline{0.8164} \tiny{$\pm$0.0295}  
& \underline{0.6882} \tiny{$\pm$0.0342}  
& \textbf{0.6473} \tiny{$\pm$0.0343}  
\\

GATSBI 
& 0.7474 \tiny{$\pm$0.0364}  
& 0.6977 \tiny{$\pm$0.0260}  
& 0.7234 \tiny{$\pm$0.0599}  
& 0.9930 \tiny{$\pm$0.0038}  
& 0.9916 \tiny{$\pm$0.0046}  
& 0.9929 \tiny{$\pm$0.0059}  
& 0.9989 \tiny{$\pm$0.0004}  
& 0.9976 \tiny{$\pm$0.0016}  
& 0.9988 \tiny{$\pm$0.0010}  
& 0.9687 \tiny{$\pm$0.0286}  
& 0.9436 \tiny{$\pm$0.0203}  
& 0.9723 \tiny{$\pm$0.0140}  
\\

Simformer     
& \textbf{0.5166} \tiny{$\pm$0.0210}  
& \textbf{0.5100} \tiny{$\pm$0.0090}  
& \textbf{0.5126} \tiny{$\pm$0.0070}  
& 0.6130 \tiny{$\pm$0.0281}  
& 0.6226 \tiny{$\pm$0.0398}  
& 0.5866 \tiny{$\pm$0.0430}  
& 0.6501 \tiny{$\pm$0.0519}  
& 0.6207 \tiny{$\pm$0.0519}  
& 0.5780 \tiny{$\pm$0.0211}  
& \textbf{0.7661} \tiny{$\pm$0.0685}  
& \textbf{0.6201} \tiny{$\pm$0.0490}  
& \underline{0.7192} \tiny{$\pm$0.0721}  
\\

FMPE
& 0.7894 \tiny{$\pm$ 0.0254}  
& 0.7872 \tiny{$\pm$ 0.0049}  
& 0.5821 \tiny{$\pm$ 0.0292}  
& 0.9605\tiny{$\pm$ 0.0019}  
& 0.9716 \tiny{$\pm$ 0.0023}  
& 0.9708 \tiny{$\pm$ 0.0025}  
& 0.9218 \tiny{$\pm$ 0.0122}  
& 0.9100 \tiny{$\pm$ 0.0174}  
& 0.9082 \tiny{$\pm$ 0.0178}
& 0.9698 \tiny{$\pm$ 0.0132}  
& 0.9686 \tiny{$\pm$ 0.0139}  
& 0.9686 \tiny{$\pm$ 0.0139}
\\

\rowcolor{blue!5}        
ConDiSim 
& 0.6776 \tiny{$\pm$0.0283}  
& 0.6420 \tiny{$\pm$0.0247}  
& 0.6087 \tiny{$\pm$0.0262}  
& 0.5970 \tiny{$\pm$0.0228}  
& 0.5733 \tiny{$\pm$0.0250}  
& 0.5655 \tiny{$\pm$0.0117}  
& 0.7030 \tiny{$\pm$0.0469}  
& 0.6641\tiny{$\pm$0.0498}  
& 0.5990\tiny{$\pm$0.0501}  
& 0.92995 \tiny{$\pm$0.00680}  
& 0.87111 \tiny{$\pm$0.07559}  
& 0.72778\tiny{$\pm$0.03867}  
\\

\bottomrule
\end{tabular}
\end{adjustbox}
\end{subtable}

\vspace{0.1cm}
\begin{subtable}{\textwidth}
\centering
 \begin{adjustbox}{max width=\textwidth}
\begin{tabular}{>{\columncolor{gray!10}}l|c|c|c|c|c|c|c|c|c}
\hline
\rowcolor{gray!10}
\textbf{Method} 
  & \multicolumn{3}{c|}{\textbf{Two Moons}} 
  & \multicolumn{3}{c|}{\textbf{Bernoulli GLM}} 
  & \multicolumn{3}{c}{\textbf{SIR}} \\
\hline
\rowcolor{gray!10}
\textbf{Budget}
  & \textbf{10,000} & \textbf{20,000} & \textbf{30,000}
  & \textbf{10,000} & \textbf{20,000} & \textbf{30,000}
  & \textbf{10,000} & \textbf{20,000} & \textbf{30,000} \\
\hline

NPE  
& 0.5749 \tiny{$\pm$0.0331}  
& 0.5931 \tiny{$\pm$0.0329}  
& 0.5411 \tiny{$\pm$0.0195}  
& 0.9121 \tiny{$\pm$0.0338}  
& 0.8988 \tiny{$\pm$0.0306}  
& 0.8898 \tiny{$\pm$0.0565}  
& \underline{0.8378} \tiny{$\pm$0.0396}  
& 0.9122 \tiny{$\pm$0.0207}  
& 0.9405 \tiny{$\pm$0.0125}  
\\

APT  
& \textbf{0.5189} \tiny{$\pm$0.0080}  
& \underline{0.5202} \tiny{$\pm$0.0114}  
& \underline{0.5147} \tiny{$\pm$0.0125}  
& 0.8404 \tiny{$\pm$0.0107}  
& 0.8328 \tiny{$\pm$0.0156}  
& 0.8353 \tiny{$\pm$0.0164}  
& 0.9433 \tiny{$\pm$0.0076}  
& 0.9434 \tiny{$\pm$0.0061}  
& 0.9457 \tiny{$\pm$0.0068}  
\\

GATSBI  
& 0.7770 \tiny{$\pm$0.0630}  
& 0.6634 \tiny{$\pm$0.0503}  
& 0.7467 \tiny{$\pm$0.1036}  
& 0.9999 \tiny{$\pm$0.0001}  
& 0.9999 \tiny{$\pm$0.0000}  
& 0.9997 \tiny{$\pm$0.0005}  
& 0.9990 \tiny{$\pm$0.0015}  
& 0.9912 \tiny{$\pm$0.0161}  
& 0.9838 \tiny{$\pm$0.0157}  
\\

Simformer  
& \underline{0.5273} \tiny{$\pm$0.0229}  
& \textbf{0.5190} \tiny{$\pm$0.0174}  
& \textbf{0.5098} \tiny{$\pm$0.0084}  
& \underline{0.6584} \tiny{$\pm$0.0381}  
& \textbf{0.6131} \tiny{$\pm$0.0225}  
& \underline{0.6154} \tiny{$\pm$0.0265}  
& 0.9378 \tiny{$\pm$0.0053}  
& 0.9520 \tiny{$\pm$0.0064}  
& 0.9478 \tiny{$\pm$0.0049}  
\\

FMPE
& 0.8051 \tiny{$\pm$ 0.0506}  
& 0.7745 \tiny{$\pm$ 0.0261}  
& 0.5706 \tiny{$\pm$ 0.0209}  
& 0.9266 \tiny{$\pm$ 0.0288}  
& 0.9118 \tiny{$\pm$ 0.0376}  
& 0.8431 \tiny{$\pm$ 0.0393}  
& 0.9573 \tiny{$\pm$ 0.0013}  
& \underline{0.8454} \tiny{$\pm$ 0.0014}  
& \textbf{0.8101} \tiny{$\pm$ 0.0020}
\\


\rowcolor{blue!5}  
ConDiSim
& 0.5716 \tiny{$\pm$ 0.1088}  
& 0.5495 \tiny{$\pm$ 0.0802}  
& 0.5226 \tiny{$\pm$ 0.0158}  
& \textbf{0.6276} \tiny{$\pm$ 0.0432}  
& \underline{0.6228} \tiny{$\pm$ 0.0339} 
& \textbf{0.5884} \tiny{$\pm$ 0.0398}   
& \textbf{0.7614}\tiny{$\pm$ 0.0457}   
& \textbf{0.8142} \tiny{$\pm$ 0.0397} 
& \underline{0.8271} \tiny{$\pm$ 0.0426} 

\\
\hline
\end{tabular}
\end{adjustbox}
\end{subtable}
\end{table*}
\newpage
\begin{table*}[h]
\centering
\caption{Training time (in minutes) for different methods across simulation budgets (mean $\pm$ std.\ dev.); best in \textbf{bold}, second-best \underline{underlined}.}
\renewcommand{\arraystretch}{1}
\setlength{\tabcolsep}{10pt}
\label{tab:train_times_all}

\begin{subtable}[t]{\textwidth}
\centering

\begin{adjustbox}{max width=\textwidth}

\begin{tabular}{l|ccc|ccc|ccc}
\rowcolor{gray!10}
\hline
\textbf{Method}
  & \multicolumn{3}{c|}{\textbf{Two Moons}}
  & \multicolumn{3}{c|}{\textbf{Bernoulli GLM}}
  & \multicolumn{3}{c}{\textbf{SIR}} \\
\rowcolor{gray!10}
\hline
\textbf{Budget}
  & \textbf{10,000} & \textbf{20,000} & \textbf{30,000}
  & \textbf{10,000} & \textbf{20,000} & \textbf{30,000}
  & \textbf{10,000} & \textbf{20,000} & \textbf{30,000} \\
\hline

NPE
& 7.04 \tiny{$\pm$0.21}
& 14.21 \tiny{$\pm$0.04}
& 20.95 \tiny{$\pm$0.37}
& 12.17 \tiny{$\pm$0.17}
& 24.96 \tiny{$\pm$0.64}
& 37.43 \tiny{$\pm$0.22}
& 20.10 \tiny{$\pm$5.45}
& 29.62 \tiny{$\pm$3.74}
& 44.92 \tiny{$\pm$9.24} \\

APT
& 90.31 \tiny{$\pm$8.99}
& 186.17 \tiny{$\pm$13.79}
& 268.28 \tiny{$\pm$32.88}
& 101.39 \tiny{$\pm$7.89}
& 197.30 \tiny{$\pm$9.32}
& 267.38 \tiny{$\pm$21.93}
& 40.15 \tiny{$\pm$3.90}
& 73.74 \tiny{$\pm$6.25}
& 112.51 \tiny{$\pm$2.51} \\

GATSBI
& 117.67 \tiny{$\pm$2.43}
& 164.90 \tiny{$\pm$57.10}
& 249.50 \tiny{$\pm$105.05}
& 118.62 \tiny{$\pm$0.18}
& 179.69 \tiny{$\pm$49.96}
& 253.48 \tiny{$\pm$113.38}
& 227.16 \tiny{$\pm$1.44}
& 241.24 \tiny{$\pm$7.58}
& 250.47 \tiny{$\pm$2.22} \\

Simformer
& 3.96 \tiny{$\pm$0.04}
& 7.36 \tiny{$\pm$0.11}
& 10.73 \tiny{$\pm$0.25}
& 17.87 \tiny{$\pm$0.18}
& 35.05 \tiny{$\pm$0.52}
& 52.81 \tiny{$\pm$0.91}
& 4.24 \tiny{$\pm$0.29}
& 7.67 \tiny{$\pm$0.38}
& 10.68 \tiny{$\pm$0.45} \\

FMPE
& \underline{2.63} \tiny{$\pm$ 0.05}  
& \underline{4.73}\tiny{$\pm$ 1.11}  
& \underline{10.65} \tiny{$\pm$ 0.04}  
& \underline{2.65} \tiny{$\pm$ 0.05}  
& \underline{5.75} \tiny{$\pm$ 1.43}  
& \underline{7.64} \tiny{$\pm$ 1.38}  
& \bf{2.78} \tiny{$\pm$ 0.15}  
& \bf{6.39} \tiny{$\pm$ 0.19}  
& \bf{8.94} \tiny{$\pm$ 2.94}
\\

\rowcolor{blue!5}
CondiSim

& {\bf 2.17} \tiny{$\pm$0.68}
& {\bf 4.22} \tiny{$\pm$1.09}
& {\bf 7.02} \tiny{$\pm$0.30}
& {\bf 3.518} \tiny{$\pm$0.19}
& {\bf 6.65} \tiny{$\pm$0.93}
& {\bf 9.56} \tiny{$\pm$2.99}
& \underline{ 2.15} \tiny{$\pm$0.39}
& \underline{4.47} \tiny{$\pm$0.62}
& \underline{6.18} \tiny{$\pm$1.27} \\

\bottomrule
\end{tabular}
\end{adjustbox}
\end{subtable}

\vspace{0.1cm}
\begin{subtable}[t]{\textwidth}
\centering
\begin{adjustbox}{max width=\textwidth}
\begin{tabular}{l|ccc|ccc|ccc}
\rowcolor{gray!10}
\hline
\textbf{Method}
  & \multicolumn{3}{c|}{\textbf{SLCP Distractors}}
  & \multicolumn{3}{c|}{\textbf{Bernoulli GLM Raw}}
  & \multicolumn{3}{c}{\textbf{Lotka Volterra}} \\
\rowcolor{gray!10}
\hline
\textbf{Budget}
  & \textbf{10,000} & \textbf{20,000} & \textbf{30,000}
  & \textbf{10,000} & \textbf{20,000} & \textbf{30,000}
  & \textbf{10,000} & \textbf{20,000} & \textbf{30,000} \\
\hline

NPE
& 21.92 \tiny{$\pm$2.41}
& 43.92 \tiny{$\pm$2.58}
& 64.38 \tiny{$\pm$7.13}
& 25.39 \tiny{$\pm$2.09}
& 61.43 \tiny{$\pm$2.61}
& 85.94 \tiny{$\pm$6.55}
& 80.67 \tiny{$\pm$14.61}
& 160.92 \tiny{$\pm$42.82}
& 188.04 \tiny{$\pm$29.46} \\

APT
& 26.30 \tiny{$\pm$0.36}
& 49.96 \tiny{$\pm$2.73}
& 74.31 \tiny{$\pm$2.54}
& 21.87 \tiny{$\pm$1.28}
& 40.00 \tiny{$\pm$1.37}
& 60.94 \tiny{$\pm$3.00}
& 45.52 \tiny{$\pm$4.85}
& 96.15 \tiny{$\pm$13.77}
& 144.33 \tiny{$\pm$18.33} \\

GATSBI
& 74.37 \tiny{$\pm$2.59}
& 100.59 \tiny{$\pm$2.39}
& 102.30 \tiny{$\pm$0.51}
& 114.76 \tiny{$\pm$1.19}
& 115.22 \tiny{$\pm$1.43}
& 144.10 \tiny{$\pm$24.15}
& 78.91 \tiny{$\pm$6.32}
& 75.73 \tiny{$\pm$0.70}
& 74.99 \tiny{$\pm$0.20} \\

Simformer
& 15.98 \tiny{$\pm$1.22}
& 31.98 \tiny{$\pm$2.34}
& 47.85 \tiny{$\pm$1.86}
& 16.96 \tiny{$\pm$0.51}
& 32.46 \tiny{$\pm$1.89}
& 48.01 \tiny{$\pm$1.44}
& 5.0 \tiny{$\pm$0.34}
& 9.23 \tiny{$\pm$0.54}
& 13.64 \tiny{$\pm$0.21} \\

FMPE
& \underline{2.76} \tiny{$\pm$ 0.08}  
& \textbf{3.08} \tiny{$\pm$ 1.76}  
& \underline{12.97} \tiny{$\pm$ 0.87}  
& \textbf{2.44} \tiny{$\pm$ 0.44}  
& \textbf{4.42} \tiny{$\pm$ 1.37}  
& \textbf{8.05} \tiny{$\pm$ 1.38}  
& \underline{3.63} \tiny{$\pm$ 0.42}  
& \textbf{6.97} \tiny{$\pm$ 0.03}  
& \textbf{9.55} \tiny{$\pm$ 0.72}

\\

\rowcolor{blue!5}
CondiSim
& {\bf 2.02} \tiny{$\pm$0.27}
& \underline{5.16} \tiny{$\pm$0.84}
& {\bf 8.44} \tiny{$\pm$1.02}
& \underline{3.22} \tiny{$\pm$0.72}
& \underline{7.22} \tiny{$\pm$0.50}
& \underline{10.67} \tiny{$\pm$0.69}
& {\bf 2.42} \tiny{$\pm$0.21}
& \underline{7.19} \tiny{$\pm$0.97}
& \underline{11.19} \tiny{$\pm$2.31} \\

\bottomrule
\end{tabular}
\end{adjustbox}
\end{subtable}

\vspace{0.1cm}
\begin{subtable}{\textwidth}
\centering
\renewcommand{\arraystretch}{1.1}

\setlength{\tabcolsep}{2pt}

\begin{adjustbox}{max width=\textwidth}

\begin{tabular}{l|ccc|ccc|ccc|ccc}
\hline
\rowcolor{gray!10}
\hline
\textbf{Method}
 & \multicolumn{3}{c|}{\textbf{Gaussian Mixture}}
 & \multicolumn{3}{c|}{\textbf{Gaussian Linear}}
 & \multicolumn{3}{c|}{\textbf{Gaussian Linear Uniform}}
 & \multicolumn{3}{c}{\textbf{SLCP}} \\
\hline
\rowcolor{gray!10}
\textbf{Budget}
 & \textbf{10,000} & \textbf{20,000} & \textbf{30,000}
 & \textbf{10,000} & \textbf{20,000} & \textbf{30,000}
 & \textbf{10,000} & \textbf{20,000} & \textbf{30,000}
 & \textbf{10,000} & \textbf{20,000} & \textbf{30,000} \\
\hline

NPE
& 19.56 \tiny{$\pm$2.47}
& 42.88 \tiny{$\pm$4.54}
& 64.03 \tiny{$\pm$6.51}
& 12.03 \tiny{$\pm$0.80}
& 24.69 \tiny{$\pm$1.63}
& 38.79 \tiny{$\pm$1.23}
& 16.55 \tiny{$\pm$0.85}
& 30.90 \tiny{$\pm$2.77}
& 49.42 \tiny{$\pm$1.49}
& 42.97 \tiny{$\pm$6.35}
& 97.70 \tiny{$\pm$6.66}
& 117.61 \tiny{$\pm$4.12} \\

APT
& 76.27 \tiny{$\pm$7.02}
& 167.88 \tiny{$\pm$15.99}
& 240.13 \tiny{$\pm$22.74}
& 70.47 \tiny{$\pm$0.88}
& 135.38 \tiny{$\pm$5.75}
& 203.27 \tiny{$\pm$5.97}
& 78.61 \tiny{$\pm$2.11}
& 158.57 \tiny{$\pm$9.32}
& 219.44 \tiny{$\pm$11.83}
& 93.40 \tiny{$\pm$1.26}
& 150.29 \tiny{$\pm$11.10}
& 216.30 \tiny{$\pm$21.38} \\

GATSBI
& 114.07 \tiny{$\pm$2.11}
& 111.30 \tiny{$\pm$0.28}
& 137.46 \tiny{$\pm$19.16}
& 48.19 \tiny{$\pm$0.24}
& 48.70 \tiny{$\pm$0.84}
& 49.12 \tiny{$\pm$0.49}
& 43.27 \tiny{$\pm$0.36}
& 66.43 \tiny{$\pm$5.70}
& 69.10 \tiny{$\pm$0.15}
& 162.72 \tiny{$\pm$14.73}
& 196.19 \tiny{$\pm$119.42}
& 99.39 \tiny{$\pm$5.63} \\

Simformer
& 3.84 \tiny{$\pm$0.04}
& 7.15 \tiny{$\pm$0.09}
& 10.48 \tiny{$\pm$0.21}
& 18.18 \tiny{$\pm$0.13}
& 35.46 \tiny{$\pm$0.17}
& 55.46 \tiny{$\pm$1.13}
& 17.89 \tiny{$\pm$0.23}
& 33.47 \tiny{$\pm$0.40}
& 51.24 \tiny{$\pm$0.89}
& 4.23 \tiny{$\pm$0.01}
& 7.67 \tiny{$\pm$0.13}
& 11.32 \tiny{$\pm$0.40} \\

FMPE
& \textbf{2.16}\tiny{$\pm$ 0.08}  
& \underline{5.36} \tiny{$\pm$ 0.78}  
& \underline{8.31} \tiny{$\pm$ 1.32}  
& \underline{1.78} \tiny{$\pm$ 0.05}  
& \textbf{4.40} \tiny{$\pm$ 0.78}  
& \textbf{5.94} \tiny{$\pm$ 4.31}  
& \textbf{2.15} \tiny{$\pm$ 0.22}  
& \textbf{4.30} \tiny{$\pm$ 3.86}  
& \textbf{8.71} \tiny{$\pm$ 2.12}
& \textbf{1.36} \tiny{$\pm$ 0.14}  
& \textbf{3.71} \tiny{$\pm$ 2.57}  
& \textbf{6.34} \tiny{$\pm$ 2.33}

\\

\rowcolor{blue!5}
ConDiSim
& \underline{2.22} \tiny{$\pm$0.41}
& {\bf 4.64} \tiny{$\pm$0.42}
& {\bf 7.02} \tiny{$\pm$0.98}
& {\bf 1.59} \tiny{$\pm$0.23}
& \underline{4.75} \tiny{$\pm$1.04}
& \underline{7.58} \tiny{$\pm$1.84}
& \underline{2.26} \tiny{$\pm$0.92}
& \underline{4.68} \tiny{$\pm$1.36}
& \underline{8.06} \tiny{$\pm$2.43}
& \underline{1.70} \tiny{$\pm$0.58}
& \underline{4.10} \tiny{$\pm$1.68}
& \underline{8.23} \tiny{$\pm$0.66} \\

\bottomrule
\end{tabular}
\end{adjustbox}
\end{subtable}

\end{table*}

\subsection{Scalability}\label{scale}
We evaluate ConDiSim across a spectrum of SBI benchmarks, ranging from low-dimensional problems (Two Moons, GMM, SIR; 2 parameters) to higher-dimensional settings (Gaussian Linear, Bernoulli GLM; 10 parameters, HH Model; 7 parameters, genetic oscillator; 15 parameters). As shown by C2ST scores in Table~\ref{tab:c2st_main} and posterior visualizations (Figures~\ref{fig:two_moons}–\ref{fig:lv}), ConDiSim maintains high posterior fidelity even as dimensionality and model complexity increase. Training times in Table~\ref{tab:train_times_all} further confirm its computational efficiency, with favorable scaling relative to problem size. Notably, the 15-dimensional genetic oscillator highlights ConDiSim's architectural efficiency and effectiveness, as it preserves inference accuracy without suffering substantial degradation owing to a relatively large parameter space and the noisy, stochastic nature of the simulator.

\subsection{Robustness}\label{robust}
In many real-world problems, observational data are inherently imperfect and often contaminated by measurement noise, confounded by irrelevant features, or available only in raw, unprocessed form. Consequently, it is essential to evaluate how inference methods perform under such adverse conditions. We assess ConDiSim's robustness on two challenging benchmarks: SLCP Distractors, where 92 uninformative variables are added to the observation vector to simulate high-dimensional nuisance features, and Bernoulli GLM Raw, which operates directly on raw binary observations rather than relying on handcrafted summary statistics, emulating realistic scenarios with noisy, unprocessed measurements. The genetic oscillator test problem is also noisy, simulated using the Stochastic Simulation Algorithm (SSA) \citep{gillespie2007stochastic}. As shown in Table~\ref{tab:c2st_main} and Figures~\ref{fig:slcp_dist_posterior}–\ref{fig:glm_raw_posterior}, ConDiSim maintains accurate posterior recovery across both settings, demonstrating resilience to observation noise and feature redundancy that demonstrates its suitability for practical scientific applications where data are high-dimensional, unprocessed, or corrupted by measurement uncertainty.

\subsection{Reliability}\label{reliable}
Beyond predictive accuracy, it is crucial to verify that the learned posteriors are statistically well-calibrated, meaning that the reported uncertainty faithfully captures the true variability of the inference problem. To assess calibration, we use Simulation-Based Calibration (SBC) \citep{talts2018validating}, which is a standard diagnostic that evaluates whether the rank statistics of posterior samples follow their expected uniform distribution. The empirical cumulative distribution function (ECDF) differences shown in Figures~\ref{fig:two_moons_ecdf}–\ref{fig:ecdf_joint_3} remain within the expected bounds, with fluctuations centered around zero, indicating that ConDiSim produces unbiased and well-calibrated posteriors across benchmarks. Such calibration is essential for trustworthy inference in scientific applications, where both accurate point estimates and reliable uncertainty quantification are required.

\section{Conclusion}\label{Conclusion}
We introduce \texttt{ConDiSim}, a DDPM-based conditional diffusion model for simulation-based inference. Across ten benchmark tasks and two real-world problems, \texttt{ConDiSim} effectively handles ill-posed inference, multimodal and high-dimensional posteriors, and demonstrates robustness even with raw, unprocessed observations, or in the presence of noise and distractors. It also adapts naturally to diverse data modalities, while offering substantially faster training time than competing methods. We demonstrate that ConDiSim is particularly suited for inference of stochastic simulators with time series output, where it outperforms other methods and delivers accuracy, well-calibrated posterior estimates, and fast amortized inference.

\section*{Acknowledgments}
The computations/data handling were enabled by the Berzelius resource
provided by the Knut and Alice Wallenberg Foundation at
the National Supercomputer Centre and by the National
Academic Infrastructure for Supercomputing in Sweden
(NAISS) at Chalmers e-Commons at Chalmers, and Uppsala Multidisciplinary Center for Advanced Computational
Science (UPPMAX) at Uppsala University, partially funded
by the Swedish Research Council through grant agreement
nos. 2022-06725 and 2018-05973. Andreas Hellander and Prashant Singh acknowledge
support from the Swedish Research Council through grant
agreement nos. 2023-05167 and 2023-05593 respectively.


\bibliography{ref}

\newpage

\onecolumn

\title{ConDiSim: Conditional Diffusion Models for Simulation-Based Inference\\(Supplementary Material)}
\maketitle


\appendix

\section{Connection between ConDiSim and Score based Diffusion Model}
In this section, we establish a formal theoretical connection between ConDiSim, a DDPM based diffusion model, and the score based diffusion models. We demonstrate its equivalence to a variance preserving stochastic differential equation (SDE) by deriving the continuous time limit of ConDiSim’s discrete forward process, providing a unified perspective on diffusion based generative modeling.

\subsection{Forward Process and Continuous-Time Limit}
ConDiSim employs a discrete time forward process, defined as:
\begin{equation}
    \boldsymbol{\theta}_t = \sqrt{1 - \beta_t} \, \boldsymbol{\theta}_{t-1} + \sqrt{\beta_t} \, \bm{\epsilon}, \quad \bm{\epsilon} \sim \mathcal{N}(0, \mathbf{I}),
\end{equation}

where $\boldsymbol{\theta}_t$ represents the state at time step $t$, $\beta_t$ is the noise schedule, and $\boldsymbol{\epsilon}$ is standard Gaussian noise. To convert this discrete time process to continuous time, we consider the limit as the number of time steps $T \to \infty$ and define a small time step $\Delta t = \frac{1}{T} \to 0$. We introduce a continuous noise schedule $\beta(t)$ such that $\beta_t \approx \beta(t) \Delta t$. Using a first-order Taylor expansion for small $\beta_t$, we approximate $\sqrt{1 - \beta_t} \approx 1 - \frac{1}{2} \beta(t) \Delta t$. Substituting into the forward process:
\begin{equation}
    \boldsymbol{\theta}_t = \left(1 - \frac{1}{2} \beta(t) \Delta t \right) \boldsymbol{\theta}_{t-1} + \sqrt{\beta(t) \Delta t} \, \bm{\epsilon},
\end{equation}

Defining the increment $\Delta \boldsymbol{\theta}_t = \boldsymbol{\theta}_t - \boldsymbol{\theta}_{t-1}$, and noting that the noise term $\sqrt{\beta(t) \Delta t} \, \boldsymbol{\epsilon}$ corresponds to a Wiener process increment $\Delta \mathbf{W} \sim \mathcal{N}(\mathbf{0}, \Delta t \mathbf{I})$, we rewrite the equation as:
\begin{equation}
    \Delta \boldsymbol{\theta}_t = -\frac{1}{2} \beta(t) \boldsymbol{\theta}_{t-1} \Delta t + \sqrt{\beta(t)} \, \Delta \mathbf{W}.
\end{equation}

In the continuous-time limit (\(\Delta t \to 0\)), we replace \(\Delta \boldsymbol{\theta}_t \to d\boldsymbol{\theta}_t\), \(\Delta t \to dt\), \(\Delta \mathbf{W} \to d\mathbf{w}_t\), and \(\boldsymbol{\theta}_{t-1} \simeq \boldsymbol{\theta}_t\), yielding the following stochastic differential equation (SDE):
\begin{equation}
    d\boldsymbol{\theta}_t = -\frac{1}{2} \beta(t) \boldsymbol{\theta}_t \, dt + \sqrt{\beta(t)} \, d\mathbf{w}_t,
\label{eq:ddpm_sde}
\end{equation}
where $\mathbf{w}_t$ denotes standard Brownian motion and $d\mathbf{w}_t \sim \mathcal{N}(0, dt\,\mathbf{I})$. Comparing Equation~\ref{eq:ddpm_sde} with the general form of the stochastic differential equations (SDEs):
\begin{equation}
    d\boldsymbol{\theta}_t = \mathbf{f}(\boldsymbol{\theta}_t, t) \, dt + g(t) \, d\mathbf{w}_t,
\label{eq:forward_sde}
\end{equation}
where, $\mathbf{f}(\boldsymbol{\theta}_t, t)$ is the deterministic drift term and $g(t)$ is the scalar valued diffusion coefficient. Equation~\eqref{eq:ddpm_sde} is a special case of the general SDE, with $\mathbf{f}(\boldsymbol{\theta}_t, t) = -\frac{1}{2} \beta(t) \boldsymbol{\theta}_t$ and $g(t) = \sqrt{\beta(t)}$; this is known as the Variance-Preserving SDE (VP-SDE) \citep{song2021scorebased} and corresponds to an Ornstein–Uhlenbeck (OU) process. The solution to the VP-SDE can be derived using \text{It\^{o}} calculus and is given by:
\begin{equation}
    \boldsymbol{\theta}_t = \sqrt{\alpha_t} \, \boldsymbol{\theta}_0 + \sqrt{1 - \alpha_t} \, \bm{\epsilon},
    \quad \text{where } \alpha_t = \exp\left(-\int_0^t \beta(s) \, ds \right), \quad \bm{\epsilon} \sim \mathcal{N}(0, \mathbf{I}).
\label{eq:forward_sol}
\end{equation}

This expression represents the marginal distribution $q(\boldsymbol{\theta}_t \mid \boldsymbol{\theta}_0)$ under the continuous diffusion process and serves as the continuous time analog of the discrete DDPM forward process:
\begin{equation}
    \boldsymbol{\theta}_t = \sqrt{\alpha_t} \, \boldsymbol{\theta}_0 + \sqrt{1 - \alpha_t} \, \bm{\epsilon}, \quad \alpha_t = \prod_{i=1}^t (1 - \beta_i).
\end{equation}
Thus, the VP-SDE generalizes the discrete Markov chain of ConDiSim to continuous time.

\subsection{Reverse Diffusion Process and Score Matching}
Given a general forward SDE as in Equation~\ref{eq:forward_sde}, the associated \emph{reverse-time SDE} describes the dynamics when running the process backward in time (from $t = T$ to $t = 0$) and corresponds to the reverse diffusion process of ConDiSim. Following \citep{ANDERSON1982313}, the general form of the reverse SDE is:
\begin{equation}
    d\boldsymbol{\theta}_t = \left[\mathbf{f}(\boldsymbol{\theta}_t, t) - g^2(t)\, \nabla_{\boldsymbol{\theta}_t} \log q_t(\boldsymbol{\theta}_t)\right] dt + g(t)\, d\mathbf{w}_t,
\end{equation}
where $q_t(\boldsymbol{\theta}_t)$ is the marginal probability density of $\boldsymbol{\theta}_t$ at time $t$ and \(d\mathbf{w}_t\) is Brownian motion in reverse time. For the VP-SDE (Equation~\eqref{eq:ddpm_sde}), the reverse SDE becomes:
\begin{equation}
    d\boldsymbol{\theta}_t = \left[-\frac{1}{2} \beta(t)\, \boldsymbol{\theta}_t - \beta(t)\, \nabla_{\boldsymbol{\theta}_t} \log q_t(\boldsymbol{\theta}_t)\right] dt + \sqrt{\beta(t)}\, d\mathbf{w}_t.
\end{equation}
In this formulation, all terms are known except for the score function $s(\boldsymbol{\theta}_t, t) = \nabla_{\boldsymbol{\theta}_t} \log q_t(\boldsymbol{\theta}_t)$, which is crucial for simulating the reverse process. Approximating this score is a standard optimization problem known as score matching, which is well-studied in probabilistic sampling, especially in the context of Langevin dynamics. However, since the true score depends on the intractable marginal $q_t(\boldsymbol{\theta}_t)$, we instead introduce a parameterized approximation $s_\phi(\boldsymbol{\theta}_t, t)$ and learn it by minimizing the Fisher divergence (explicit score matching loss):
\begin{equation}
    \phi^* = \arg\min_{\phi}\; \mathbb{E}_{t} \;\mathbb{E}_{q_t(\boldsymbol{\theta}_t)} \left[
        \left\| s_\phi(\boldsymbol{\theta}_t, t) - \nabla_{\boldsymbol{\theta}_t} \log q_t(\boldsymbol{\theta}_t) \right\|^2
    \right].
\end{equation}
To address the intractability of the true score, several methods have been developed, including implicit score matching \citep{hyvarinen2005estimation}, sliced score matching\citep{pmlr-v115-song20a}, and denoising score matching \citep{10.1162/NECO_a_00142}. To align the training objective with ConDiSim, we adopt denoising score matching (DSM), which leverages the tractable conditional score $\nabla_{\boldsymbol{\theta}_t} \log q_t(\boldsymbol{\theta}_t \mid \boldsymbol{\theta}_0)$. The corresponding objective is given by:
\begin{equation}
    \phi^* = \arg\min_{\phi}\; \mathbb{E}_{t} \; \mathbb{E}_{q_0(\boldsymbol{\theta}_0)} \; \mathbb{E}_{q_t(\boldsymbol{\theta}_t | \boldsymbol{\theta}_0)} \left[
        \left\| s_\phi(\boldsymbol{\theta}_t, t) - \nabla_{\boldsymbol{\theta}_t} \log q_t(\boldsymbol{\theta}_t | \boldsymbol{\theta}_0) \right\|^2
    \right].
\end{equation}
Given the closed-form solution for the forward process in Equation~\eqref{eq:forward_sol}, the score function $\nabla_{\boldsymbol{\theta}_t} \log q_t(\boldsymbol{\theta}_t \mid \boldsymbol{\theta}_0)$ becomes tractable and can be written as:
\begin{equation}
    \nabla_{\boldsymbol{\theta}_t} \log q_t(\boldsymbol{\theta}_t | \boldsymbol{\theta}_0) = -\frac{\boldsymbol{\theta}_t - \sqrt{\alpha(t)}\, \boldsymbol{\theta}_0}{1 - \alpha(t)}.
\end{equation}
Reparameterizing $\boldsymbol{\theta}_t = \sqrt{\alpha_t}\, \boldsymbol{\theta}_0 + \sqrt{1 - \alpha_t}\, \bm{\epsilon}$ with $\bm{\epsilon} \sim \mathcal{N}(0, \mathbf{I})$, the DSM objective becomes:
\begin{equation}
    \phi^* = \arg\min_{\phi} \; \mathbb{E}_{t} \; \mathbb{E}_{q_0(\boldsymbol{\theta}_0)} \; \mathbb{E}_{\bm{\epsilon} \sim \mathcal{N}(0, \mathbf{I})} \left[
        \left\| s_\phi(\boldsymbol{\theta}_t, t) + \frac{1}{\sqrt{1-\alpha_t}}\, \bm{\epsilon} \right\|^2
    \right].
\end{equation}
This objective aligns closely with the ConDiSim training objective, where the score is approximated as:
\begin{equation}
    s_\phi(\boldsymbol{\theta}_t, t) \approx -\frac{1}{\sqrt{1 - \alpha_t}} \hat{\bm{\epsilon}}_\phi(\boldsymbol{\theta}_t, t),
\end{equation}
with \(\hat{\bm{\epsilon}}_\phi(\boldsymbol{\theta}_t, t)\) being the predicted noise. This equivalence demonstrates that ConDiSim's training objective is a rescaled form of the DSM objective, unifying the discrete DDPM framework with continuous-time score-based diffusion models. This analysis reveals that, although ConDiSim is formulated as a discrete DDPM, it is fundamentally equivalent to a score-based diffusion model in the continuous-time limit. The VP-SDE provides a principled framework for analyzing ConDiSim's dynamics: the forward diffusion process can be understood through the Fokker–Planck equation, which characterizes convergence properties of the marginal distributions, while the reverse diffusion process connects naturally to annealed Langevin dynamics.

\section{SBI Problem Description}\label{sbi_problems}
We present the general setup for simulation-based inference (SBI), which is employed across various benchmarking problems to estimate the posterior distribution \( p(\boldsymbol{\theta} | \mathbf{y}) \) over parameters \(\boldsymbol{\theta}\), given observed data \(\mathbf{y}\). As outlined in \citep{lueckmann2021benchmarking}, the framework for SBI problems typically includes the following components:
\begin{itemize}
    \item \textbf{Prior Distribution:} A prior distribution \( p(\boldsymbol{\theta}) \) that defines the initial beliefs about the parameter vector before any data is observed. This distribution serves as the starting point for the inference process.
    \item \textbf{Simulation Model:} A stochastic simulation model \( \mathcal{M} \) that allows empirical evaluation of the likelihood function \( p(\mathbf{y} | \boldsymbol{\theta}) \), illustrating how synthetic data \(\mathbf{y}\) is generated given specific parameter values \(\boldsymbol{\theta}\). This model encapsulates the underlying mechanisms of the problem and provides samples from the joint distribution \( p(\boldsymbol{\theta}, \mathbf{y}) \), enabling inference without direct likelihood computation.
    \item \textbf{Dimensionality:} The dimensionality of both the parameter vector \(\boldsymbol{\theta}\) and the observed data \(\mathbf{y}\) is crucial for parameter inference. It can vary greatly between problems, and directly impacts the complexity and computational cost of the inference task. High-dimensional spaces often present greater challenges in accurately estimating the posterior distribution.
\end{itemize}

\subsection{Two Moons}
This parameter inference task is designed to test the performance of inference methods under multi-modal conditions, featuring a posterior distribution with both global (bi-modal) and local (crescent-shaped) characteristics.
\begin{itemize}
    \item \textbf{Prior:} \( \boldsymbol{\theta} \sim \mathcal{U}(-1, 1) \).
    \item \textbf{Simulator:} 
    \(
    \mathbf{y} | \boldsymbol{\theta} = r \begin{bmatrix} \cos(\alpha) \\ \sin(\alpha) \end{bmatrix} + 0.25 \begin{bmatrix} -|\theta_1 + \theta_2|/\sqrt{2} \\ (-\theta_1 + \theta_2)/\sqrt{2} \end{bmatrix},
    \)
    where \( \alpha \sim \mathcal{U}(-\pi/2, \pi/2) \) and \( r \sim \mathcal{N}(0.1, 0.01^2) \).
    \item \textbf{Dimensionality:} \( \boldsymbol{\theta} \in \mathbb{R}^2, \mathbf{y} \in \mathbb{R}^2 \).
\end{itemize}

\subsection{Gaussian Mixture}
This task challenges inference methods to estimate the shared mean of a mixture of Gaussian distributions with distinct covariances, thereby assessing their performance in handling multi-modal and overlapping data distributions.
\begin{itemize}
    \item \textbf{Prior:} \( \boldsymbol{\theta} \sim \mathcal{U}(-10, 10) \).
    \item \textbf{Simulator:} 
    \(\mathbf{y} | \boldsymbol{\theta} = 0.5 \left( \mathcal{N}(\boldsymbol{\theta}, \mathbf{I}) + \mathcal{N}(\boldsymbol{\theta}, 0.01 \mathbf{I}) \right) \).
    \item \textbf{Dimensionality:} \( \boldsymbol{\theta} \in \mathbb{R}^2, \mathbf{y} \in \mathbb{R}^2 \).
\end{itemize}

\subsection{Gaussian Linear}
This task involves estimating the mean vector \(\boldsymbol{\theta}\) of a multivariate Gaussian distribution with a fixed and known covariance matrix. Since both the prior and the likelihood are Gaussian, they form a conjugate pair, resulting in a Gaussian posterior distribution. This conjugacy makes the task an ideal benchmark for evaluating inference methods, as it allows for a clear assessment of their ability to accurately capture known posterior distributions.
\begin{itemize}
    \item \textbf{Prior:} \( \boldsymbol{\theta} \sim \mathcal{N}(\mathbf{0}, 0.1 \mathbf{I}) \).
    \item \textbf{Simulator:} Generates data \(\mathbf{y}\) using \( \mathbf{y} | \boldsymbol{\theta} \sim \mathcal{N}(\boldsymbol{\theta}, 0.1 \mathbf{I}) \).
    \item \textbf{Dimensionality:} \( \boldsymbol{\theta} \in \mathbb{R}^{10}, \mathbf{y} \in \mathbb{R}^{10} \).
\end{itemize}

\subsection{Gaussian Linear Uniform}
This task is a variant of the Gaussian Linear problem, where the mean parameter \(\boldsymbol{\theta}\) is assigned a uniform prior distribution  \(\boldsymbol{\theta} \sim \mathcal{U}(-1, 1)\)  instead of a Gaussian prior. The introduction of a uniform prior imposes bounded support on the parameters, altering the posterior distribution from a Gaussian to a truncated Gaussian. This modification tests the inference method’s ability to handle non-Gaussian priors and to accurately capture the resulting posterior distributions.

\subsection{SLCP}
The SLCP task presents a more challenging inference scenario where, despite the simplicity of the likelihood function, the posterior distribution exhibits complex, non-linear dependencies among parameters. The parameter vector \(\boldsymbol{\theta}\) is five-dimensional, and the simulator generates two-dimensional data points by applying non-linear transformations to \(\boldsymbol{\theta}\). This setup assesses the inference method’s capability to accurately capture intricate posterior structures arising from non-linear relationships between parameters and observations.
\begin{itemize}
    \item \textbf{Prior:} \( \boldsymbol{\theta} \sim \mathcal{U}(-3, 3) \).
    \item \textbf{Simulator:} Outputs data \(\mathbf{y} | \boldsymbol{\theta} = (y_1, \ldots, y_4)\), where:
    \[
        y_i \sim \mathcal{N} \left( \begin{bmatrix} \theta_1 \\ \theta_2 \end{bmatrix}, \begin{bmatrix} \theta_3^2 & \tanh(\theta_5) \theta_3 \theta_4 \\ \tanh(\theta_5) \theta_3 \theta_4 & \theta_4^2 \end{bmatrix} \right)
    \].
    \item \textbf{Dimensionality:} \( \boldsymbol{\theta} \in \mathbb{R}^5, \mathbf{y} \in \mathbb{R}^8 \).
\end{itemize}

\subsection{SLCP with Distractors}
An extension of the SLCP task that incorporates additional non-informative dimensions (distractors) into the observed data. This setup evaluates the robustness of inference methods when faced with high-dimensional data containing redundant information.
\begin{itemize}
    \item \textbf{Prior:} \( \boldsymbol{\theta} \sim \mathcal{U}(-3, 3) \).
    \item \textbf{Simulator:} Outputs data $y|\theta = (y_1, \ldots, y_{100})$, where a permutation function $p(y)$ reorders $y$ with a fixed random permutation.
    \item \textbf{Dimensionality:} \( \boldsymbol{\theta} \in \mathbb{R}^5, \mathbf{y} \in \mathbb{R}^{100} \).
\end{itemize}

\subsection{Bernoulli GLM}
This task involves performing inference in a Generalized Linear Model (GLM) with Bernoulli-distributed observations and a 10-dimensional parameter vector \(\boldsymbol{\theta}\). In this model, each binary observation is linked to the parameters through a logistic function, modeling the probability of success in a Bernoulli trial. Gaussian priors are imposed on the coefficients \(\beta\) and the function f, promoting smoothness in the parameter estimates. This setup evaluates the inference method’s ability to handle discrete data and accurately capture the posterior distribution in models with latent smoothness constraints.
\begin{itemize}
    \item \textbf{Prior:} \( \beta \sim \mathcal{N}(\mathbf{0}, 2) \), \( f \sim \mathcal{N}(\mathbf{0}, (\mathbf{F}^{\top} \mathbf{F})^{-1}) \).
    \item \textbf{Simulator:} \( \mathbf{y} | \boldsymbol{\theta} = (y_1, \ldots, y_{100}) \), where \( y_i \sim \text{Bern}(\eta(\mathbf{V}_i^{\top} f + \beta)) \), with \( \eta(z) = \frac{\exp(z)}{1 + \exp(z)} \).
    \item \textbf{Dimensionality:} \( \boldsymbol{\theta} \in \mathbb{R}^{10}, \mathbf{y} \in \mathbb{R}^{10} \).
    \item \textbf{Fixed Parameters:} \( T = 100 \).
\end{itemize}

\subsection{Bernoulli GLM Raw}
This task is similar to the Bernoulli GLM described above but utilizes raw binary observations instead of sufficient statistics. Without the simplifying benefit of sufficient statistics, the inference method must directly model the relationship between the parameters and the raw data, increasing the complexity of the inference task. This scenario tests the method’s capability to accurately infer parameters in the presence of high-dimensional, discrete, and unprocessed data, challenging its robustness and flexibility.
\begin{itemize}
    \item \textbf{Prior:} Same as Bernoulli GLM.
    \item \textbf{Simulator:} \( \mathbf{y} | \boldsymbol{\theta} = (y_1, \ldots, y_{100}) \), where \( y_i \sim \text{Bern}(\eta(\mathbf{V}_i^{\top} f + \beta)) \).
    \item \textbf{Dimensionality:} \( \boldsymbol{\theta} \in \mathbb{R}^{10}, \mathbf{y} \in \mathbb{R}^{100} \).
    \item \textbf{Fixed Parameters:} \( T = 100 \).
\end{itemize}

\subsection{SIR Model}
The SIR model is an epidemiological framework that describes the number of individuals in three compartments: Susceptible (\(S\)), Infectious (\(I\)), and Recovered (\(R\)). The dynamics of the system are governed by the following differential equations:
\begin{equation}
\frac{d}{dt} \begin{pmatrix} S \\ I \\ R \end{pmatrix} 
= \begin{pmatrix} -\beta S \\ \beta S - \gamma I \\ \gamma I \end{pmatrix} \frac{I}{N},
\end{equation}
where \(N\) represents the total population size. The parameter vector \(\boldsymbol{\theta}\) is defined as:
\[
\boldsymbol{\theta} = \begin{pmatrix} \beta \\ \gamma \end{pmatrix} \sim \text{LogNormal} \left( \begin{pmatrix} \log(0.4) \\ \log(1/8) \end{pmatrix}, \begin{pmatrix} 0.5 \\ 0.2 \end{pmatrix} \right).
\]
\begin{itemize}
    \item \textbf{Simulator:} \( \mathbf{y} | \boldsymbol{\theta} = (y_1, \ldots, y_{10}) \), where \( y_i \sim \mathcal{B}(1000, \frac{I}{N}) \).
    \item \textbf{Dimensionality:} \( \boldsymbol{\theta} \in \mathbb{R}^2, \mathbf{y} \in \mathbb{R}^{10} \).
    \item \textbf{Fixed Parameters:} Population size \( N = 1,000,000 \), task duration \( T = 160 \).
\end{itemize}

\subsection{Lotka-Volterra}
The Lotka-Volterra (LV) model is a classical ecological framework that describes the interaction between prey (\(X\)) and predator (\(Y\)) populations. The dynamics are expressed as:
\begin{equation}
\frac{d}{dt} \begin{pmatrix} X \\ Y \end{pmatrix} 
= \begin{pmatrix} \alpha X - \beta XY \\ -\gamma Y + \delta XY \end{pmatrix},
\end{equation}
with the parameter vector defined as:
\[
\boldsymbol{\theta} = \begin{pmatrix} \alpha \\ \beta \\ \gamma \\ \delta \end{pmatrix} \sim \text{LogNormal} \left( \begin{pmatrix} -0.125 \\ -3 \\ -0.125 \\ -3 \end{pmatrix}, \begin{pmatrix} 0.5 \\ 0.5 \\ 0.5 \\ 0.5 \end{pmatrix} \right).
\]
\begin{itemize}
    \item \textbf{Simulator:} \( \mathbf{y} | \boldsymbol{\theta} = (y_1, \ldots, y_{10}) \), where \( y_i \sim \text{LogNormal}(\log(X), 0.1) \) and \( y_{2i} \sim \text{LogNormal}(\log(Y), 0.1) \).
    \item \textbf{Dimensionality:} \( \boldsymbol{\theta} \in \mathbb{R}^4, \mathbf{y} \in \mathbb{R}^{20} \).
    \item \textbf{Fixed Parameters:} Task duration \( T = 20 \), initial conditions \( (X(0), Y(0)) = (30, 1) \).
\end{itemize}

\subsection{Hodgkin-Huxley Model}
The Hodgkin-Huxley (HH) model describes the generation and propagation of action potentials in neurons by modeling the membrane voltage \( V(t) \) as influenced by ionic currents through voltage-gated sodium (\(\text{Na}^+\)), potassium (\(\text{K}^+\)), and leak channels. We adopt the formulation from \citep{gloecklerall}. The model is given by the following set of stochastic differential equations:
\begin{equation}
\frac{dV}{dt} = \frac{I_{\text{inj}}(t) - g_{\text{Na}} m^3 h (V - E_{\text{Na}}) - g_{\text{K}} n^4 (V - E_{\text{K}}) - g_{\text{L}} (V - E_{\text{L}})}{C_m} + 0.05 \, dw_t,
\end{equation}
where \( dw_t \) represents a Gaussian white noise term that captures stochastic fluctuations in the membrane potential.

The gating variables \( m(t), h(t), n(t) \) follow first-order kinetics:
\begin{align}
\frac{dm}{dt} &= \alpha_m(V)(1 - m) - \beta_m(V)m, \\
\frac{dh}{dt} &= \alpha_h(V)(1 - h) - \beta_h(V)h, \\
\frac{dn}{dt} &= \alpha_n(V)(1 - n) - \beta_n(V)n,
\end{align}
The voltage-dependent rate functions \(\alpha_x(V)\) and \(\beta_x(V) (x \in \{m,h,n\})\) defined as follows:
\[
\alpha_m(V) = 0.32 \times \frac{\text{efun}(-0.25 (V - V_0 - 13.0))}{0.25}, \quad 
\beta_m(V) = 0.28 \times \frac{\text{efun}(0.2 (V - V_0 - 40.0))}{0.2},
\]
\[
\alpha_h(V) = 0.128 \times \exp \left( \frac{-(V - V_0 - 17.0)}{18.0} \right), \quad 
\beta_h(V) = \frac{4.0}{1 + \exp \left( \frac{-(V - V_0 - 40.0)}{5.0} \right)},
\]
\[
\alpha_n(V) = 0.032 \times \frac{\text{efun}(-0.2 (V - V_0 - 15.0))}{0.2}, \quad 
\beta_n(V) = 0.5 \times \exp \left( \frac{-(V - V_0 - 10.0)}{40.0} \right).
\]
Here, \(\text{efun}(x)\) is defined to ensure numerical stability when evaluating small arguments:
\[
\text{efun}(x) = 
\begin{cases} 
1 - \frac{x}{2} & \text{if } x < 1 \text{e-4}, \\
\frac{x}{\exp(x) - 1.0} & \text{otherwise}.
\end{cases}
\]
The metabolic cost \( H(t) \), representing the energy consumption due to sodium channel activity, is modeled as:
\begin{equation}
\frac{dH}{dt} = g_{\text{Na}} m^3 h (V - E_{\text{Na}}).
\end{equation}
\begin{itemize}
    \item \textbf{Parameters:} The parameter vector is \(\boldsymbol{\theta} = (C_m, g_{\text{Na}}, g_{\text{K}}, g_{\text{L}}, E_{\text{Na}}, E_{\text{K}}, E_{\text{L}})\), where \( C_m \) is the membrane capacitance, \( g_{\text{Na}}, g_{\text{K}}, g_{\text{L}} \) are the conductances for sodium, potassium, and leak currents, and \( E_{\text{Na}}, E_{\text{K}}, E_{\text{L}} \) are the corresponding reversal potentials.

    \item \textbf{Prior:} The parameters \(\boldsymbol{\theta}\) are sampled from a uniform distribution:
    \[
    \boldsymbol{\theta} \sim \mathcal{U}\left([1.0, 60, 10, 0.1, 40, -100, -90], [2.0, 120, 30, 0.5, 70, -60, -60]\right).
    \]
    \item \textbf{Simulator:} The HH model is simulated for 200 ms. An input current of \(I_{\text{inj}}(t) = 4 \, \text{mA}\) is applied between \(50\) ms and \(150\) ms. The voltage equation is integrated with added Gaussian white noise \((0.05\,d w_t)\). To reduce dimensionality, the raw voltage trace is summarized into a set of statistics:
    \begin{itemize}
        \item Spike count,
        \item Mean and standard deviation of the resting potential,
        \item Mean voltage during spiking,
        \item Higher-order moments (normalized skewness and kurtosis) of the voltage response,
        \item Total energy consumption due to sodium current.
    \end{itemize}
    \item \textbf{Dimensionality:} The parameter space is \(\boldsymbol{\theta} \in \mathbb{R}^7\), and the output summary statistics \(\mathbf{y} \in \mathbb{R}^8\).
    \item \textbf{Fixed Parameters:} The initial membrane voltage is \( V_0 = -65.0 \, \text{mV} \), with gating variables initialized to their steady-state values at \(V_0\). Simulations use a time resolution of \( 5000 \, \text{steps} \) over \( 200 \, \text{ms} \) simulation window.
\end{itemize}


\subsection{Genetic Oscillator}

The genetic oscillator model describes a biochemical reaction network involving \(9\) species governed by \(18\) reactions, parameterized by \(15\) reaction rate constants. This network is based on a positive–negative feedback loop, where the activator protein \(A\) upregulates transcription of both its gene and the repressor protein \(R\). The repressor \(R\) sequesters \(A\) by forming a complex \(C\), thereby inhibiting further activation \citep{doi:10.1073/pnas.092133899,akesson2022convolutional}. The species in the network are:
\begin{itemize}
    \item \(D_A, D_A^*\): Activator gene in unbound (\(D_A\)) and bound states (\(D_A^*\)).
    \item \(D_R, D_R^*\): Repressor gene in unbound (\(D_R\)) and bound states (\(D_R^*\)).
    \item \(M_A, M_R\): mRNA transcribed from activator (\(M_A\)) and repressor genes (\(M_R\)).
    \item \(A\): Activator protein.
    \item \(R\): Repressor protein.
    \item \(C\): Complex formed by \(A\) and \(R\).
\end{itemize}

\begin{figure}[h]
    \centering
    \includegraphics[width=0.6\textwidth]{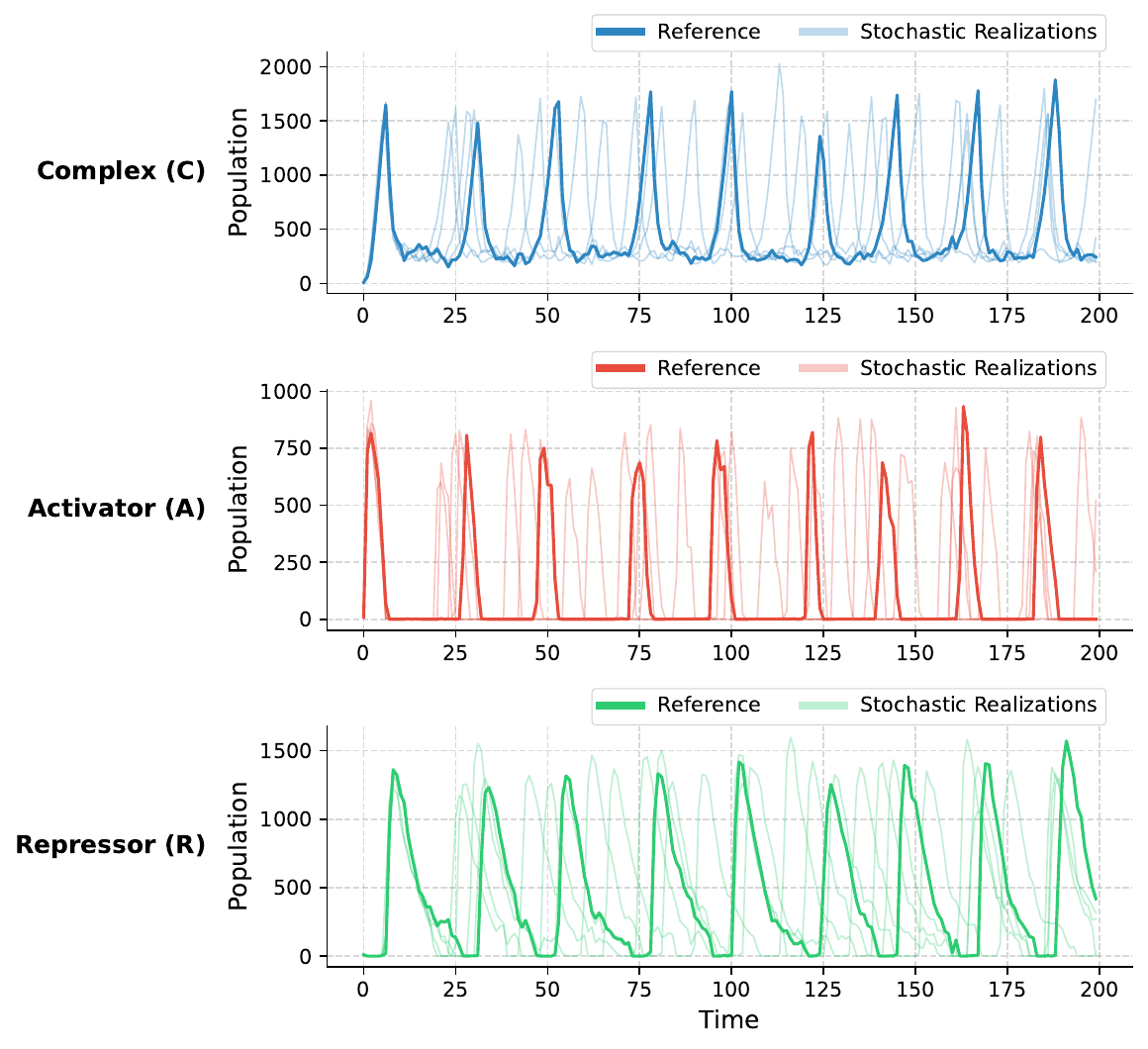}
    \caption{Time series trajectories for three key species in the Vilar oscillator system: Complex (C), Activator (A), and Repressor (R). The darker line shows a reference trajectory obtained using the true parameter values, while the four lighter lines represent additional stochastic realizations under identical conditions. This visualization highlights the intrinsic variability of the system while preserving its consistent oscillatory behavior.}

    \label{fig:vilar_stochastic}
\end{figure}

The network includes the following reactions:
\begin{align*}
D_A, A &\xrightarrow{\gamma_A} D_A^*, & D_A^* &\xrightarrow{\theta_A} D_A, \\
D_A^* &\xrightarrow{\theta_A} D_A,\, A, & D_A^* &\xrightarrow{\alpha_A^*} D_A^*,\, M_A, \\
D_A   &\xrightarrow{\alpha_A} D_A,\, M_A, \\
D_R, A &\xrightarrow{\gamma_R} D_R^*, & D_R^* &\xrightarrow{\theta_R} D_R, \\
D_R^* &\xrightarrow{\theta_A} D_R^*,\, A, & D_R^* &\xrightarrow{\alpha_R^*} D_R^*,\, M_R, \\
D_R   &\xrightarrow{\alpha_R} D_R,\, M_R, \\
M_A &\xrightarrow{\beta_A} M_A + A, & M_R &\xrightarrow{\beta_R} M_R + R, \\
M_A &\xrightarrow{\delta_{M_A}} \phi, & M_R &\xrightarrow{\delta_{M_R}} \phi, \\
A, R &\xrightarrow{\gamma_C} C, & C &\xrightarrow{\delta_A} R, \\
A   &\xrightarrow{\delta_A} \phi, & R &\xrightarrow{\delta_R} \phi.
\end{align*}

The reaction rates include:
\[
\begin{aligned}
\theta_A, \theta_R &: \text{Gene unbinding rates}, \quad \gamma_A, \gamma_R, \gamma_C: \text{Binding rates}, \\
\alpha_a, \alpha_a', \alpha_r, \alpha_r' &: \text{Transcription rates}, \quad \beta_a, \beta_r: \text{Translation rates}, \\
\delta_{ma}, \delta_{mr} &: \text{mRNA degradation rates}, \quad \delta_a, \delta_r: \text{Protein degradation rates}.
\end{aligned}
\]
\paragraph{Prior:} 
The parameter vector \(\boldsymbol{\theta} \in \mathbb{R}^{15}\) is defined with prior bounds:
\[
\boldsymbol{\theta} \sim \mathcal{U}([0, 100, 0, 20, 10, 1, 1, 0, 0, 0, 0.5, 0, 0, 0, 0], [80, 600, 4, 60, 60, 7, 12, 2, 3, 0.7, 2.5, 4, 3, 70, 300])
\].

\paragraph{Simulator:} 
The model is simulated using Gillespie's stochastic simulation algorithm (SSA) over \(200\) time steps. The key observables are the copy numbers of the complex \(C\), activator \(A\), and repressor \(R\). 

\paragraph{Dimensionality:}
The parameter vector \(\boldsymbol{\theta} \in \mathbb{R}^{15}\), and the output time series data has the shape \(N \times S \times T\), where \(N\) is the number of samples, \(S = 3\) is the number of observed species, and \(T = 200\) is the number of time steps. To reduce the dimensionality and make the data more manageable, we train a simple autoencoder. The autoencoder compresses the data from \(N \times 3 \times 200\) to \(N \times 15\), effectively replacing the time series with a set of summary statistics.

\paragraph{Fixed Parameters:} 
The model is initialized with the following parameter values and species concentrations:

\begin{itemize}
    \item \textbf{ True parameters:}
    \[
    \begin{aligned}
    \alpha_a &= 50, \quad \alpha_a' = 500, \quad \alpha_r = 0.01, \quad \alpha_r' = 50, \\
    \beta_a &= 50, \quad \beta_r = 5, \quad \delta_{ma} = 10, \quad \delta_{mr} = 0.5, \\
    \delta_a &= 1, \quad \delta_r = 0.2, \quad \gamma_a = 1, \quad \gamma_r = 1, \\
    \gamma_c &= 2, \quad \theta_a = 50, \quad \theta_r = 100.
    \end{aligned}
    \]

    \item \textbf{Initial Species Values:}
    \[
    D_A = 1, \; D_A^* = 0, \; M_A = 0, \; D_R = 1, \; D_R^* = 0, \; M_R = 0, \; C = 10, \; A = 10, \; R = 10.
    \]

    \item \textbf{Simulation Time:} The system is simulated over \(200\) time steps uniformly distributed across \(200 \, \text{ms}\).

\end{itemize}



\section{Hyperparameters}\label{hyperparams}

In this section, we describe the hyperparameters and simulation environments used across all methods. Training data for the SBI benchmark problems were generated using \texttt{sbibm} \citep{lueckmann2021benchmarking}, except for the Lotka–Volterra (LV) and SIR models, which were simulated with \texttt{BayesFlow} \citep{bayesflow_2023_software}. For SimFormer, we adopt the hyperparameters reported in \citet{gloecklerall} for both the general SBI setting and the Hodgkin–Huxley (HH) model. For the Genetic Oscillator model, we follow the simulation and preprocessing setup of \citet{akesson2022convolutional}. Specifically, SimFormer is trained on a simulation budget of 30,000 using an 8-layer Transformer with 8 attention heads, 64-dimensional head projections, a 512-unit MLP, and dropout rate 0.1; it is optimized with Adam (learning rate $10^{-3}$) for 100,000 iterations with batch size 128, and posterior samples are generated via Euler–Maruyama integration over 1,000 diffusion steps. For GATSBI, we start from the task-specific hyperparameters in \citet{Ramesh22} and adjust them as needed based on problem structure and computational constraints. All experiments were run on a system with dual 16-core Intel Xeon Gold 6226R CPUs, 576\,GB RAM, and an NVIDIA Tesla T4 GPU (16\,GB).

For NPE, we adopt the setup of \citet{lueckmann2021benchmarking}, who evaluate NPE across ten \texttt{sbibm} benchmark tasks. Appendix H of their work shows that Neural Spline Flows (NSF) consistently outperform Masked Autoregressive Flows (MAF), which were originally proposed as the density estimator for NPE. We therefore use the NSF implementation from the \texttt{sbi} package \citep{tejero-cantero2020sbi} in all experiments. We also use \texttt{sbi}'s implementation of Automatic Posterior Transformation (APT, or SNPE-C). As a sequential method, APT allocates the total simulation budget uniformly across rounds.



For \texttt{ConDiSim}, we tuned hyperparameters individually for each task in the \texttt{sbibm} benchmark suite \citep{lueckmann2021benchmarking} while keeping the training pipeline fixed. We optimized with AdamW \citep{loshchilov2019decoupled} using a YOLOX-inspired learning rate schedule: a quadratic warmup over the first $10\%$ of steps from $0.2\eta$ to the base learning rate $\eta$, followed by cosine annealing down to $10^{-6}$, and a constant phase at $10^{-6}$ for the final $5\%$ of steps; weight decay was set to $\lambda = 10^{-4}$. We use scaled linear and scaled quadratic noise schedule \citep{NEURIPS2020_4c5bcfec} with $\beta_{\text{start}} = 10^{-4}s$ and $\beta_{\text{end}} = 0.02s$ with $s = 1000/T$ to maintain consistent noise levels across varying diffusion steps $T$. Inputs were standardized to zero mean and unit variance, and a $70/30$ train--validation split was employed to handle the high variance of posterior estimation. Training stability was ensured via gradient clipping (max norm $5.0$) and SNR-based loss weighting \citep{Hang2023EfficientDT} with $\gamma = 5.0$, where timestep weights $w_t = \min(\mathrm{SNR}_t, \gamma)/\mathrm{SNR}_t$ and $\mathrm{SNR}_t = \bar{\alpha}_t/(1 - \bar{\alpha}_t)$ emphasize learning at noisier timesteps. Early stopping (patience $30$ after epoch $50$) and SiLU activations throughout the network were also used. Task-specific hyperparameters are provided in Table~\ref{tab:condisim_hyperparams}.
\begin{table*}[ht]
\centering
\caption{ConDiSim hyperparameters.}
\label{tab:condisim_hyperparams}
\renewcommand{\arraystretch}{1.2}  
\setlength{\tabcolsep}{10pt}  
\scriptsize 
\begin{adjustbox}{max width=\textwidth}  
\begin{tabular}{>{\columncolor{gray!10}}l|c|c|c|c|c|c}
\hline
\rowcolor{gray!10}
\textbf{Benchmark} & \textbf{\# Diffusion Blocks} & \textbf{Hidden Dim} & \textbf{Scheduler} & \textbf{\# Timesteps} & \textbf{Batch Size} & \textbf{Learning Rate} \\
\hline
SLCP                       & 6 & 128  & linear     & 200  & 32  & 1e-3 \\
SLCP with Distractors      & 6 & 128  & quadratic     & 1000 & 50  & 1e-3 \\
Two Moons                  & 4  & 64  & cosine & 160  & 32  & 1e-3 \\
Gaussian Mixture           & 4 & 64  & cosine     & 160   & 32  & 1e-3 \\
Gaussian Linear           & 6 & 64  & linear     & 100   & 50  & 2e-4 \\
Gaussian Linear Uniform    & 6 & 64  & cosine    & 200   & 50  & 2e-4 \\
SIR                        & 4 & 64  & linear     & 100   & 32  & 1e-4 \\
Bernoulli GLM              & 6 & 128  & linear     & 200  & 32 & 1e-3 \\
Bernoulli GLM Raw          & 6 & 128  & linear     & 200  & 32 & 1e-3 \\
Lotka-Volterra             & 6 & 128  & cosine     & 200  & 50 & 2e-4 \\
HH Model                   & 6 & 128  & quadratic     & 1000  & 50  & 1e-3 \\
Genetic Oscillator         & 6 & 128  & cosine     & 300  & 64  & 1e-4 \\
\hline
\end{tabular}
\end{adjustbox}
\end{table*}

\newpage
\section{Additional Results}
\label{addtional_results}
\subsection{Posterior Evolution for Two Moons}
\begin{figure}[!h]
    \centering
    \includegraphics[width=\linewidth]{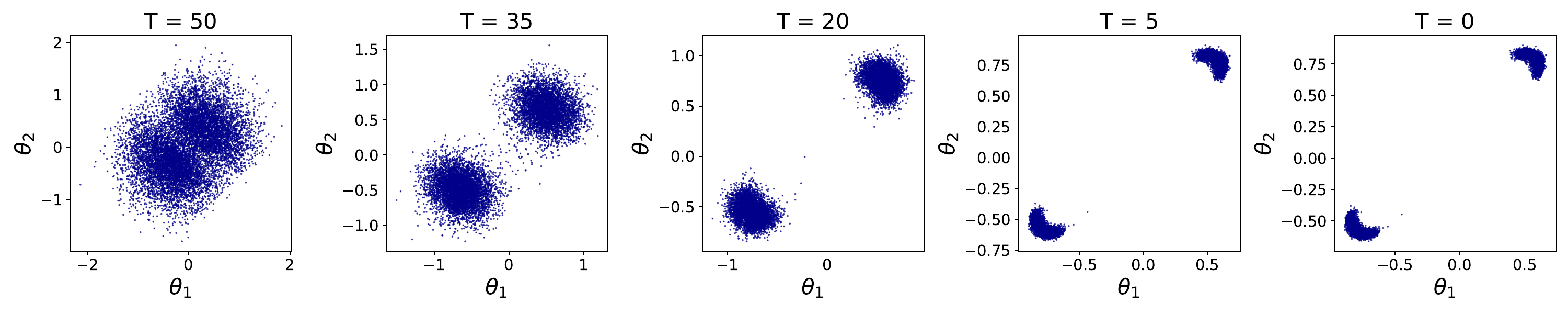}
    \caption{Evolution of the posterior distribution for the Two Moons task, showing the transition from a broad, noisy distribution to a well-defined bi-modal structure as \( T \to 0 \).}
    \label{fig:posterior_evolution}
\end{figure}

The evolution of the approximated posterior for the Two Moons test problem is shown in Fig. \ref{fig:posterior_evolution}. 
\subsection{Comparison with Simformer on HH and Genetic Oscillator Models}
Unlike standard SBI benchmarks with analytical posteriors, the Hodgkin–Huxley (HH) and Genetic Oscillator models lack closed-form solutions, preventing direct evaluation via C2ST or MMD. Having benchmarked ConDiSim on ten standard SBI tasks (Section~\ref{posterior_plots_main}), we now compare it against Simformer, the closest competing method, on these real-world problems.

\begin{figure*}[!t]
    \centering
    \begin{subfigure}[t]{0.45\linewidth}
        \centering
        \includegraphics[width=\linewidth]{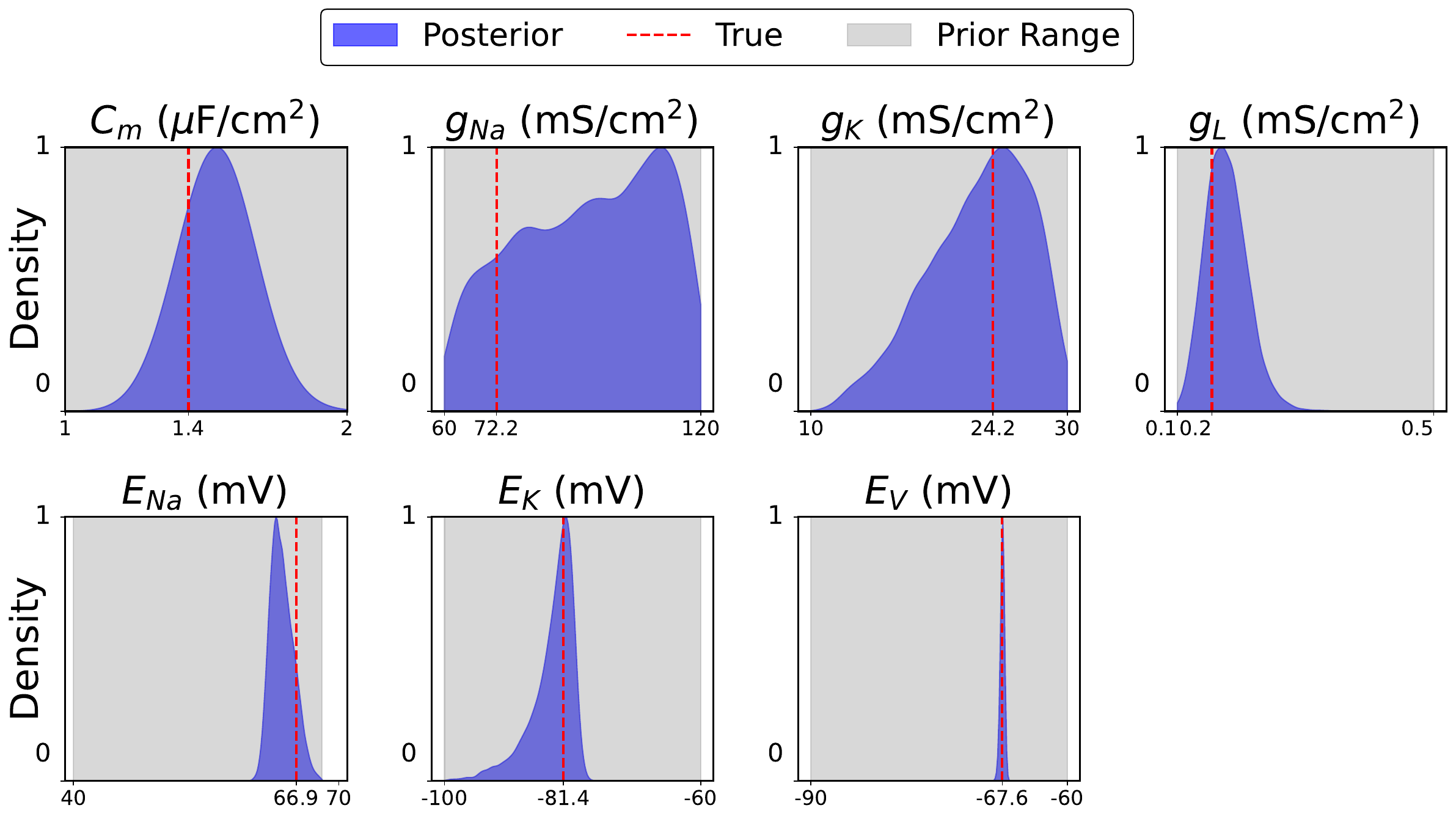}
        \caption{Hodgkin–Huxley: Approximated Posterior.}
        \label{fig:simformer_hh_posterior}
    \end{subfigure}%
    \hspace{4mm}
    \begin{subfigure}[t]{0.45\linewidth}
        \centering
        \includegraphics[width=\linewidth]{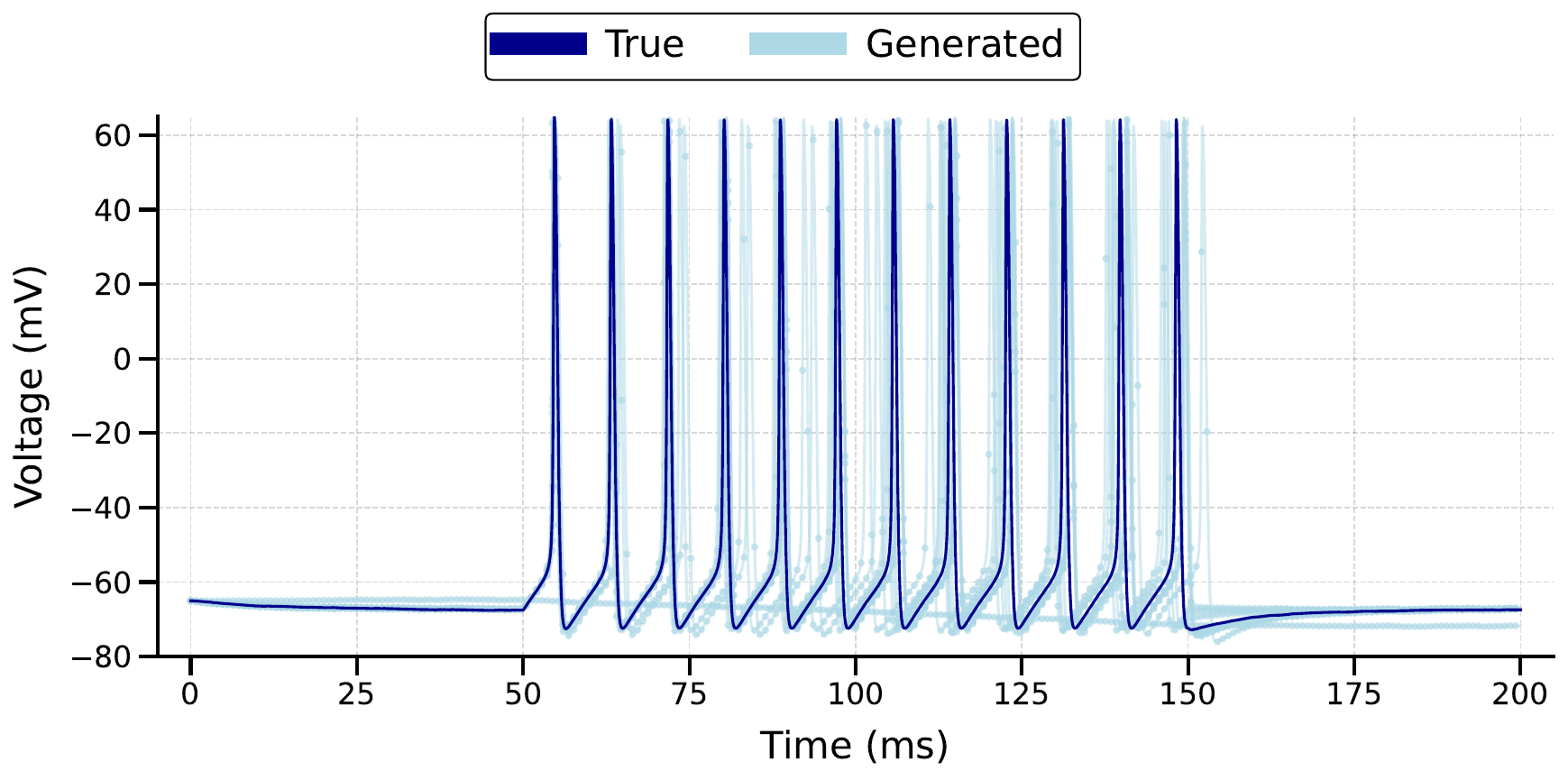}
        \caption{Posterior Predictive Voltage Traces.}
        \label{fig:simforme_volatge_plot}
    \end{subfigure}
    \caption{Simformer: Posterior Distributions for Hodgkin–Huxley model.}
    \label{fig:simformer_hh_posteriors}
\end{figure*}

\begin{figure*}[h]
    \centering
    \includegraphics[width=0.7\linewidth]{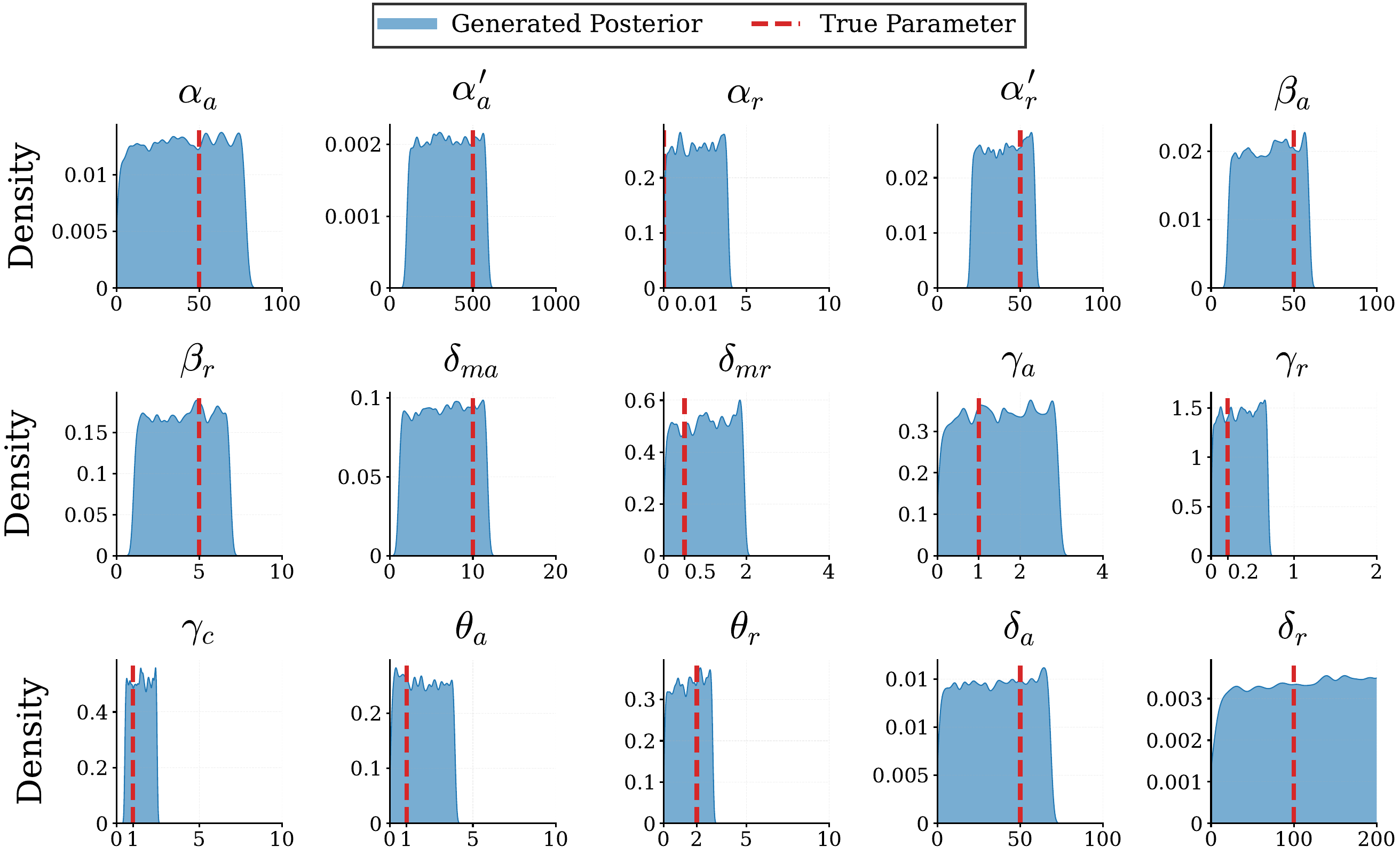}
    \caption{Simformer: Posterior Distributions for the Genetic Oscillator test problem.}
    \label{fig:vilar_posterior_simformer}
\end{figure*}

Both methods are trained with 30k simulations. For the HH model with fixed parameters $\boldsymbol{\theta^*} = (1.437, 72.177, 24.209, 0.154, 66.87, -81.435, -67.606)$, posterior samples and predictive checks (Figures~\ref{fig:simformer_hh_posteriors},~\ref{fig:hh_posteriors}) show both methods recover plausible voltage traces, with ConDiSim yielding tighter posterior support for several parameters. For the genetic oscillator, posterior distributions for Simformer and ConDiSim are shown in Figures~\ref{fig:vilar_posterior_simformer} and~\ref{fig:vilar_posterior}, respectively. Both produce reasonable posteriors, but ConDiSim achieves sharper concentration around true parameters for several rates (e.g., $\alpha_r, \beta_r, \gamma_c, \theta_a$), while Simformer yields broader supports. This demonstrates ConDiSim's more precise uncertainty quantification without sacrificing coverage, particularly effective for complex oscillatory dynamics.

\newpage
\section{Posterior Plots for SBI benchmarking 
problems}\label{posterior_plots_supp}
In this section, we summarize the unique challenges posed by the remaining SBI benchmark problem and highlight the performance of our model. Detailed posterior distributions are provided in the corresponding figures for visual reference.

\begin{figure}[!ht]
    \centering
    \subcaptionbox{GMM.\label{fig:gmm}}{%
        \includegraphics[width=0.45\textwidth]{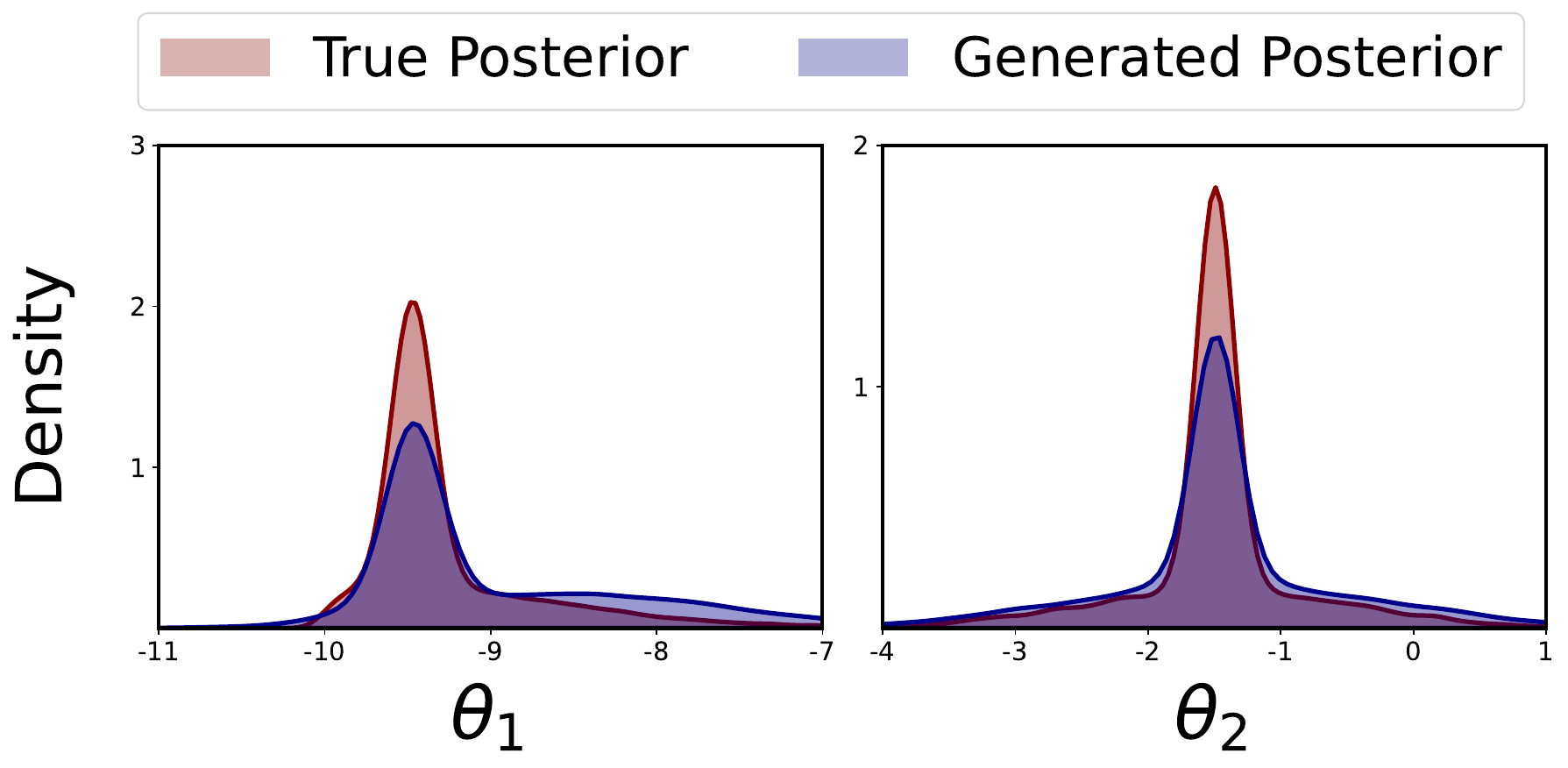}
    }
    \hspace{1mm}
    \subcaptionbox{SIR Model.\label{fig:sir}}{%
        \includegraphics[width=0.45\textwidth]{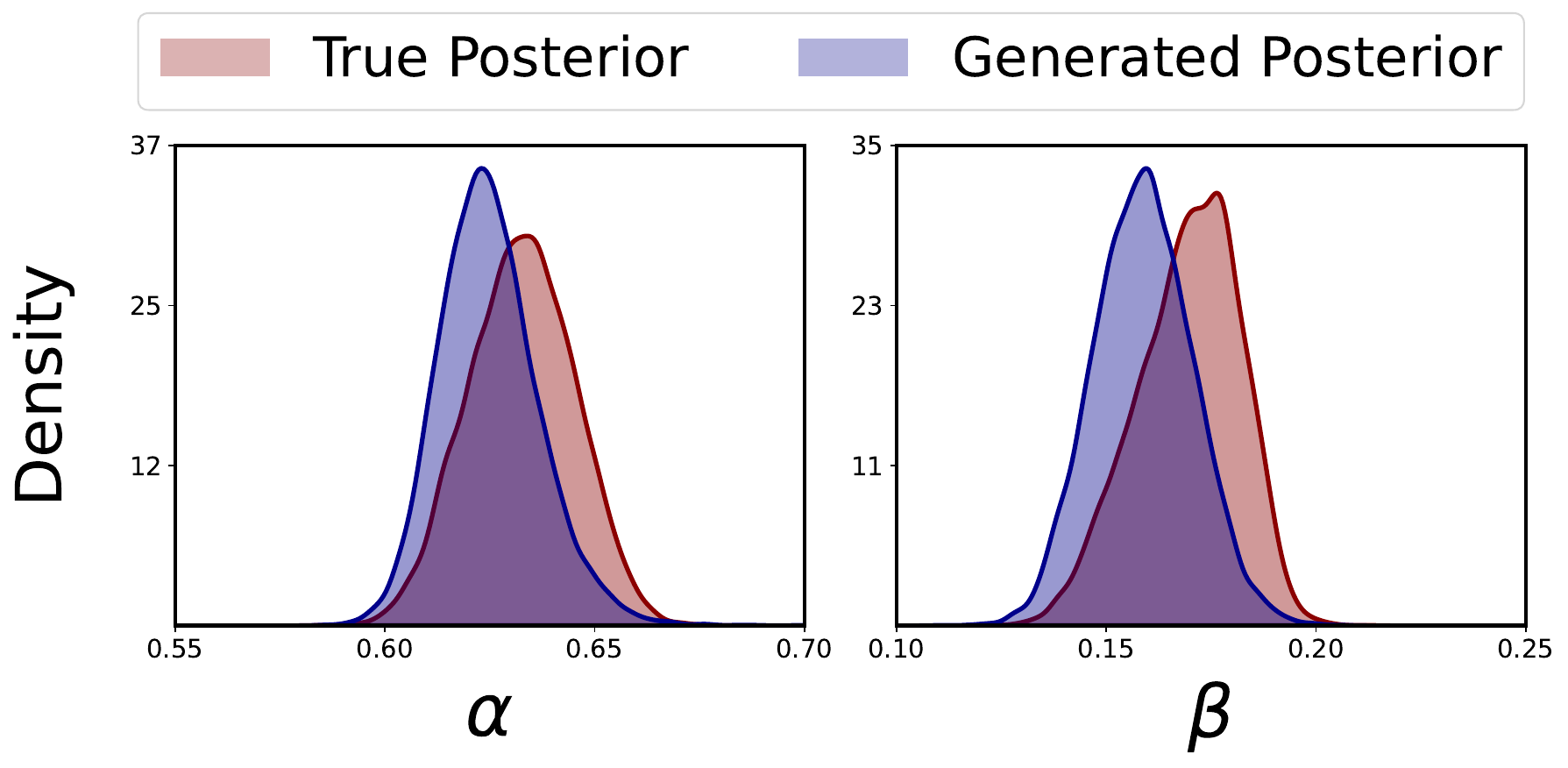}
    }
    \caption{\textbf{Posterior Plots:} Comparison of generated vs. reference posterior distributions for the GMM and SIR models.}
    \label{fig:gmm_sir}
\end{figure}

In the \textbf{Gaussian Mixture} (GMM) problem, the main challenge is identifying the shared mean between two Gaussian distributions with different variances. Figure~\ref{fig:gmm} demonstrates that our model effectively captures the primary modes and shapes of these distributions.

\begin{figure}[!ht]
    \centering
    \begin{subfigure}[t]{\linewidth}
        \centering
        \includegraphics[width=0.7\linewidth]{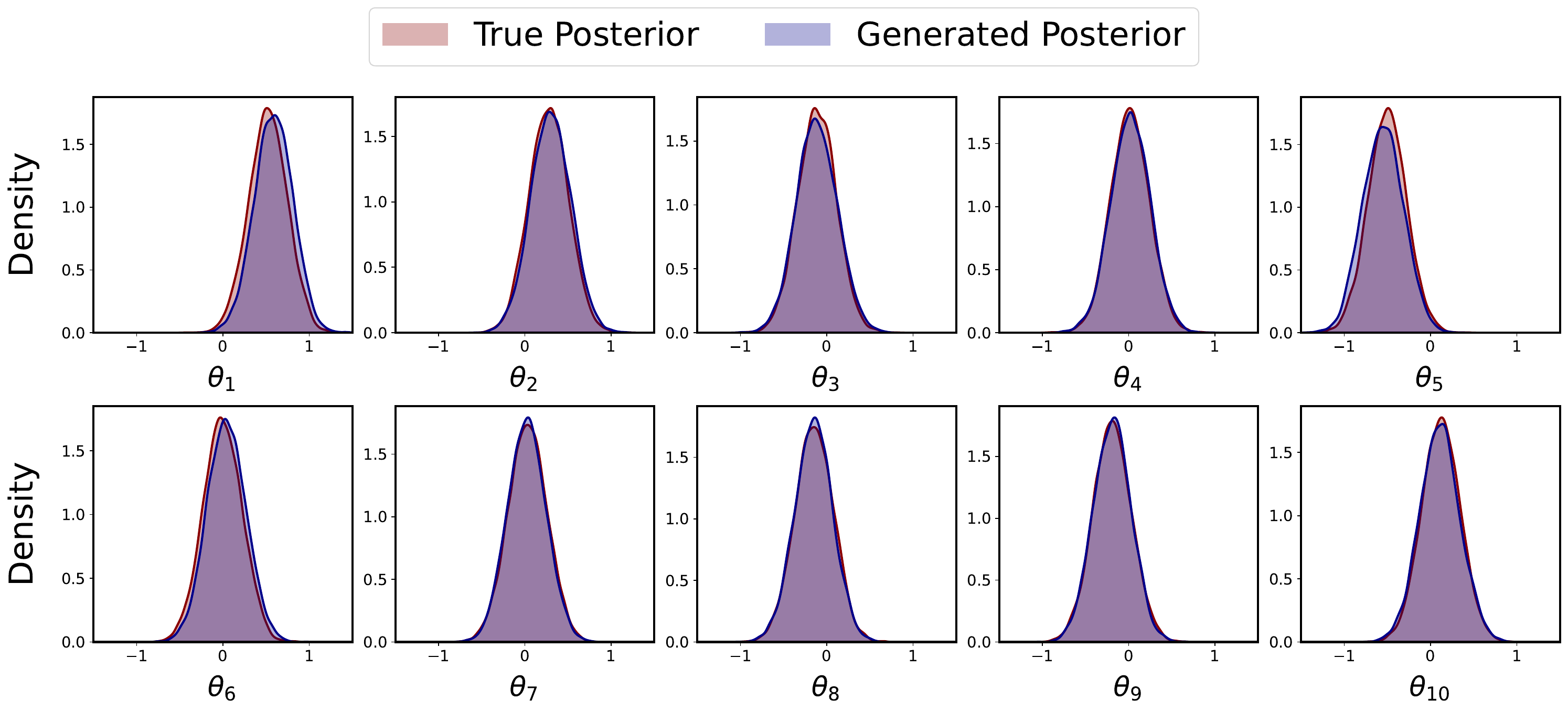} %
        \caption{Gaussian Linear Model.}
        \label{fig:gaussian_linear_posterior}
    \end{subfigure}
    \vfill
    \vspace*{0.01\linewidth}
    \begin{subfigure}[t]{\linewidth}
        \centering
        \includegraphics[width=0.7\linewidth]{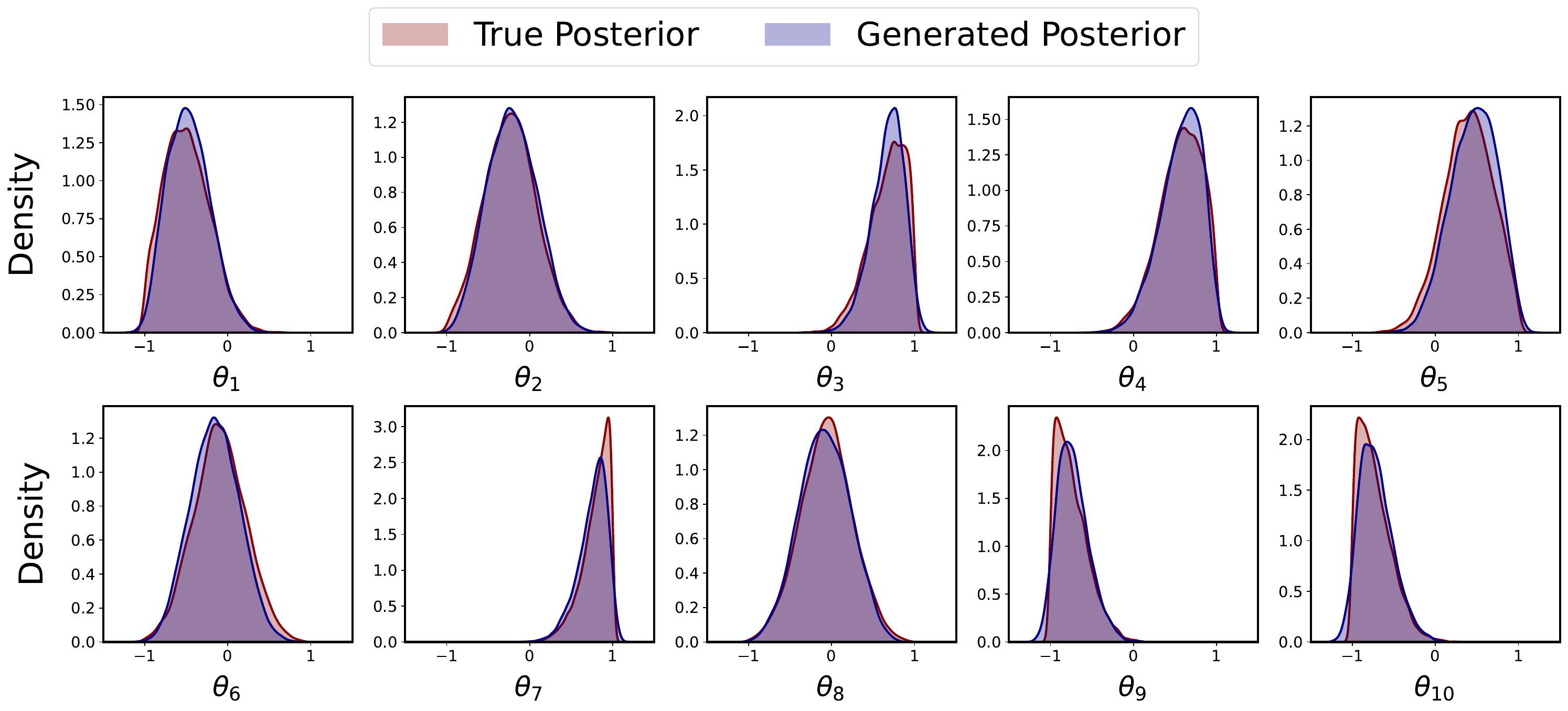}
        \caption{Gaussian Linear Uniform Model.}
        \label{fig:gaussian_linear_uniform_posterior}
    \end{subfigure}
    \caption{Posterior Distributions for Gaussian Linear and Gaussian Linear Uniform Models: Generated vs. Reference Posterior.}
    \label{fig:posteriors}
\end{figure}

For \textbf{Gaussian Linear} and \textbf{Gaussian Linear Uniform}, the challenge is accurately modeling the posterior under both conjugate and non-conjugate settings. The Gaussian Linear Uniform variant complicates posterior inference with a uniform prior. Figure~\ref{fig:gaussian_linear_posterior} and Figure~\ref{fig:gaussian_linear_uniform_posterior} show that our model effectively handles these differences.

\begin{figure}[!ht]
    \centering
    \begin{subfigure}[t]{\linewidth}
        \centering
        \includegraphics[width=0.5\linewidth]{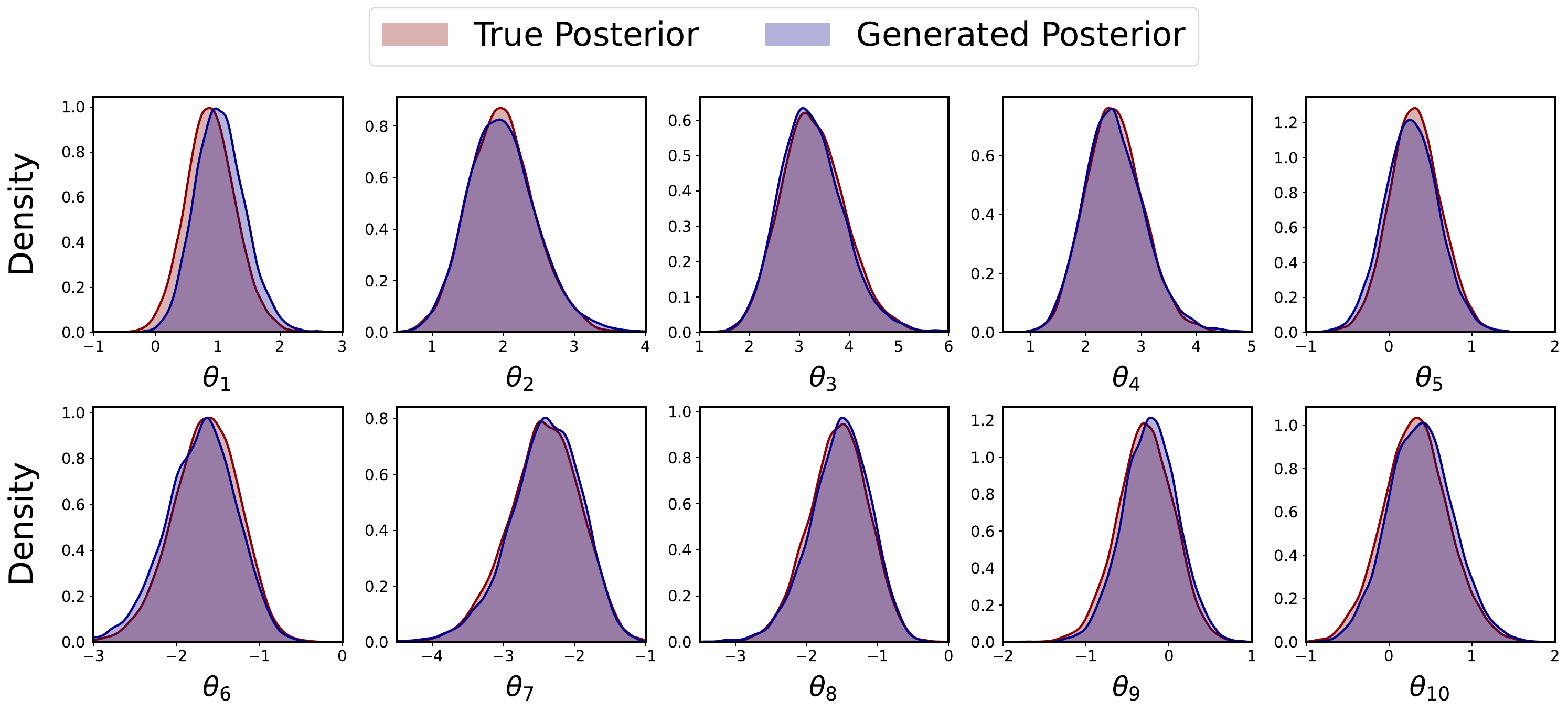} 
        \caption{Bernoulli GLM.}
        \label{fig:glm_posterior}
    \end{subfigure}
    \vfill
    \vspace*{0.01\linewidth}
    \begin{subfigure}[t]{\linewidth}
        \centering
        \includegraphics[width=0.5\linewidth]{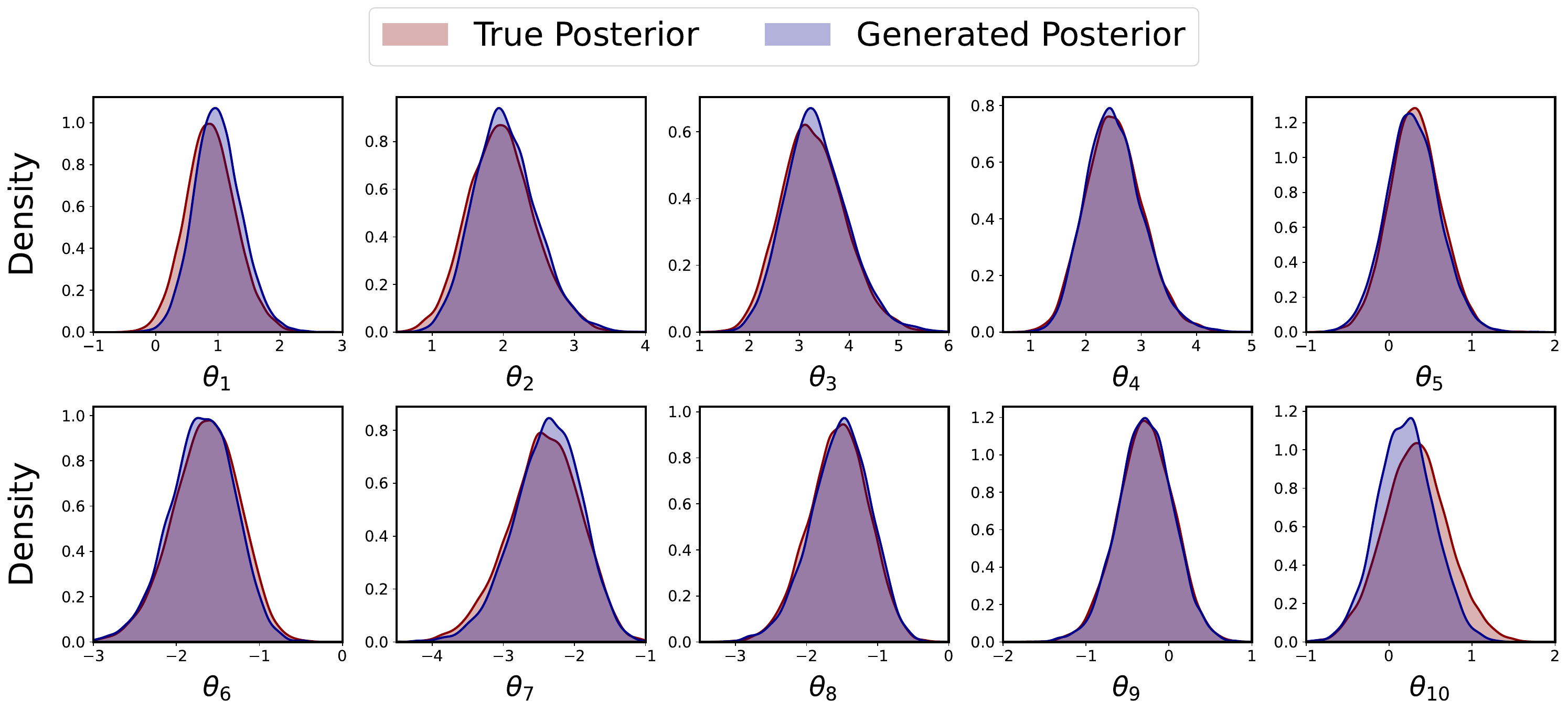}
        \caption{Bernoulli GLM Raw.}
        \label{fig:glm_raw_posterior}
    \end{subfigure}
    \caption{Posterior Distributions for Bernoulli GLM and Bernoulli GLM Raw: Generated vs. Reference Posterior.}
    \label{fig:posteriors}
\end{figure}

In \textbf{Bernoulli GLM} and \textbf{Bernoulli GLM Raw}, the challenge involves estimating posteriors from Bernoulli observations with Gaussian priors. The ``Raw'' variant introduces complexity by using raw data instead of sufficient statistics. Figures~\ref{fig:glm_posterior} and \ref{fig:glm_raw_posterior} indicate accurate posterior modeling for both cases.

The \textbf{SIR} model aims to infer contact and recovery rates from noisy infection data over time. Figure~\ref{fig:sir} illustrates that our model robustly estimates the parameters despite the high variability in the data.

\begin{figure}[!ht]
    \centering
    \includegraphics[width=0.8\textwidth]{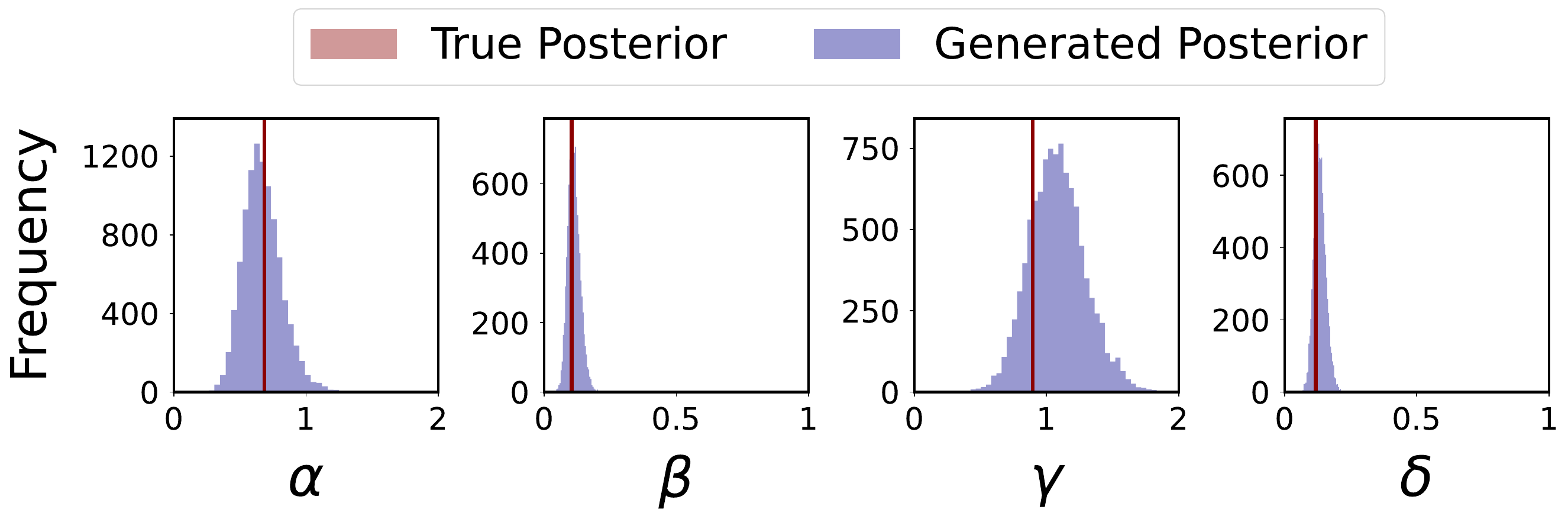}
    \caption{\textbf{Lotka Volterra Posterior:} Visualization of the posterior for the Lotka--Volterra model. The true posterior mean is indicated by a vertical line (due to low variance).}
    \label{fig:lv}
\end{figure}

Lastly, the \textbf{Lotka-Volterra} model is challenging due to its nonlinear dynamics and potential chaotic behavior. Figure~\ref{fig:lv} shows that our model successfully infers the underlying parameters, effectively navigating the model's complexity.

These experiments demonstrate that \textbf{ConDiSim} effectively infers posterior distributions across all benchmark problems with high accuracy.

\newpage
\section{Simulation Based Calibration}\label{sbc_supp}
Simulation-Based Calibration (SBC) is an important metric for generative models in SBI, as it verifies alignment between inferred posterior distributions and true parameter values, assisting in detecting potential biases in model-generated samples. For SBC, we generate \( M = 500 \) prior samples, simulate corresponding observations, and draw \( L = 250 \) posterior samples for each observation using the learned ConDiSim model. Calibration is assessed through rank statistics of true parameters within their posterior distributions, with empirical cumulative distribution functions (ECDFs) providing a robust, bin-free visualization of these ranks. Following \citep{S_ilynoja_2022}, ECDF-based SBC transforms posterior draws into fractional rank statistics, avoiding arbitrary binning and facilitating interpretation. A uniform ECDF indicates well-calibrated posteriors, while deviations suggest miscalibration: curves above the diagonal indicate overconfidence, and those below indicate underconfidence. Using \texttt{BayesFlow} \citep{bayesflow_2023_software}, we generate ECDF plots with 90\% confidence interval bands as calibration benchmarks. Solid blue lines represent rank ECDFs of posterior draws, with shaded grey areas denoting the 90\% confidence intervals. The SBC results for SBI benchmarking problems are presented below:

\begin{figure}[htbp]
    \centering

    \begin{subfigure}[t]{\linewidth}
        \centering
        \includegraphics[width=0.9\linewidth]{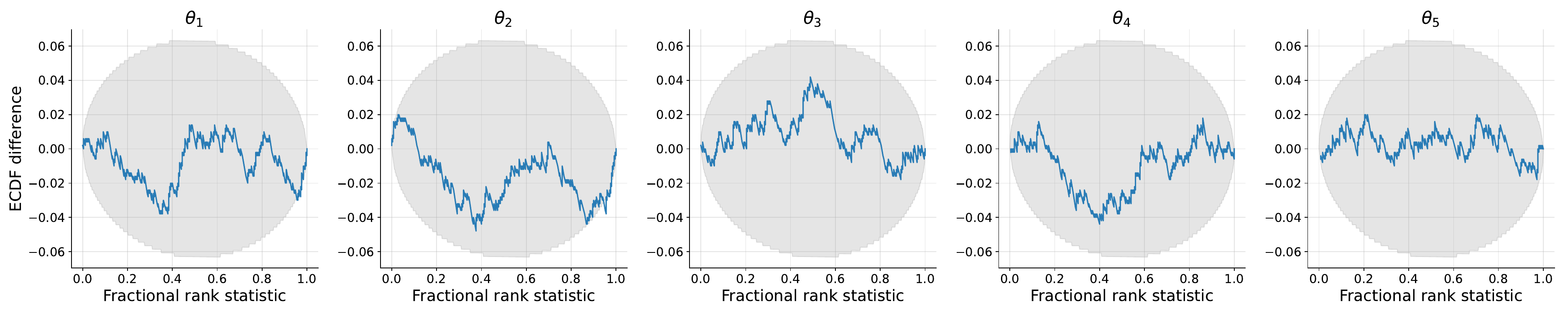} 
        \caption{SLCP.}
        \label{fig:slcp_ecdf}
    \end{subfigure}

    \vfill
    \vspace*{0.02\linewidth}
    \begin{subfigure}[t]{\linewidth}
        \centering
        \includegraphics[width=0.9\linewidth]{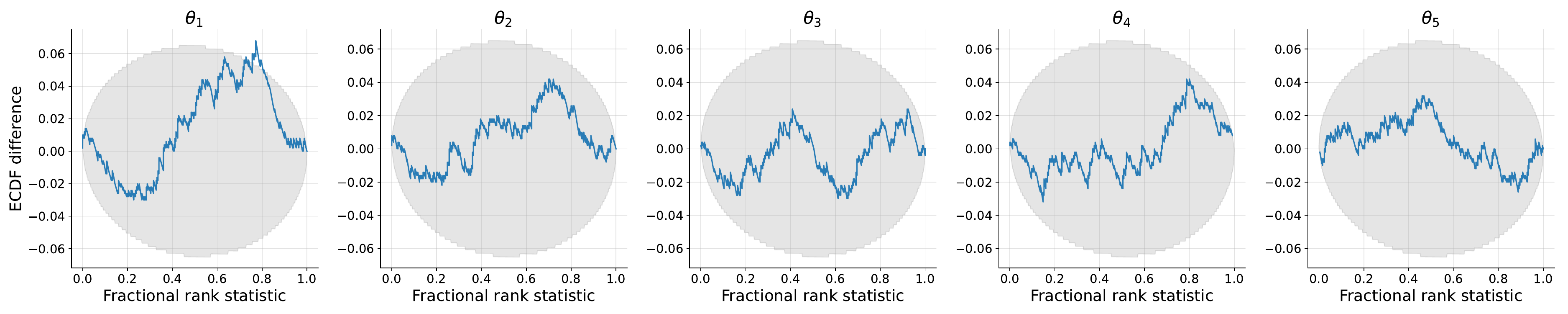} 
        \caption{SLCP Distractors.}
        \label{fig:slcp_dist_ecdf}
    \end{subfigure}

    \vfill
    \vspace*{0.02\linewidth}
    \begin{subfigure}[t]{\linewidth}
        \centering
        \includegraphics[width=0.9\linewidth]{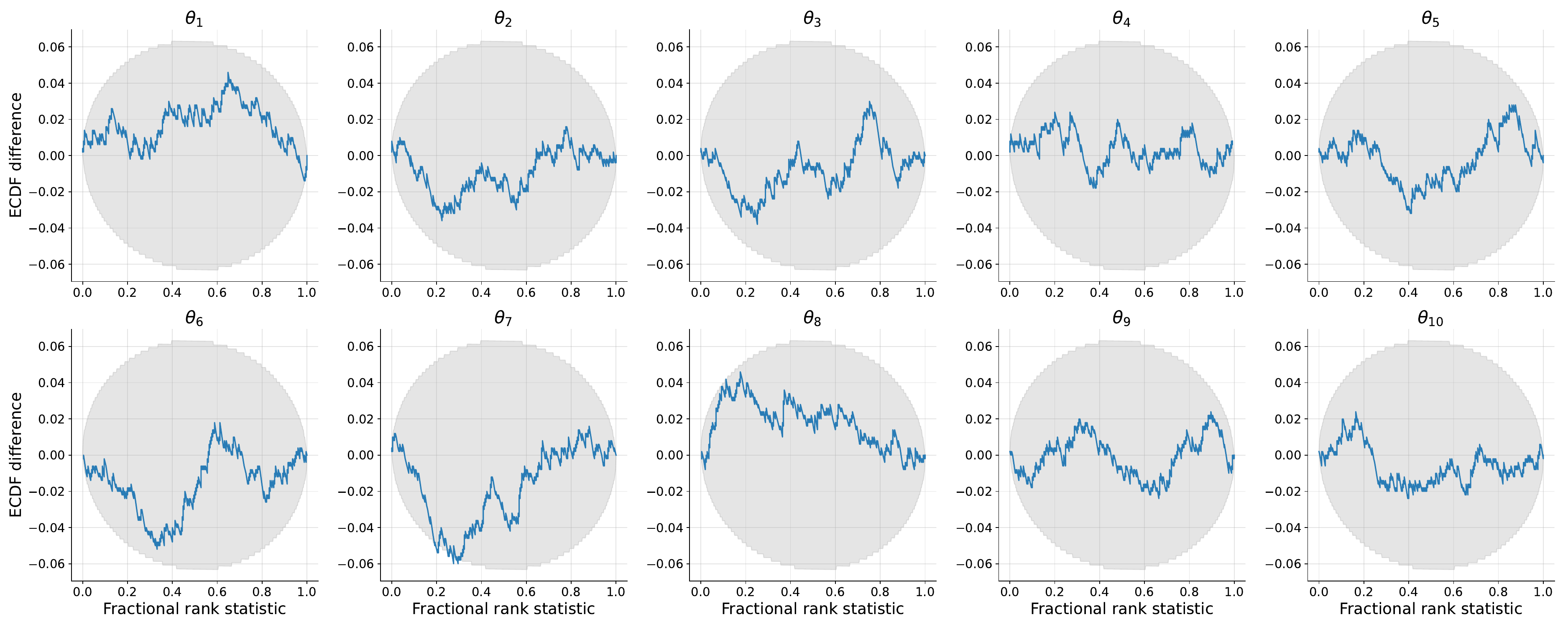} 
        \caption{Gaussian Linear.}
        \label{fig:gaussian_linear_ecdf}
    \end{subfigure}
    \vfill
    \vspace*{0.02\linewidth}
    \begin{subfigure}[t]{\linewidth}
        \centering
        \includegraphics[width=0.9\linewidth]{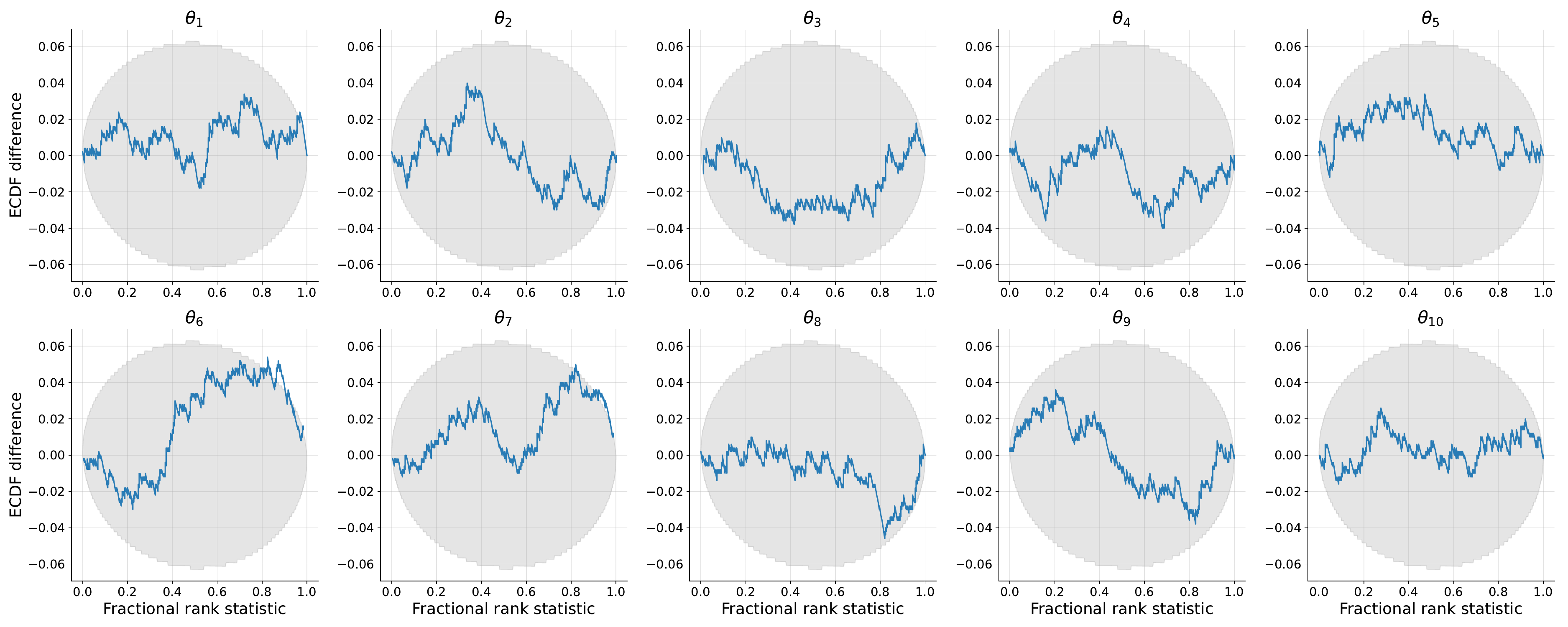} 
        \caption*{(d) Gaussian Linear Uniform.}
        \label{fig:ecdf_ecdf}
    \end{subfigure}

    \label{fig:ecdf_joint}
\end{figure}

\begin{figure}[htbp]
    \centering
    \begin{subfigure}[t]{\linewidth}
        \centering
        \includegraphics[width=0.9\linewidth]{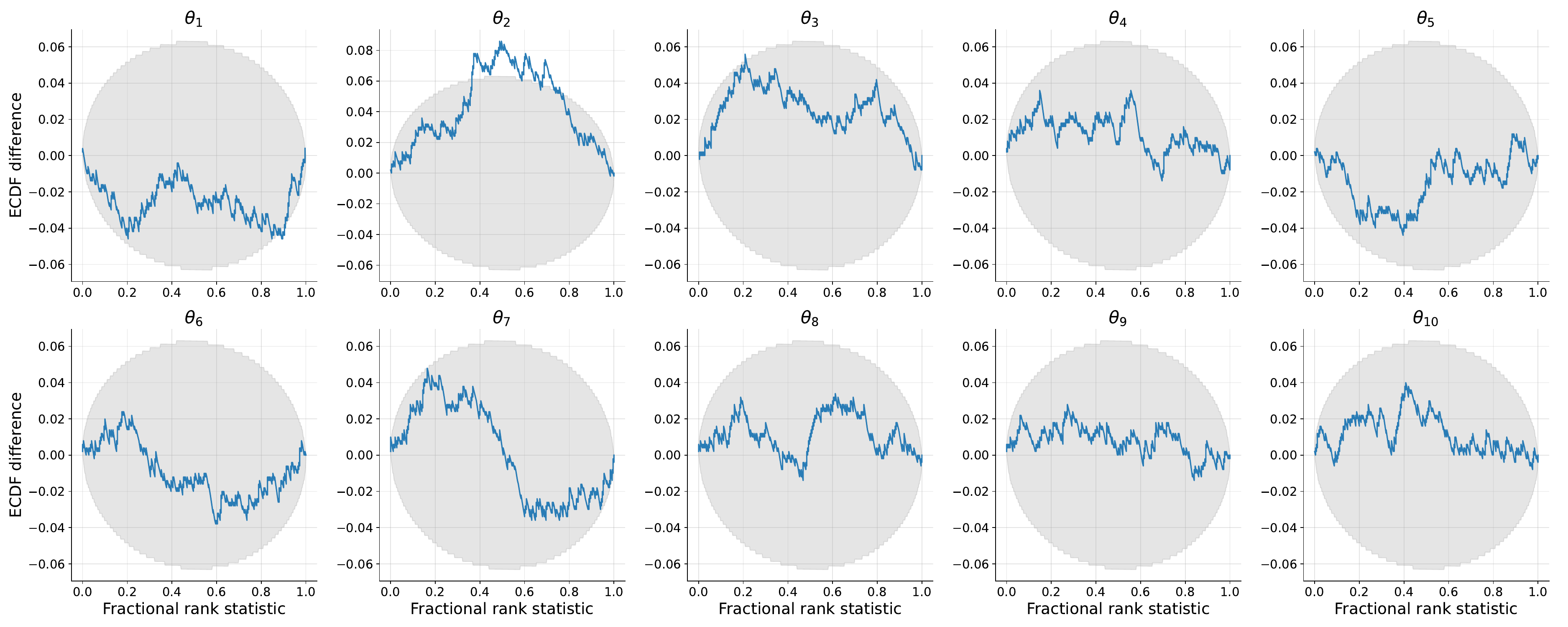} 
        \caption*{(e) Bernoulli GLM.}
        \label{fig:bernoulli_glm_ecdf}
    \end{subfigure}
    \vfill
    \vspace*{0.02\linewidth}
    \begin{subfigure}[t]{\linewidth}
        \centering
        \includegraphics[width=0.9\linewidth]{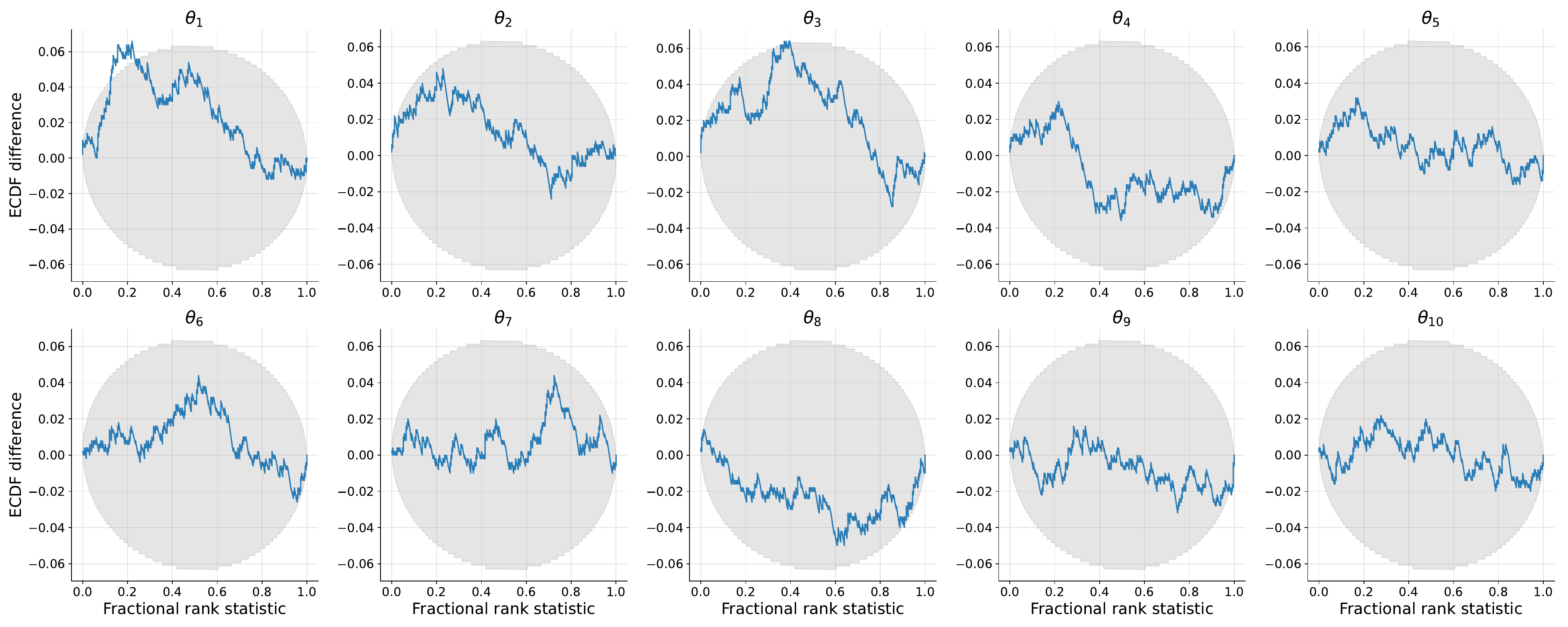} 
        \caption*{(f) Bernoulli GLM Raw.}
        \label{fig:bernoulli_glm_raw_ecdf}
    \end{subfigure}
    \begin{subfigure}[t]{0.48\linewidth}
        \centering
        \includegraphics[width=\linewidth]{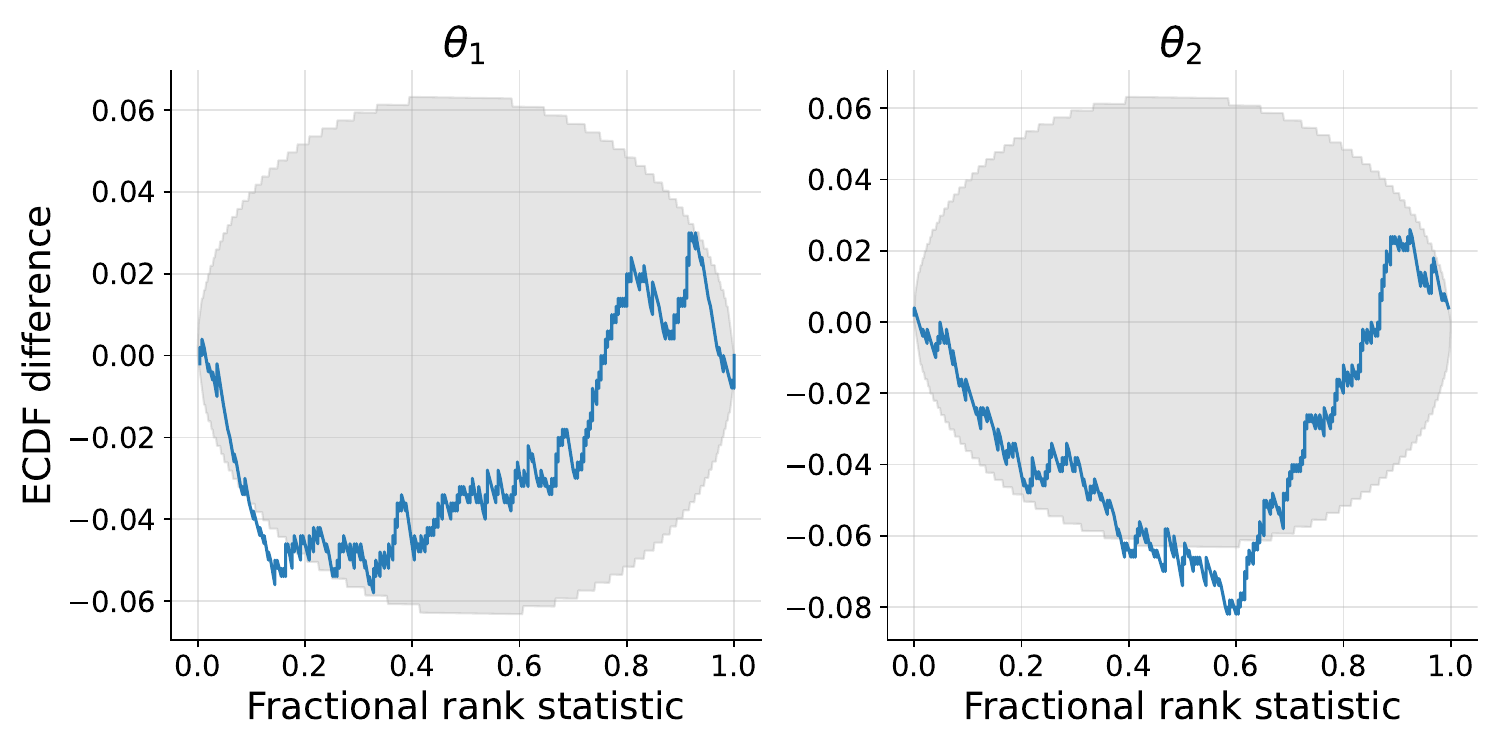}
        \caption*{(g) SIR.}
        \label{fig:sir_ecdf}
    \end{subfigure}%
    \hfill
    \begin{subfigure}[t]{0.48\linewidth}
        \centering
        \includegraphics[width=\linewidth]{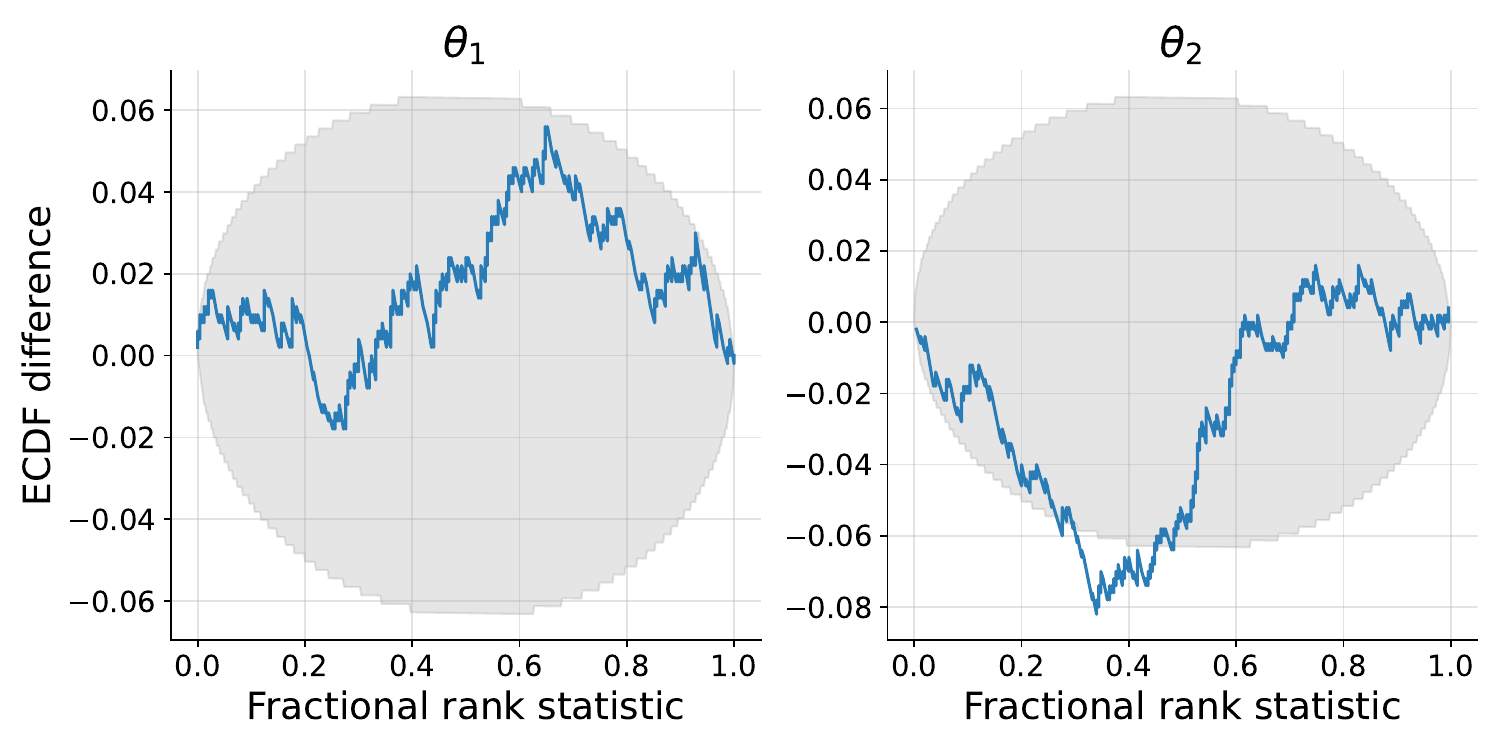}
        \caption*{(h) Gaussian Mixture.}
        \label{fig:gmm_ecdf}
    \end{subfigure}

    \vspace{1em} 

    \begin{subfigure}[t]{0.8\linewidth}
        \centering
        \includegraphics[width=\linewidth]{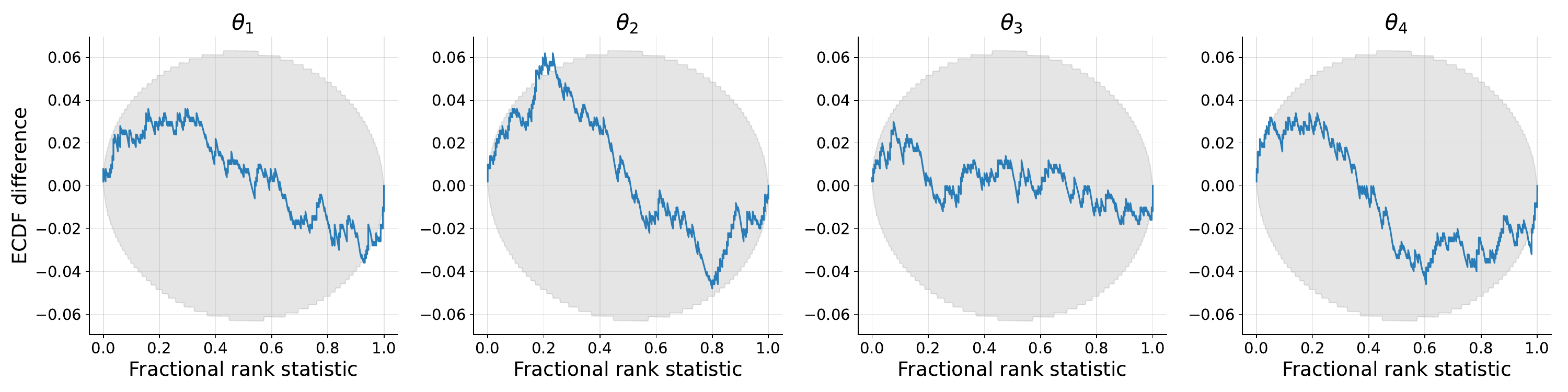}
        \caption*{(i) Lotka Volterra.}
        \label{fig:lv_ecdf}
    \end{subfigure}

    \caption{ECDF results for the 30,000 simulation budget and the first run for each problem.}

    \label{fig:ecdf_joint_3}
\end{figure}

\newpage
\section{SBI Metrics}
\subsection{Limitations of the C2ST Metric}
The Classifier Two-Sample Test (C2ST) evaluates the similarity between a generated posterior \(P\) and the true posterior \(Q\) by training a classifier to differentiate their samples. A score near 0.5 indicates that the two distributions are similar, while higher values indicate discrepancies. Although widely used in SBI literature for its straightforward interpretability providing a clear score between 0.5 and 1, the metric can be overly sensitive, as a single poorly estimated dimension may distort the results. Furthermore, its reliability depends significantly on the design and training of the classifier \citep{Friedman03, Lopez16}. However, C2ST remains the most widely used metric in SBI, as it is both interpretable and well established for evaluating benchmark problems. For completeness, we also report MMD results in the following section, although MMD is kernel dependent, less commonly used, and often difficult to interpret solely from numerical values.

\subsection{MMD}
The Maximum Mean Discrepancy (MMD) measures the distance between the mean embeddings of \(P\) and \(Q\) in a reproducing kernel Hilbert space (RKHS). MMD is defined as,
\begin{equation}
\mathrm{MMD}^2(P, Q; \mathcal{H}) = \left\| \mathbb{E}_{x \sim P}[\phi(x)] - \mathbb{E}_{y \sim Q}[\phi(y)] \right\|_{\mathcal{H}}^2,
\end{equation}
where \(\phi\) is a feature mapping into the RKHS \(\mathcal{H}\). A low MMD value indicates that the generated posterior closely approximates the true posterior. MMD therefore provides a  holistic measure of similarity between distributions without being overly influenced by discrepancies in a single dimension. A low MMD value indicates that the distributions \(P\) and \(Q\) are similar, while a larger value suggests significant differences.

\begin{table*}[h!]
\centering
\caption{MMD metric for different methods across simulation budgets (mean $\pm$ std. dev.).}
\label{tab:mmd_global}
\renewcommand{\arraystretch}{1.25}
\setlength{\tabcolsep}{5pt}

\begin{subtable}[t]{\textwidth}
\centering
\caption{Two Moons, Bernoulli GLM, and SIR.}
\label{tab:mmd_global_1}
\begin{adjustbox}{max width=\textwidth}
\begin{tabular}{l|ccc|ccc|ccc}
\rowcolor{gray!10}
\textbf{Method}
  & \multicolumn{3}{c|}{\textbf{Two Moons}}
  & \multicolumn{3}{c|}{\textbf{Bernoulli GLM}}
  & \multicolumn{3}{c}{\textbf{SIR}} \\

\rowcolor{gray!10}
\textbf{Budget}
  & \textbf{10,000} & \textbf{20,000} & \textbf{30,000}
  & \textbf{10,000} & \textbf{20,000} & \textbf{30,000}
  & \textbf{10,000} & \textbf{20,000} & \textbf{30,000} \\
\hline

NPE
& 0.0254 \tiny{$\pm$0.0025}
& 0.0241 \tiny{$\pm$0.0099}
& 0.0125 \tiny{$\pm$0.0045}
& 1.0371 \tiny{$\pm$0.1830}
& 0.9379 \tiny{$\pm$0.1001}
& 0.8675 \tiny{$\pm$0.1945}
& 0.6611 \tiny{$\pm$0.2236}
& 1.0312 \tiny{$\pm$0.1255}
& 1.1903 \tiny{$\pm$0.1007} \\

APT
& \underline{0.0073} \tiny{$\pm$0.0030}
& 0.0069 \tiny{$\pm$0.0030}
& \underline{0.0037} \tiny{$\pm$0.0012}
& 0.7383 \tiny{$\pm$0.0853}
& 0.4944 \tiny{$\pm$0.0663}
& 0.5808 \tiny{$\pm$0.0539}
& 1.1003 \tiny{$\pm$0.0712}
& 1.1093 \tiny{$\pm$0.0311}
& 1.1291 \tiny{$\pm$0.0481} \\

GATSBI
& 0.1213 \tiny{$\pm$0.0650}
& 0.0582 \tiny{$\pm$0.0122}
& 0.1432 \tiny{$\pm$0.1161}
& 3.6381 \tiny{$\pm$0.5428}
& 4.8299 \tiny{$\pm$0.6825}
& 3.9965 \tiny{$\pm$1.8510}
& 2.9470 \tiny{$\pm$0.4891}
& 2.7502 \tiny{$\pm$0.7268}
& 2.1466 \tiny{$\pm$0.5625} \\
Simformer
& \textbf{0.0005} \tiny{$\pm$0.0003}
& \textbf{0.0001} \tiny{$\pm$0.0001}
& \textbf{0.0003} \tiny{$\pm$0.0004}
& \textbf{0.0191} \tiny{$\pm$0.0098}
& \textbf{0.0156} \tiny{$\pm$0.0087}
& \textbf{0.0116} \tiny{$\pm$0.0061}
& \textbf{0.0018} \tiny{$\pm$0.0028}
& \textbf{0.0016} \tiny{$\pm$0.0042}
& \textbf{0.0016} \tiny{$\pm$0.0085} \\

FMPE
& 0.0133 \tiny{$\pm$ 0.0197}  
& \underline{0.0039} \tiny{$\pm$ 0.0060}  
& 0.019 \tiny{$\pm$ 0.0014}  
& 0.1076 \tiny{$\pm$ 0.0283}  
& 0.0826 \tiny{$\pm$ 0.0138}  
& 0.0746 \tiny{$\pm$ 0.0247}  
& \underline{0.2058} \tiny{$\pm$ 0.0024}  
& \underline{0.3618} \tiny{$\pm$ 0.0048}  
& \underline{0.1819} \tiny{$\pm$ 0.0044}

\\

\rowcolor{blue!5}
ConDiSim
& 0.0172 \tiny{$\pm$0.0226}
& 0.0091 \tiny{$\pm$0.0099}
& 0.0052 \tiny{$\pm$0.0018}
& \underline{0.0393} \tiny{$\pm$0.0095}
& \underline{0.0525} \tiny{$\pm$0.0547}
& \underline{0.0190} \tiny{$\pm$0.0134}
& 0.2413 \tiny{$\pm$0.0681}
& 0.3853 \tiny{$\pm$0.1195}
& 0.4860 \tiny{$\pm$0.1198} \\

\end{tabular}
\end{adjustbox}
\end{subtable}

\vspace{0.3cm} 

\begin{subtable}[t]{\textwidth}
\centering
\caption{SLCP Distractors, Bernoulli GLM Raw, and Lotka Volterra.}
\label{tab:mmd_global_2}
\begin{adjustbox}{max width=\textwidth}
\begin{tabular}{l|ccc|ccc|ccc}
\rowcolor{gray!10}
\textbf{Method}
  & \multicolumn{3}{c|}{\textbf{SLCP Distractors}}
  & \multicolumn{3}{c|}{\textbf{Bernoulli GLM Raw}}
  & \multicolumn{3}{c}{\textbf{Lotka Volterra}} \\

\rowcolor{gray!10}
\textbf{Budget}
  & \textbf{10,000} & \textbf{20,000} & \textbf{30,000}
  & \textbf{10,000} & \textbf{20,000} & \textbf{30,000}
  & \textbf{10,000} & \textbf{20,000} & \textbf{30,000} \\
\hline

NPE
& 1.2167 \tiny{$\pm$0.2397}
& 0.7834 \tiny{$\pm$0.3453}
& 0.6428 \tiny{$\pm$0.1779}
& 1.1227 \tiny{$\pm$0.0987}
& 1.1389 \tiny{$\pm$0.1878}
& 1.1448 \tiny{$\pm$0.1287}
& 2.6357 \tiny{$\pm$0.3468}
& 3.8635 \tiny{$\pm$0.2770}
& 4.3310 \tiny{$\pm$0.0709} \\

APT
& 1.8380 \tiny{$\pm$0.4448}
& 0.8278 \tiny{$\pm$0.2373}
& 0.4422 \tiny{$\pm$0.1428}
& 0.7233 \tiny{$\pm$0.2131}
& 0.6968 \tiny{$\pm$0.0826}
& 0.5437 \tiny{$\pm$0.1333}
& 3.1921 \tiny{$\pm$0.8198}
& 3.2722 \tiny{$\pm$0.8169}
& 4.4739 \tiny{$\pm$0.9642} \\

GATSBI
& 5.1993 \tiny{$\pm$1.2320}
& 7.2478 \tiny{$\pm$1.2260}
& 3.0631 \tiny{$\pm$0.3244}
& 3.4369 \tiny{$\pm$0.8517}
& 4.3479 \tiny{$\pm$0.5433}
& 3.5950 \tiny{$\pm$0.8613}
& 7.2684 \tiny{$\pm$0.5733}
& 5.5183 \tiny{$\pm$1.7126}
& 4.8858 \tiny{$\pm$1.2146} \\
Simformer
& \textbf{0.0540} \tiny{$\pm$0.0173}
& \textbf{0.0541} \tiny{$\pm$0.0161}
& \underline{0.0552} \tiny{$\pm$0.0195}
& 0.2056 \tiny{$\pm$0.0486}
& 0.0926 \tiny{$\pm$0.0493}
& \underline{0.0731} \tiny{$\pm$0.0238}
& \textbf{0.9696} \tiny{$\pm$0.4434}
& \textbf{0.8342} \tiny{$\pm$0.4742}
& \textbf{0.9634} \tiny{$\pm$0.2588} \\

FMPE
& 0.1598 \tiny{$\pm$ 0.0501}  
& 0.1403 \tiny{$\pm$ 0.0471}  
& 0.0930 \tiny{$\pm$ 0.0561}  
& \textbf{0.0859} \tiny{$\pm$ 0.0167}  
& \underline{0.0831} \tiny{$\pm$ 0.0181}  
& 0.0781 \tiny{$\pm$ 0.0102}  
& \underline{1.0537} \tiny{$\pm$ 0.6277}  
& \underline{1.0858} \tiny{$\pm$ 0.5639}  
& \underline{1.1696} \tiny{$\pm$ 0.4336}

\\

\rowcolor{blue!5}
ConDiSim
& \underline{0.1378} \tiny{$\pm$0.0404}
& \underline{0.0580} \tiny{$\pm$0.0205}
& \textbf{0.0275} \tiny{$\pm$0.0182}
& \underline{0.0968} \tiny{$\pm$0.1012}
& \textbf{0.0359} \tiny{$\pm$0.0212}
& \textbf{0.0305} \tiny{$\pm$0.0279}
& 2.3226 \tiny{$\pm$0.1486}
& 2.1965 \tiny{$\pm$0.3856}
& 2.1180 \tiny{$\pm$0.2584} \\

\end{tabular}
\end{adjustbox}
\end{subtable}

\vspace{0.3cm} 

\begin{subtable}[t]{\textwidth}
\centering
\caption{Gaussian Mixture, Gaussian Linear, Gaussian Linear Uniform, and SLCP.}
\renewcommand{\arraystretch}{1.5}
\setlength{\tabcolsep}{2pt}
\label{tab:mmd_global_3}
\begin{adjustbox}{max width=\textwidth}
\begin{tabular}{l|ccc|ccc|ccc|ccc}
\rowcolor{gray!10}
\textbf{Method}
  & \multicolumn{3}{c|}{\textbf{Gaussian Mixture}}
  & \multicolumn{3}{c|}{\textbf{Gaussian Linear}}
  & \multicolumn{3}{c|}{\textbf{Gaussian Linear Uniform}}
  & \multicolumn{3}{c}{\textbf{SLCP}} \\

\rowcolor{gray!10}
\textbf{Budget}
  & \textbf{10,000} & \textbf{20,000} & \textbf{30,000}
  & \textbf{10,000} & \textbf{20,000} & \textbf{30,000}
  & \textbf{10,000} & \textbf{20,000} & \textbf{30,000}
  & \textbf{10,000} & \textbf{20,000} & \textbf{30,000} \\
\hline
GATSBI
& 0.2242 \tiny{$\pm$0.0296}
& 0.1149 \tiny{$\pm$0.0549}
& 0.1733 \tiny{$\pm$0.1082}
& 0.4999 \tiny{$\pm$0.1595}
& 0.6476 \tiny{$\pm$0.3050}
& 0.7182 \tiny{$\pm$0.2839}
& 0.9044 \tiny{$\pm$0.2353}
& 0.7712 \tiny{$\pm$0.1483}
& 1.1197 \tiny{$\pm$0.4795}
& 0.7197 \tiny{$\pm$0.5598}
& 0.3375 \tiny{$\pm$0.1473}
& 0.7948 \tiny{$\pm$0.5992} \\
NPE
& 0.0604 \tiny{$\pm$0.0260}
& 0.0418 \tiny{$\pm$0.0211}
& 0.0554 \tiny{$\pm$0.0347}
& 0.0105 \tiny{$\pm$0.0036}
& 0.0074 \tiny{$\pm$0.0018}
& 0.0082 \tiny{$\pm$0.0011}
& 0.0258 \tiny{$\pm$0.0098}
& 0.0117 \tiny{$\pm$0.0041}
& 0.0121 \tiny{$\pm$0.0051}
& 0.4021 \tiny{$\pm$0.0551}
& 0.1632 \tiny{$\pm$0.0684}
& 0.2333 \tiny{$\pm$0.0935} \\

APT
& \underline{0.0137} \tiny{$\pm$0.0031}
& \underline{0.0108} \tiny{$\pm$0.0088}
& 0.0110 \tiny{$\pm$0.0034}
& \textbf{0.0075} \tiny{$\pm$0.0013}
& \underline{0.0070} \tiny{$\pm$0.0018}
& \underline{0.0080} \tiny{$\pm$0.0008}
& \underline{0.0166} \tiny{$\pm$0.0087}
& \underline{0.0138} \tiny{$\pm$0.0053}
& \underline{0.0113} \tiny{$\pm$0.0039}
& 0.1405 \tiny{$\pm$0.0347}
& \underline{0.0487} \tiny{$\pm$0.0296}
& 0.0287 \tiny{$\pm$0.0166} \\

GATSBI
& 0.2242 \tiny{$\pm$0.0296}
& 0.1149 \tiny{$\pm$0.0549}
& 0.1733 \tiny{$\pm$0.1082}
& 0.4999 \tiny{$\pm$0.1595}
& 0.6476 \tiny{$\pm$0.3050}
& 0.7182 \tiny{$\pm$0.2839}
& 0.9044 \tiny{$\pm$0.2353}
& 0.7712 \tiny{$\pm$0.1483}
& 1.1197 \tiny{$\pm$0.4795}
& 0.7197 \tiny{$\pm$0.5598}
& 0.3375 \tiny{$\pm$0.1473}
& 0.7948 \tiny{$\pm$0.5992} \\

Simformer
& \textbf{0.0009} \tiny{$\pm$0.0007}
& \textbf{0.0009} \tiny{$\pm$0.0006}
& \textbf{0.0007} \tiny{$\pm$0.0010}
& \underline{0.0092} \tiny{$\pm$0.0116}
& \textbf{0.0037} \tiny{$\pm$0.0035}
& \textbf{0.0037} \tiny{$\pm$0.0028}
& \textbf{0.0014} \tiny{$\pm$0.0006}
& \textbf{0.0015} \tiny{$\pm$0.0011}
& \textbf{0.0008} \tiny{$\pm$0.0003}
& \textbf{0.0323} \tiny{$\pm$0.0203}
& \textbf{0.0025} \tiny{$\pm$0.0034}
& \textbf{0.0021} \tiny{$\pm$0.0030} \\

FMPE
& 0.1893 \tiny{$\pm$ 0.0985}  
& 0.0208 \tiny{$\pm$ 0.0078}  
& \underline{0.0013} \tiny{$\pm$ 0.0011}  
& 0.0511 \tiny{$\pm$ 0.0053}  
& 0.0500 \tiny{$\pm$ 0.0041}  
& 0.0497 \tiny{$\pm$ 0.0034}  
& 0.1425 \tiny{$\pm$ 0.0219}  
& 0.1261 \tiny{$\pm$ 0.0206 } 
& 0.1236 \tiny{$\pm$ 0.0247}
& 0.1573 \tiny{$\pm$ 0.0543}  
& 0.1489 \tiny{$\pm$ 0.0528}  
& 0.1489 \tiny{$\pm$ 0.0528}

\\

\rowcolor{blue!5}
ConDiSim
& 0.1451 \tiny{$\pm$0.0269}
& 0.1047 \tiny{$\pm$0.0314}
& 0.0686 \tiny{$\pm$0.0228}
& 0.0591 \tiny{$\pm$0.0249}
& 0.0165 \tiny{$\pm$0.0083}
& 0.0116 \tiny{$\pm$0.0032}
& 0.0462\tiny{$\pm$0.0321}
& 0.0297 \tiny{$\pm$0.0252}
& 0.0291 \tiny{$\pm$0.0459}
& \underline{0.0996} \tiny{$\pm$0.0287}
& 0.0510 \tiny{$\pm$0.0345}
& \underline{0.0267} \tiny{$\pm$0.0178} \\

\end{tabular}
\end{adjustbox}
\end{subtable}
\end{table*}

\end{document}